%% file: main.tex
\documentclass[twoside]{article}

\usepackage[accepted]{style/aistats2021/aistats2021}

\usepackage[round]{natbib}

\usepackage[utf8]{inputenc}  % allow utf-8 input
\usepackage[T1]{fontenc}     % use 8-bit T1 fonts
\usepackage{url}             % simple URL typesetting
\usepackage{amsfonts}        % blackboard math symbols
\usepackage{booktabs}        % professional-quality tables
\usepackage{nicefrac}        % compact symbols for 1/2, etc.
\usepackage{microtype}       % microtypography

\usepackage{bibunits}

\defaultbibliography{references}
\defaultbibliographystyle{humannat} % humannat

\input{preamble/preamble.tex}
\input{preamble/math.tex}

\usepackage[page,header]{appendix}

\begin{document}

\runningauthor{Jan-Matthis Lueckmann, Jan Boelts, David S. Greenberg, Pedro J. Gon\c{c}alves, Jakob H. Macke}

\twocolumn[

\aistatstitle{Benchmarking Simulation-Based Inference}

\aistatsauthor{%
    \textbf{Jan-Matthis Lueckmann}$^{1,2}$\hspace{0.4cm} \textbf{Jan Boelts}$^{2}$\hspace{0.4cm}
    David S. \textbf{Greenberg}$^{2,3}$
    \\
    \textbf{Pedro J. Gon\c{c}alves}$^{4}$\hspace{0.4cm}
    \textbf{Jakob H. Macke}$^{1,2,5}$\hspace{0.4cm}\vspace{0.2cm}
}

\aistatsaddress{$^{1}$University of Tübingen\hspace{0.4cm}$^{2}$Technical University of Munich\hspace{0.4cm}$^{3}$Helmholtz Centre Geesthacht\\$^{4}$Research Center caesar\hspace{0.4cm}$^{5}$Max Planck Institute for Intelligent Systems, Tübingen}

]

\setlength{\pdfpageheight}{11in}
\setlength{\pdfpagewidth}{8.5in}

\input{00_abstract}
 
\begin{bibunit}

\input{01_introduction}
\input{02_benchmark}
\input{03_results}
\input{05_discussion}

\addcontentsline{toc}{section}{\protect\numberline{\thesection}{References}}
\putbib
\end{bibunit}

\onecolumn
\newpage
\clearpage

\appendix
\appendixpage
\startcontents[section]
\printcontents[section]{l}{0}{\setcounter{tocdepth}{3}}
\setcounter{page}{1}
\setcounter{section}{0}

\begin{bibunit}

\newpage
\setcounter{figure}{0}

\input{appendix/A_algorithms}
\newpage

\input{appendix/B_benchmark}
\newpage

\input{appendix/F_figures}
\newpage

\input{appendix/H_hyperparams}

\newpage

\input{appendix/M_metrics}

\newpage

\input{appendix/R_runtimes}
\newpage

\input{appendix/T_tasks}
\newpage

\input{appendix/W_website}

\newpage

\clearpage
\refstepcounter{section}
\renewcommand{\thesection}{}
\addcontentsline{toc}{section}{\protect\numberline{\thesection}{References}}

\putbib
\end{bibunit}

\stopcontents[section]

\end{document}

%% file: preamble/preamble.tex
\newcommand{\link}[3][blue]{\texttt{\href{#2}{\color{#1}{#3}}}}
\newcommand{\linkrepo}{\link{https://github.com/sbi-benchmark/sbibm}{github.com/sbi-benchmark/sbibm}}
\newcommand{\linkwebsite}{\link{https://sbi-benchmark.github.io}{sbi-benchmark.github.io}}

\usepackage{titletoc}
\usepackage{hyperref}  % Needs to be loaded after titletoc (!)
\usepackage[ruled,vlined]{algorithm2e}
\usepackage{parnotes}
\usepackage[pdftex]{graphicx}
\usepackage{tabularx}
\usepackage{subcaption}
\usepackage{lipsum}
\usepackage{pifont}

\usepackage{tcolorbox}
\tcbuselibrary{skins}
\newtcolorbox{tbox}[1][]{enhanced, notitle, standard jigsaw, boxrule=0.1mm, top=3mm, bottom=4mm, colframe=black!50!white, colback=black!2, coltitle=black, fonttitle=\bfseries, colbacktitle=black!2, subtitle style={frame hidden, opacityframe=0.}, #1}

\usepackage{float}
\floatstyle{plain}
\newfloat{advice}{h}{lop}
\floatname{advice}{Box}

\usepackage{tikz}
\usetikzlibrary{shapes,arrows,arrows.meta}
\usepackage{adjustbox}

\newcommand{\suppfig}[1]{Suppl. Fig. \ref{fig:#1}}

\newcommand{\algo}[1]{Algorithm \ref{algorithm:#1}}

\usepackage{marginnote}

\newcommand{\bigcell}[2]{\begin{tabular}[t]{@{}#1@{}}#2\end{tabular}}
\newcommand{\newcell}[2]{%
  \expandafter\newcommand\csname rmk-#1\endcsname{#2}%
}
\newcommand{\cell}[1]{\csname rmk-#1\endcsname}

\definecolor{ABC}{RGB}{74, 140, 251}
\definecolor{SL}{RGB}{0, 0, 0}
\definecolor{NLE}{RGB}{96, 208, 152}
\definecolor{NPE}{RGB}{204, 102, 204}
\definecolor{NRE}{RGB}{255, 166, 10}
\definecolor{SABC}{RGB}{33, 95, 198}
\definecolor{RFABC}{RGB}{0, 0, 0}  % 0, 0, 102
\definecolor{SNLE}{RGB}{51, 153, 102}
\definecolor{SNPE}{RGB}{153, 0, 153}
\definecolor{SNRE}{RGB}{255, 125, 22}

\newcommand{\ABC}{{\color{ABC}{REJ-ABC}}}
\newcommand{\SABC}{{\color{SABC}{SMC-ABC}}}
\newcommand{\NLE}{{\color{NLE}{NLE}}}
\newcommand{\SNLE}{{\color{SNLE}{SNLE}}}
\newcommand{\NPE}{{\color{NPE}{NPE}}}
\newcommand{\SNPE}{{\color{SNPE}{SNPE}}}
\newcommand{\NRE}{{\color{NRE}{NRE}}}
\newcommand{\SNRE}{{\color{SNRE}{SNRE}}}
\newcommand{\RFABC}{{\color{RFABC}{RF-ABC}}}
\newcommand{\SL}{{\color{SL}{SL}}}

\newcommand{\XABC}{({\color{SABC}{S}}){\color{ABC}{ABC}}}
\newcommand{\XNLE}{({\color{SNLE}{S}}){\color{NLE}{NLE}}}
\newcommand{\XNPE}{({\color{SNPE}{S}}){\color{NPE}{NPE}}}
\newcommand{\XNRE}{({\color{SNRE}{S}}){\color{NRE}{NRE}}}

%% file: preamble/math.tex
\usepackage{amsmath}
\usepackage{amssymb}
\newcommand{\mathbold}[1]{\ensuremath{\boldsymbol{\mathbf{#1}}}}

\newcommand{\E}{\mathbb{E}}
\newcommand{\KL}{D_{\text{KL}}}

\newcommand{\Uniform}{\mathcal{U}}
\newcommand{\Normal}{\mathcal{N}}

\renewcommand{\d}[1]{\ensuremath{\operatorname{d}\!{#1}}}

\DeclareMathOperator*{\argmin}{arg\,min}

\newcommand{\bff}{\mathbold{f}}

\newcommand{\bmm}{\mathbold{m}}

\newcommand{\bv}{\mathbold{v}}

\newcommand{\bx}{\mathbold{x}}
\newcommand{\by}{\mathbold{y}}
\newcommand{\bz}{\mathbold{z}}

\newcommand{\bF}{\mathbold{F}}

\newcommand{\bI}{\mathbold{I}}

\newcommand{\bS}{\mathbold{S}}

\newcommand{\bV}{\mathbold{V}}

\newcommand{\bepsilon}{\mathbold{\epsilon}}

\newcommand{\bmu}{\mathbold{\mu}}

\newcommand{\bphi}{\mathbold{\phi}}

\newcommand{\bpsi}{\mathbold{\psi}}

\newcommand{\btheta}{\mathbold{\theta}}

\newcommand{\bSigma}{\mathbold{\Sigma}}

\newcommand{\bzero}{\mathbold{0}}
\newcommand{\bone}{\mathbold{1}}

\newcommand{\bthree}{\mathbold{3}}

\newcommand{\bbR}{\mathbb{R}}

\newcommand{\mcD}{\mathcal{D}}

%% file: 00_abstract.tex
\begin{abstract}
Recent advances in probabilistic modelling have led to a large number of simulation-based inference algorithms which do not require numerical evaluation of likelihoods. However, a public benchmark with appropriate performance metrics for such `likelihood-free' algorithms has been lacking. This has made it difficult to compare algorithms and identify their strengths and weaknesses. We set out to fill this gap: We provide a benchmark with inference tasks and suitable performance metrics, with an initial selection of algorithms including recent approaches employing neural networks and classical Approximate Bayesian Computation methods. We found that the choice of performance metric is critical, that even state-of-the-art algorithms have substantial room for improvement, and that sequential estimation improves sample efficiency. Neural network-based approaches generally exhibit better performance, but there is no uniformly best algorithm. We provide practical advice and highlight the potential of the benchmark to diagnose problems and improve algorithms. The results can be explored interactively on a companion website. All code is open source, making it possible to contribute further benchmark tasks and inference algorithms.
\end{abstract}

%% file: 01_introduction.tex
\section{Introduction}

%
% Description of SBI
% 

Many domains of science, engineering, and economics make extensive use of models implemented as stochastic numerical simulators \citep{gourieroux1993indirect,ratmann2007using,alsing2018,brehmer2018constraining,karabatsos2018approximate,gonccalves2019training}. A key challenge when studying and validating such simulation-based models is the statistical identification of parameters which are consistent with observed data. In many cases, calculation of the likelihood is intractable or impractical, rendering conventional approaches inapplicable. The goal of simulation-based inference (SBI), also known as `likelihood-free inference', is to perform Bayesian inference without requiring numerical evaluation of the likelihood function \citep{sisson2018_chapter1,cranmer2019}. In SBI, it is generally not required that the simulator is differentiable, nor that one has access to its internal random variables. 

%
% Progress and issues
%

In recent years, several new SBI algorithms have been developed \citep[e.g., ][]{gutmann2015,papamakarios2016,lueckmann2017,chan2018,greenberg2019,papamakarios2019a,prangle2019distilling,brehmer2020a,hermans2019,jarvenpaa2020,picchini2020,rodrigues2020,thomas2020}, energized, in part, by advances in probabilistic machine learning \citep{rezende2016,papamakarios2017,papamakarios2019c}. Despite---or possibly \emph{because}---of these rapid and exciting developments, it is currently difficult to assess how different approaches relate to each other theoretically and empirically: First, different studies often use different tasks and metrics for comparison, and comprehensive comparisons on multiple tasks and simulation budgets are rare.
Second, some commonly employed metrics might not be appropriate or might be biased through the choice of hyperparameters.
Third, the absence of a benchmark has made it necessary to reimplement tasks and algorithms for each new study. This practice is wasteful, and makes it hard to rapidly evaluate the potential of new algorithms. 
Overall, it is difficult to discern the most promising approaches and decide on which algorithm to use when. These problems are exacerbated by the interdisciplinary nature of research on SBI, which has led to independent development and co-existence of closely-related algorithms in different disciplines.

%
% Potential of having a benchmark for SBI
%

There are many exciting challenges and opportunities ahead, such as the scaling of these algorithms to high-dimensional data, active selection of simulations, and gray-box settings, as outlined in \citet{cranmer2019}. To tackle such challenges, researchers will need an extensible framework to compare existing algorithms and test novel ideas. Carefully curated, a benchmark framework will make it easier for researchers to enter SBI research, and will fuel the development of new algorithms through community involvement, exchange of expertise and collaboration. Furthermore, benchmarking results could help practitioners to decide which algorithm to use on a given problem of interest, and thereby contribute to the dissemination of SBI.

%
% Why should people care Pt. II? Potential of benchmarks in general.
%

The catalyzing effect of benchmarks has been evident, e.g.,~in computer vision 
%\citep{krizhevsky2009CIFAR,deng2009imagenet}, 
\citep{russakovsky2015imagenet}, 
speech recognition \citep{hirsch2000aurora,wang2018glue}, reinforcement learning \citep{bellemare2013arcade, duan2016benchmarking},
Bayesian deep learning \citep{filos2019systematic,wenzel2020}, and many other fields drawing on machine learning. Open benchmarks can be an important component of transparent and reproducible computational research. 
Surprisingly, a benchmark framework for SBI has been lacking, possibly due to the challenging endeavor of designing benchmarking tasks and defining suitable performance metrics.  % took out widely accepted 

% Figure 1 on page 2
\input{figs/01_algorithms}

%
% How do we solve it? (Outlines flow of the paper)
%

Here, we begin to address this challenge, and provide a benchmark framework for SBI to allow rapid and transparent comparisons of current and future SBI algorithms: First, we selected a set of initial algorithms representing distinct approaches to SBI \citep[\autoref{fig:algorithms}; ][]{cranmer2019}. Second, we analyzed multiple performance metrics which have been used in the SBI literature. Third, we implemented ten tasks including tasks popular in the field.
The shortcomings of commonly used metrics led us to focus on tasks for which a likelihood \emph{can} be evaluated, which allowed us to calculate reference (`ground-truth') posteriors. % via MCMC sampling or numerical integration.
These reference posteriors are made available to allow rapid evaluation of SBI algorithms. Code for the framework is available at \linkrepo{} and we maintain an interactive version of all results at \linkwebsite{}. %\footnote{Code submitted as supplementary material. Screenshots and description of website in \autoref{appendix:website}.}

The full potential of the benchmark will be realized when it is populated with additional community-contributed algorithms and tasks. However, our initial version already provides useful insights: 1) the choice of performance metric is critical; 2) the performance of the algorithms on some tasks leaves substantial room for improvement; 3) sequential estimation generally improves sample efficiency; 4) for small and moderate simulation budgets, neural-network based approaches outperform classical ABC algorithms, confirming recent progress in the field; and 5) there is no algorithm to rule them all. The performance ranking of algorithms is task-dependent, pointing to a need for better guidance or automated procedures for choosing  which algorithm to use when. We highlight examples of how the benchmark can be used to diagnose shortcomings of algorithms and facilitate improvements. We end with a discussion of the limitations of the benchmark. 

%% file: figs/01_algorithms.tex
\begin{figure*}[t]
    \centering
    \includegraphics[trim=0 0.25cm 0 0.3cm,clip,width=\textwidth]{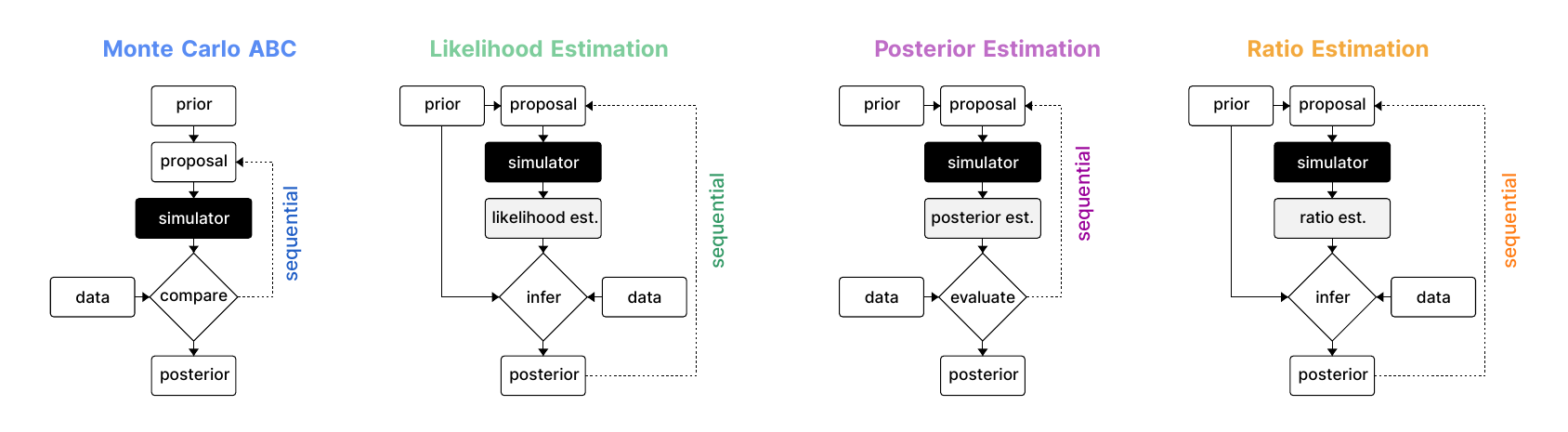}
    \caption{
        {\bf Overview of algorithms.} We compare algorithms belonging to four distinct approaches to SBI: Classical ABC approaches as well as model-based approaches approximating likelihoods, posteriors, or density ratios. We contrast algorithms that use the prior distribution to propose parameters against ones that sequentially adapt the proposal. Classification and schemes following \cite{cranmer2019}. % 
    }
    \label{fig:algorithms}
\end{figure*}

%% file: 02_benchmark.tex
\section{Benchmark}

The benchmark consists of a set of algorithms, performance metrics and tasks. Given a prior $p(\btheta)$ over parameters $\btheta$, a simulator to sample $\bx \sim p(\bx|\btheta)$ and an observation $\bx_o$, an algorithm returns an approximate posterior $q(\btheta|\bx_o)$, or samples from it, $\btheta \sim q$. The approximate solution is tested, according to a performance metric, against a reference posterior $p(\btheta|\bx_o)$.

\subsection{Algorithms}

Following the classification introduced in the review by \citet{cranmer2019}, we selected algorithms addressing SBI in four distinct ways, as schematically depicted in \autoref{fig:algorithms}. An important difference between algorithms is how new simulations are acquired: Sequential algorithms adaptively choose informative simulations to increase sample efficiency. While crucial for expensive simulators, it can require non-trivial algorithmic steps and hyperparameter choices. To evaluate whether the potential is realized empirically and justifies the algorithmic burden, we included sequential and non-sequential counterparts for algorithms of each category. 

Keeping our initial selection focused allowed us to carefully consider implementation details and hyperparameters: We extensively explored performance and sensitivity to different choices in more than 10k runs, all results and details of which can be found in \autoref{appendix:hyperparams}. Our selection is briefly described below, full algorithm details are in \autoref{appendix:algorithms}.

\textbf{\ABC{} and \SABC{}.} Approximate Bayesian Computation \citep[ABC,][]{sisson2018_chapter1} is centered around the idea of Monte Carlo rejection sampling \citep{tavere1997,pritchard1999}. Parameters $\btheta$ are sampled from a proposal distribution, simulation outcomes $\bx$ are compared with observed data $\bx_o$, and are accepted or rejected depending on a (user-specified) distance function and rejection criterion. While rejection ABC (\ABC{}) uses the prior as a proposal distribution, the efficiency can be improved by using sequentially refined proposal distributions \citep[\SABC{},][]{beaumont2002, marjoram2006, sisson2007, toni2009, beaumont2009}. We implemented \ABC{} with quantile-based rejection and used the scheme of \citet{beaumont2009} for \SABC{}. We extensively varied hyperparameters and compared the implementation of an ABC-toolbox \citep{klinger2018} against our own (\autoref{appendix:hyperparams}). We investigated linear regression adjustment \citep{blum2010} and the summary statistics approach by \citet{prangle2014semi} (\suppfig{abc_additional}).

\textbf{\NLE{} and \SNLE{}.} Likelihood estimation (or `synthetic likelihood') algorithms learn an approximation to the intractable likelihood, for an overview see \citet{sisson2018_chapter12}. While early incarnations focused on Gaussian approximations \citep[\SL{}; ][]{wood2010}, recent versions utilize deep neural networks \citep{papamakarios2019a,lueckmann2019} to approximate a density over $\bx$, followed by MCMC to obtain posterior samples. Since we primarily focused on these latter versions, we refer to them as neural likelihood estimation (\NLE{}) algorithms, and denote the sequential variant with proposals as \SNLE{}. In particular, we used the scheme proposed by \citet{papamakarios2019a} which uses masked autoregressive flows \citep[MAFs, ][]{papamakarios2017} for density estimation. We improved MCMC sampling for \XNLE{} and compared MAFs against Neural Spline Flows \citep[NSFs; ][]{durkan2019neural}, see \autoref{appendix:hyperparams}.

\textbf{\NPE{} and \SNPE{}.} Instead of approximating the likelihood, these approaches directly target the posterior. Their origins date back to regression adjustment approaches \citep{blum2010}. Modern variants \citep{papamakarios2016,lueckmann2017,greenberg2019} use neural networks for density estimation (approximating a density over $\btheta$). Here, we used the recent algorithmic approach proposed by \citet{greenberg2019} for sequential acquisitions. We report performance using NSFs for density estimation, which outperformed MAFs (\autoref{appendix:hyperparams}).

\textbf{\NRE{} and \SNRE{}.} Ratio Estimation approaches to SBI use classifiers to approximate density ratios \citep{izbicki2014high,pham2014note,cranmer2015,dutta2016likelihood,durkan2020,thomas2020}.
Here, we used the recent approach proposed by \citet{hermans2019} as implemented in \citet{durkan2020}: A neural network-based classifier approximates probability ratios and MCMC is used to obtain samples from the posterior. \SNRE{} denotes the sequential variant of neural ratio estimation (\NRE{}). In \autoref{appendix:hyperparams} we compare different classifier architectures for \XNRE{}.

In addition, we benchmarked Random Forest ABC \citep[\RFABC{}; ][]{raynal2018}, a recent ABC variant, and Synthetic Likelihood \citep[SL; ][]{wood2010}, mentioned above. However, \RFABC{} only targets individual parameters (i.e.~assumes posteriors to factorize), and \SL{} requires new simulations for every MCMC step, thus requiring orders of magnitude more simulations than other algorithms. Therefore, we report results for these algorithms separately, in \suppfig{rf_abc} and \suppfig{sl}, respectively. 

Algorithms can be grouped with respect to how their output is represented: 1) some return samples from the posterior, $\btheta \sim q(\btheta|\bx_o)$ (\ABC{}, \SABC{}); 2) others return samples and allow evaluation of unnormalized posteriors $\tilde{q}(\btheta|\bx_o)$ (\XNLE{}, \XNRE{}); and 3) for some, the posterior density $q(\btheta|\bx_o)$ can be evaluated and sampled directly, without MCMC (\XNPE{}). As discussed below, these properties constrain the metrics that can be used for comparison.

\subsection{Performance metrics}

Choice of a suitable performance metric is central to any benchmark. As the goal of SBI algorithms is to perform full inference, the `gold standard' would be to quantify the similarity between the true posterior and the inferred one with a suitable distance (or divergence) measure on probability distributions. This would require both access to the ground-truth posterior, and a reliable means of estimating similarity between (potentially) richly structured distributions. 
Several performance metrics have been used in past research, depending on the constraints imposed by knowledge about ground-truth and the inference algorithm (see \autoref{table:metrics}). In real-world applications, typically only the observation $\bx_o$ is known. However, in a benchmarking setting, it is reasonable to assume that one has at least access to the ground-truth parameters $\btheta_o$. There are two commonly used metrics which only require $\btheta_o$ and $\bx_o$, but suffer severe drawbacks for our purposes:

\textbf{Probability $\btheta_o$}. The negative log probability of true parameters averaged over different $(\btheta_o, \bx_o)$, $-\E[\log q(\btheta_o|\bx_o)]$, has been used extensively in the literature \citep{papamakarios2016,durkan2018,greenberg2019,papamakarios2019a,durkan2020,hermans2019}. Its appeal lies in the fact that one does not need access to the ground-truth posterior. However, using it only for a small set of $(\btheta_o, \bx_o)$ is highly problematic: It is only a valid performance measure if averaged over a large set of observations sampled from the prior \citep[][detailed discussion including connection to simulation-based calibration in \autoref{appendix:metrics}]{talts2018}. For reliable results, one would require inference for hundreds of $\bx_o$ which is only feasible if inference is rapid (amortized) and the density $q$ can be evaluated directly (among the algorithms used here this applies only to \NPE{}).

% Table 1 on page 4
\input{tables/01_metrics}
\textbf{Posterior-Predictive Checks (PPCs).} As the name implies, PPCs should be considered a mere check rather than a metric, although the \textit{median distance} between predictive samples and $\bx_o$ has been reported in the SBI literature \citep{papamakarios2019a,greenberg2019,durkan2020}. A failure mode of such a metric is that an algorithm obtaining a good MAP point estimate, could perfectly pass this check even if the estimated posterior is poor. Empirically, we found median-distances (MEDDIST) to be in disagreement with other metrics (see \nameref{results}).

The shortcomings of these commonly-used metrics led us to focus on tasks for which it is possible to get samples from ground-truth posterior $\btheta \sim p$, thus allowing us to use metrics based on two-sample tests:

\textbf{Maximum Mean Discrepancy (MMD).} MMD \citep{gretton2012,sutherland2017} is a kernel-based 2-sample test. Recent papers \citep{papamakarios2019a,greenberg2019,hermans2019} reported MMD using translation-invariant Gaussian kernels with length scales determined by the median heuristic \citep{ramdas2015}. We empirically found that MMD can be sensitive to hyperparameter choices, in particular on posteriors with multiple modes and length scales \citep[see \nameref{results} and][]{liu2020}. 

\textbf{Classifier 2-Sample Tests (C2ST).} C2STs \citep{friedman2004, lopez-paz2018} train a classifier to discriminate samples from the true and inferred posteriors, which makes them simple to apply and easy to interpret. Therefore, we prefer to report and compare algorithms in terms of accuracy in classification-based tests. In the context of SBI, C2ST has e.g.~been used in \citet{gutmann2018likelihood,dalmasso2019a}.

Other metrics that could be used include:

\textbf{Kernelized Stein Discrepancy (KSD).} KSD \citep{liu2016,chwialkowski2016} is a 1-sample test, which require access to $\nabla_{\btheta}\ \tilde{p}(\btheta|\bx_o)$ rather than samples from $p$ ($\tilde{p}$ is the unnormalized posterior). Like MMD, current estimators use translation-invariant kernels. 

\textbf{$f$-Divergences.} Divergences such as Total Variation (TV) divergence and KL divergences can only be computed when the densities of true and approximate posteriors can be evaluated (\autoref{table:metrics}). Thus, we did not use $f$-divergences for the benchmark.

Full discussion and details of metrics in  \autoref{appendix:metrics}.

\subsection{Tasks}

The preceding considerations guided our selection of inference tasks: We focused on tasks for which reference posterior samples $\btheta \sim p$ can be obtained, to allow calculation of 2-sample tests. We focused on eight purely statistical problems and two problems relevant in applied domains, with diverse dimensionalities of parameters and data (details in \autoref{appendix:tasks}):

\textbf{Gaussian Linear/Gaussian Linear Uniform.} We included two versions of simple, linear, 10-d Gaussian models, in which the parameter $\btheta$ is the mean, and the covariance is fixed. The first version has a Gaussian (conjugate) prior, the second one a uniform prior. These tasks allow us to test how algorithms deal with trivial scaling of dimensionality, as well as truncated support.

\textbf{SLCP/SLCP Distractors.} A challenging inference task designed to have a simple likelihood and a complex posterior \citep{papamakarios2019a, greenberg2019}: The prior is uniform over five parameters $\btheta$ and the data are a set of four two-dimensional points sampled from a Gaussian likelihood whose mean and variance are nonlinear functions of $\btheta$. This induces a complex posterior with four symmetrical modes and vertical cut-offs. We included a second version with 92 additional, non-informative outputs (distractors) to test the ability to detect informative features. 

\textbf{Bernoulli GLM/Bernoulli GLM Raw.} 10-parameter Generalized Linear Model (GLM) with Bernoulli observations. Inference was either performed on sufficient statistics (10-d) or raw data (100-d).

\textbf{Gaussian Mixture.} This inference task, introduced by \citet{sisson2007}, has become common in the ABC literature \citep{beaumont2009, toni2009, simola2020}. It consists of a mixture of two two-dimensional Gaussian distributions, one with much broader covariance than the other. 

\textbf{Two Moons.} A two-dimensional task with a posterior that exhibits both global (bimodality) and local (crescent shape) structure \citep{greenberg2019} to illustrate how algorithms deal with multimodality.

\textbf{SIR.} Dynamical systems represent paradigmatic use cases for SBI. SIR is an influential epidemiological model describing the dynamics of the number of individuals in three possible states: susceptible $S$, infectious $I$, and recovered or deceased, $R$. We infer the contact rate $\beta$ and the mean recovery rate $\gamma$, given observed infection counts $I$ at 10 evenly-spaced time points.

\textbf{Lotka-Volterra.} An influential model in ecology describing the dynamics of two interacting species, widely used in SBI studies. We infer four parameters $\btheta$ related to species interaction, given the number of individuals in both populations at 10 evenly-spaced points in time.

% Figure 2
\input{figs/02_two_moons_metrics}

\subsection{Experimental Setup}

For each task, we sampled 10 sets of true parameters from the prior and generated corresponding observations $(\btheta_o, \bx_o)_{1 {:}10}$. For each observation, we generated 10k samples from the reference posterior. Some reference posteriors required a customised (likelihood-based) approach (\autoref{appendix:benchmark}).

In SBI, it is typically assumed that total computation cost is dominated by simulation time.
We therefore report performance at different simulation budgets. For each observation, each algorithm was run with a simulation budget ranging from 1k to 100k simulations.

For each run, we calculated metrics described above. To estimate C2ST accuracy, we trained a multilayer perceptron to tell apart approximate and reference posterior samples and performed five-fold cross-validation. We used two hidden layers, each with 10 times as many ReLu units as the dimensionality of the data.
We also measured and report runtimes (\autoref{appendix:runtimes}).

\subsection{Software}

\textbf{Code.} All code is released publicly at \linkrepo{}. Our framework includes tasks, reference posteriors, metrics, plotting, and infrastructure tooling and is designed to be 1) easily extensible, 2) used with external toolboxes implementing algorithms. All tasks are implemented as probabilistic programs in \texttt{Pyro} \citep{bingham2018}, so that likelihoods and gradients for reference posteriors can be extracted automatically. To make this possible for tasks that use ODEs, we developed a new interface between \texttt{DifferentialEquations.jl} \citep{rackauckas2017,bezanson2017julia} and \texttt{PyTorch} \citep{paszke2019}. In addition, specifying simulators in a probabilistic programming language has the advantage that `gray-box' algorithms \citep{brehmer2020a,cranmer2019} can be added in the future. We here evaluated algorithms implemented in \texttt{pyABC} \citep{klinger2018}, \texttt{pyabcranger} \citep{collin2020}, and \texttt{sbi} \citep{sbi}. See \autoref{appendix:benchmark} for details and existing SBI toolboxes.

\textbf{Reproducibility.} Instructions to reproduce experiments on cloud-based infrastructure are in \autoref{appendix:benchmark}.

\textbf{Website.} Along with the code, we provide a web interface which allows interactive exploration of all the results (\linkwebsite{}; \autoref{appendix:website}). %Here, we include a small selection of figures. 

% Figure 3
\input{figs/03_results_metrics}

%% file: tables/01_metrics.tex
%
% Algorithm categories
%
\newcell{a0}{\multicolumn{1}{l}{$\downarrow$~ \textbf{Algorithm}}}
\newcell{a1}{\multicolumn{1}{l}{$\btheta \sim q$}}            % Returns samples
\newcell{a2}{\multicolumn{1}{l}{$\tilde{q}(\btheta|\bx_o)$}}  % Returns unnormalized density
\newcell{a3}{\multicolumn{1}{l}{$q(\btheta|\bx_o)$}}          % Returns normalized density

%
% Task categories
%
\newcell{t0}{\multicolumn{5}{l}{\textbf{Ground truth} $\rightarrow$}}
\newcell{t1}{\multicolumn{1}{l}{$\bx_o$}}                      % Only observed data known
\newcell{t2}{\multicolumn{1}{l}{$\btheta_o$}}         % Only ground-truth parameter known
\newcell{t3}{\multicolumn{1}{l}{$\btheta \sim p$}}             % Samples from posterior available
\newcell{t4}{\multicolumn{1}{l}{$\nabla \tilde{p}(\btheta|\bx_o)$}}   % Posterior up to normalizing constant
\newcell{t5}{\multicolumn{1}{l}{$p(\btheta|\bx_o)$}}           % Closed-form posterior

%
% Metrics
%
\newcell{m0}{$\downarrow$~ \textbf{Metric}}
\newcell{m1}{{$f$-divergences}\\TV, KL}
\newcell{m2}{{1-sample tests}\\KSD}
\newcell{m3}{{2-sample tests}\\C2ST, MMD}
\newcell{m4}{{Probability} $\btheta_o$\\$-\E[\log q(\btheta_o|\bx_o)]$}
\newcell{m5}{{PPCs}\\Median distance}  % $|\bx-\bx_o|$

%
% Additional column types
%
\newcolumntype{b}{X}
\newcolumntype{s}{>{\hsize=0.4\hsize}X}

%
% Table
%
\begin{table*}[t]
\caption{\textbf{Applicability of metrics given knowledge about ground truth and algorithm.} Whether a metric can be used depends on \emph{both} what is known about the ground-truth of an inference task and what an algorithm returns: Information about ground truth can vary between just having observed data $\bx_o$ (typical setting in practice), knowing the generating parameter $\btheta_o$, having posterior samples, gradients, or being able to evaluate the true posterior $p$. Tilde denotes unnormalized distributions. Access to information is cumulative. 
\vspace{0.2cm}
}
\begin{tabularx}{\textwidth}{@{}sssssssss@{}}
%\toprule
           & \cell{t0} \\ \cmidrule[0.8pt](l){2-6}
\cell{a0}  & \cell{t1} & \cell{t2} & \cell{t3} & \cell{t4} & \cell{t5} \\ \midrule[0.9pt]
\cell{a1}  & 1 & 1 & 1, 3 & 1, 3, 4 & 1, 3, 4 \\ \midrule[0.5pt]
\cell{a2}  & 1 & 1 & 1, 3 & 1, 3, 4 & 1, 3, 4 \\ \midrule[0.5pt]
\cell{a3}  & 1 & 1, 2 & 1, 2, 3 & 1, 2, 3, 4 & 1, 2, 3, 4, 5 \\ \midrule[0.9pt]
\end{tabularx}
\label{table:metrics}
\vspace{0.5cm}
1 = PPCs, 2 = Probability $\btheta_0$, 3 = 2-sample tests, 4 = 1-sample tests, 5 = $f$-divergences.
\vspace{-0.2cm}
\end{table*}

%% file: figs/02_two_moons_metrics.tex
\begin{figure*}[t!]
    \centering
    \begin{subfigure}[htbp]{\textwidth}
        \includegraphics[trim=20 49 0 5,clip,width=\textwidth]{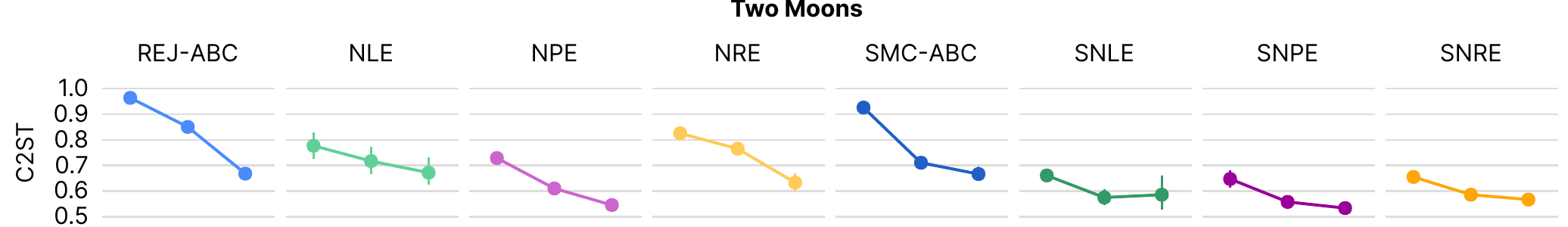}
    \end{subfigure}
    \begin{subfigure}[htbp]{\textwidth}
        \includegraphics[trim=20 50 0 17,clip,width=\textwidth]{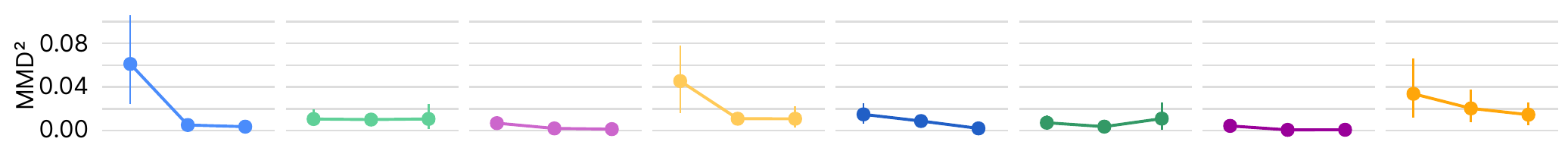}
    \end{subfigure}
    \begin{subfigure}[htbp]{\textwidth}    
        \includegraphics[trim=20 19 0 17,clip,width=\textwidth]{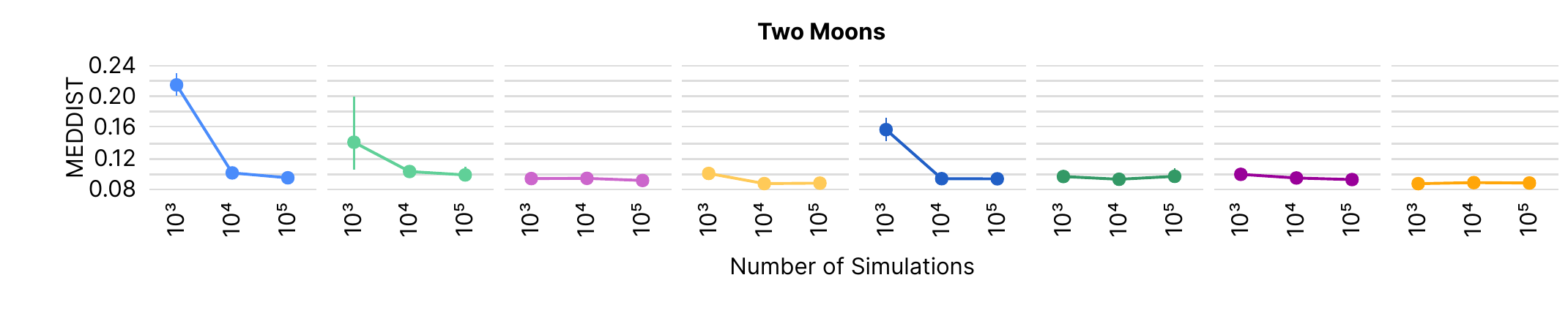}  
    \end{subfigure}
    \caption{
        {\bf Performance on Two Moons according to various metrics.} Best possible performance would be 0.5 for C2ST, 0 for MMD$^2$ and MEDDIST. Results for 10 observations each, means and 95\% confidence intervals. 
    }
    \label{fig:two_moons_metrics}
\end{figure*}

%% file: figs/03_results_metrics.tex
\begin{figure*}[t!]
    \centering
    \begin{subfigure}[htbp]{\textwidth}
        \includegraphics[trim=20 49 0 4,clip,width=\textwidth]{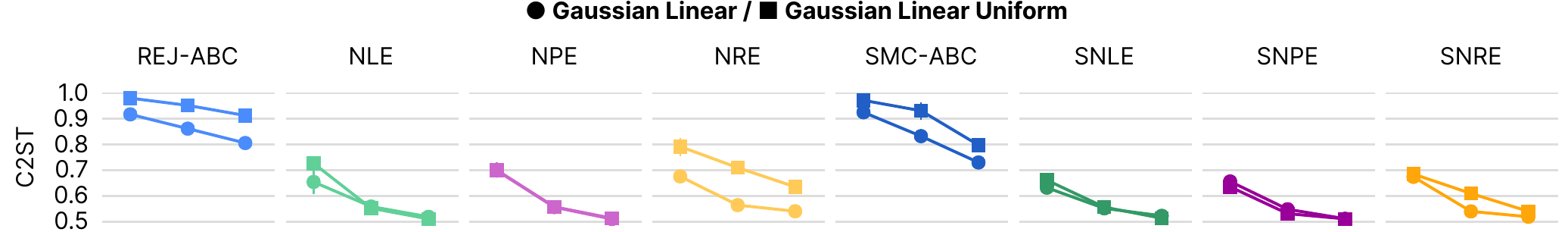}
    \end{subfigure}
    \begin{subfigure}[htbp]{\textwidth}
        \includegraphics[trim=20 49 0 4,clip,width=\textwidth]{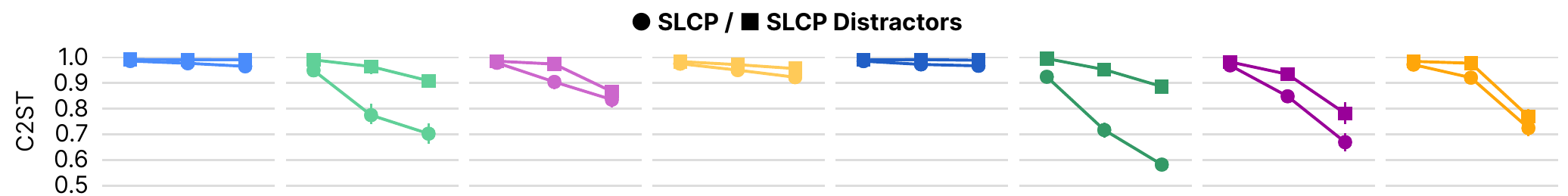}
    \end{subfigure}    
    \begin{subfigure}[htbp]{\textwidth}    
        \includegraphics[trim=20 49 0 4,clip,width=\textwidth]{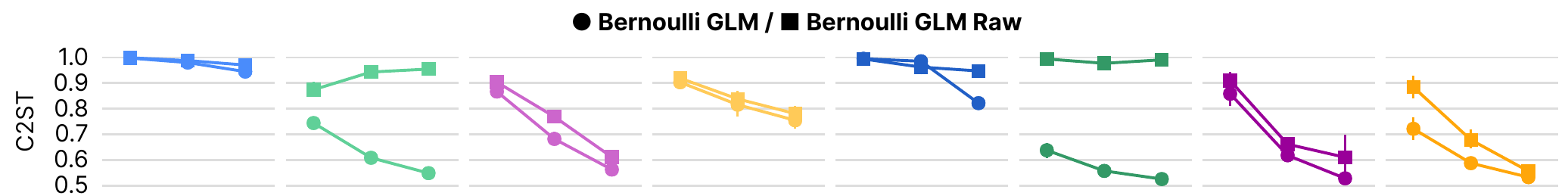}  
    \end{subfigure}
    \begin{subfigure}[htbp]{\textwidth}
        \includegraphics[trim=20 49 0 4,clip,width=\textwidth]{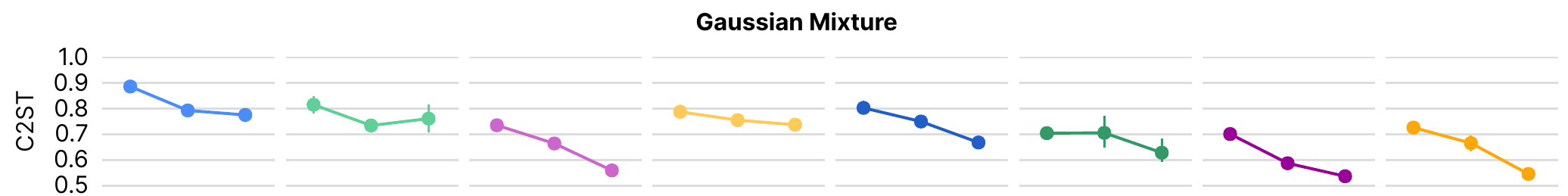}
    \end{subfigure}
    \begin{subfigure}[htbp]{\textwidth}    
        \includegraphics[trim=20 49 0 4,clip,width=\textwidth]{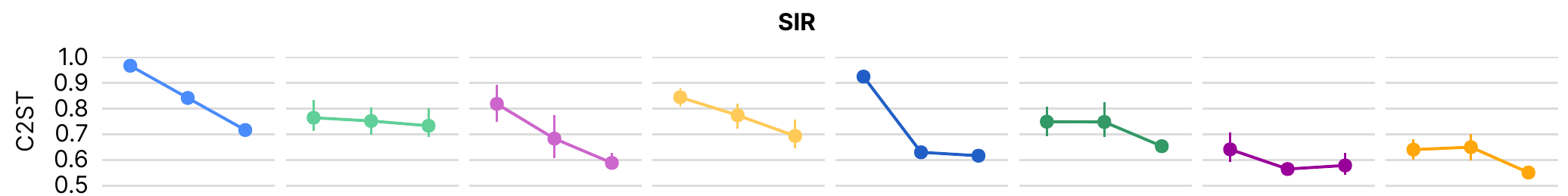}  
    \end{subfigure}
    \begin{subfigure}[htbp]{\textwidth}    
        \includegraphics[trim=20 19 0 4,clip,width=\textwidth]{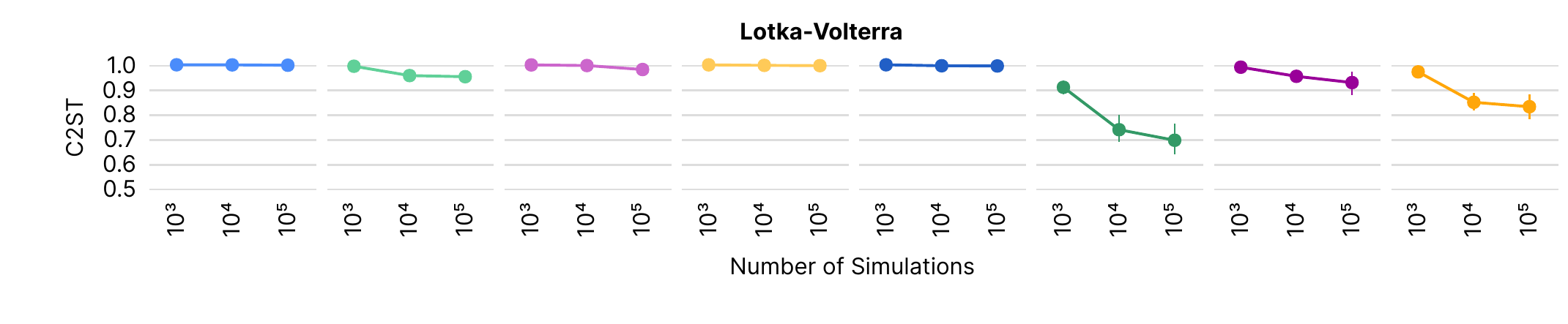}  
    \end{subfigure}
    \caption{
        {\bf Performance on other benchmark tasks.} Classification accuracy (C2ST) of \ABC{}, \SABC{}, \NLE{}, \SNLE{}, \NPE{}, \SNPE{}, \NRE{}, \SNRE{} for 10 observations each, means and 95\% confidence intervals. 
    }
    \label{fig:results_metrics}
\end{figure*}

%% file: 03_results.tex
\section{Results}
\label{results}

We first consider empirical results on a single task, Two Moons, according to different metrics, which illustrate the following important insight:

\textbf{\#1: Choice of performance metric is key.} While C2ST results on Two Moons show that performance increases with higher simulation budgets and that sequential algorithms outperform non-sequential ones for low to medium budgets, these results were not reflected in MMD and MEDDIST (\autoref{fig:two_moons_metrics}): In our analyses, we found MMD to be sensitive to hyperparameter choices, in particular on tasks with complex posterior structure. When using the commonly employed median heuristic to set the kernel length scale on a task with multi-modal posteriors (like Two Moons), MMD had difficulty discerning markedly different posteriors. This can be `fixed' by using hyperparameters adapted to the task (\suppfig{mmd}). As discussed above, the median distance (though commonly used) can be `gamed' by a good point estimate even if the estimated posterior is poor and is thus not a suitable performance metric. Computation of KSD showed numerical problems on Two Moons, due to the gradient calculation.

We assessed relationships between metrics empirically via the correlations across tasks (\suppfig{metrics_correlations}). As discussed above, the log-probability of ground-truth parameters can be problematic when averaged over too few observations (e.g.,~10, as is common in the literature): indeed, this metric had a correlation of only 0.3 with C2ST on Two Moons and 0.6 on the SLCP task. Based on these considerations, we used C2ST for reporting performance (\autoref{fig:results_metrics}; results for MMD, KSD and median distance on the website). 

Based on the comparison of the performance across all tasks, we highlight the following main points:

\textbf{\#2: These are not solved problems.} C2ST uses an interpretable scale (1 to 0.5), which makes it possible to conclude that, for several tasks, no algorithm could solve them with the specified budget (e.g.,~SLCP, Lotka-Volterra). This highlights that our problems---though conceptually simple---are challenging, and there is room for development of more powerful algorithms. 

\textbf{\#3: Sequential estimation improves sample efficiency.} Our results show that sequential algorithms outperform non-sequential ones (\autoref{fig:results_metrics}). The difference was small on simple tasks (i.e.~linear Gaussian cases), yet pronounced on most others. However, we also found these methods to exhibit diminishing returns as the simulation budget grows, which points to an opportunity for future improvements.

\textbf{\#4: Density or ratio estimation-based algorithms generally outperform classical techniques.} \ABC{} and \SABC{} were generally outperformed by more recent techniques which use neural networks for density- or ratio-estimation, and which can therefore efficiently interpolate between different simulations (\autoref{fig:results_metrics}). 
Without such model-based interpolation, even a simple 10-d Gaussian task can be challenging. However, classical rejection-based methods have a computational footprint that is orders of magnitudes smaller, as no network training is involved (\autoref{appendix:runtimes}). Thus, on low-dimensional problems and for cheap simulators, these methods can still be competitive. See \suppfig{abc_additional} for results with additional ABC variants \citep{blum2010,prangle2014semi} and \suppfig{rf_abc} for results on \RFABC{}.

\pagebreak

\textbf{\#5: No one algorithm to rule them all.} Although sequential density or ratio estimation-based algorithms performed better than their non-sequential counterparts, there was no clear-cut answer as to which sequential method (\SNLE{}, \SNRE{}, and \SNPE{}) should be preferred. To some degree, this is to be expected: these algorithms have distinct strengths that can play out differently depending on the problem structure \citep[see discussions e.g.,~in][]{greenberg2019, durkan2018, durkan2020}. However, this has not been shown systematically before.
We formulate some practical guidelines for choosing appropriate algorithms in \hyperref[box:advice]{Box~\ref*{box:advice}}.

% Figure 4 (proposing to take it out for space)
% \input{figs/04_slcp_posteriors}

\textbf{\#6: The benchmark can be used to diagnose implementation issues and improve algorithms.} 
For example, \XNLE{} and \XNRE{} rely on MCMC sampling to compute posteriors, and this sampling step can limit the performance. Access to a reference posterior can help identify and improve such issues: We found that single chains initialized by sampling from the prior with axis-aligned slice sampling \citep[as introduced in][]{papamakarios2019a} frequently got stuck in single modes. Based on this observation, we changed the MCMC strategy (details in \autoref{appendix:algorithms}), which, though simple, yielded significant performance and speed improvements on the benchmark tasks. Similarly, \XNLE{} and \XNRE{} improved by transforming parameters to be unbounded: Without transformations, runs on some tasks can get stuck during MCMC sampling (e.g.,~Lotka-Volterra). While this is common advice for MCMC \citep{hogg2017}, it has been lacking in code and papers of SBI approaches.

We used the benchmark to systematically compare hyperparameters: For example, as density estimators 

\input{04_practical_advice}

for \XNLE{} and \XNPE{}, we used NSFs \citep{durkan2020} which were developed after these algorithms were published. This revealed that higher capacity density estimators were beneficial for posterior but not likelihood estimation (detailed analysis in \autoref{appendix:hyperparams}).

These examples show how the benchmark makes it possible to diagnose problems and improve algorithms.

%% file: 04_practical_advice.tex
\onecolumn

\begin{advice}[h!]

%\centering
\begin{tbox}

\tcbsubtitle[]{Do we need the Bayesian posterior, or is a point estimate sufficient?}

Our focus was on SBI algorithms that target the Bayesian posterior. If one only aims for a single estimate, optimization methods \citep[e.g.][]{rios2013derivative,shahriari2015taking} might be more efficient. 

\tcbsubtitle[]{Is the simulator really `black-box'?}

The SBI algorithms presented in the benchmark can be applied to any `black-box' simulator.
% (no access to likelihoods or internal number generators, no gradients required). If additional information is available, it is worth using algorithms that can leverage it. For example, 
However, if the likelihood is available, methods exploiting it (e.g.~MCMC, variational inference) will generally be more efficient. Similarly, if one has access to the internal random numbers, probabilistic programming approaches \citep{le2016inference, baydin2019etalumis, wood2020} might be preferable. If additional quantities that characterize the latent process are available, i.e., the simulator is `gray-box', they can be used to augment training data and improve inference \citep{brehmer2020a, cranmer2019}.

\tcbsubtitle[]{What domain knowledge do we have about the problem?}

For any practical application of SBI, it is worth thinking carefully about domain knowledge. First, knowledge about plausible parameters should inform the choice of the prior. Second, domain knowledge can help design appropriate distance functions or summary statistics required for classical ABC algorithms. 
%When using model-based approaches, domain knowledge can help to incorporate inductive biases and known invariances into the neural network architecture.
When using model-based approaches, domain knowledge can potentially be built into the SBI algorithm itself, for example, by incorporating neural network layers with appropriate inductive biases or invariances.

%, or hard-coding known invariances into the architecture. 
% Finally, domain knowledge may also guide the design of summary statistics.
%
% \citep[e.g.~convolutional structure of images of rock-paper-scissors model in][]{greenberg2019}
% \citep[e.g.~spike-shape statistics for Hodgkin-Huxley model in][]{lueckmann2017}.

\tcbsubtitle[]{Do we have, or can we learn summary statistics?}

Summary statistics are especially important when facing problems with high-dimensional data: It is important to point out that the posterior given summary statistics $p(\btheta|s(\bx_o))$ is only equivalent to $p(\btheta|\bx_o)$ if the summary statistics are sufficient. The problem at hand can guide the manual design of summary statistics that are regarded particularly important or informative. Alternatively, many automatic approaches exist \citep[e.g.,][]{prangle2014semi,charnock2018,dinev2018} and this is an active area of research (e.g., \citealt{chen2020neural} recently proposed an approach to learn approximately sufficient statistics for \SABC{} and \XNLE{}). \XNPE{} and \XNRE{} can directly reduce high-dimensional data as part of their network architectures.

% and discuss that many existing approaches do not guarantee (global) sufficiency for all $\btheta$. 
%When using \XNPE{} or \XNRE{},
%high-dimensional data can be reduced as part of the network architecture. 

\tcbsubtitle[]{Do we have low-dimensional data and parameters, and a cheap simulator?}

If both the parameters and the data (or suitable summary-statistics thereof) are low-dimensional, and a very large number of simulations can be generated, model-free algorithms such as classical ABC can be competitive. These have the benefit of adding little computational overhead. Conversely, for limited simulation budgets and/or higher dimensionalities, approaches that train a model of the likelihood, posterior, or likelihood ratio will generally be preferable.

%, especially compared to neural network-based approaches.
%
%\tcbsubtitle[]{Are the data or the parameters higher dimensional?}
%
%For limited simulation budget, and/or higher dimensionalities, model-based approaches will generally be preferable. A guideline 
%
%Overall, we did not find strong differences between the these three model-based approaches in terms of estimation-accuracy across the current benchmark tasks. 
%For tasks in which the data dimensionality was much higher than the dimensionality of parameters, we found algorithms that can automatically learn informative summary statistics of the data, e.g., \XNPE{} or \XNRE{} to be more efficient. Conversely, if the data are of lower dimensionality, likelihood-targeting methods (e.g.~\XNLE{}) might perform better. 

\tcbsubtitle[]{Are simulations expensive? Can we simulate online?}

For time-intensive and complex simulators, it can be beneficial to use \textit{sequential} methods to increase sample efficiency: We found that sequential schemes generally outperformed non-sequential ones. While we focused on simple strategies which use the previous estimate of the posterior to propose new parameters, more sophisticated schemes \citep[e.g.,][]{gutmann2015,lueckmann2019, jarvenpaa2019} may increase sample efficiency if only few simulations can be obtained.
For some applications, inference is performed on a fixed dataset, and one cannot resort to sequential algorithms.

\tcbsubtitle[]{Do we want to carry out inference once, or repeatedly?}

To perform SBI \emph{separately} for different data points (i.e.~compute $p(\btheta|\bx_1), p(\btheta|\bx_2), \ldots)$, methods that allow `amortization' (\NPE{}) are likely preferable. While \NLE{} and \NRE{} allow amortisation of the neural network, MCMC sampling is required, which takes additional time. Conversely, if we want to run SBI conditioned on many i.i.d.~data (e.g. $p(\btheta|\bx_1,\bx_2,\ldots)$) methods based on likelihood or ratio estimation (\NLE, \NRE{}), or \NPE{} with exchangeable neural networks \citep{chan2018} would be appropriate.

\end{tbox}

\caption{\textbf{Practitioners' advice for applying SBI algorithms.} Based on our current results and understanding, we provide advice to practitioners seeking to apply SBI. 
% Note that our benchmark contains an initial selection out of a wide variety of SBI algorithms, see e.g.~the recent review by \citet{cranmer2019} for an overview and additional advice. 
There is no one-fits-all solution---which algorithm to use in practice will depend on the problem at hand. For additional advice, see \citet{cranmer2019}.
% With this advice, we seek to provide a few useful questions to help identify which classes of algorithms may be applicable.
} 
\label{box:advice}
\end{advice}

\twocolumn

%% file: 05_discussion.tex
\section{Limitations}
\label{limitations}

Our benchmark, in its current form, has several limitations. First, the algorithms considered here do not cover the entire spectrum of SBI algorithms: We did not include sequential algorithms using active learning or Bayesian Optimization \citep{gutmann2015,jarvenpaa2019,lueckmann2019,aushev2020}, or `gray-box' algorithms, which use additional information about or from the simulator \citep[e.g.,][]{baydin2019etalumis,brehmer2020a}. We focused on approaches using neural networks for density estimation and did not compare to alternatives using Gaussian Processes \citep[e.g.,][]{meeds2014,wilkinson2014}. There are many other algorithms which the benchmark is currently lacking \citep[e.g.,][]{nott2014,ong2018,clarte2019, prangle2019distilling, priddle2019, picchini2020, radev2020bayesflow, rodrigues2020}. Keeping our initial selection small allowed us to carefully investigate hyperparameter choices. We focused on sequential algorithms with less sophisticated acquisition schemes and the black-box scenario, since we think these are important baselines for future comparisons.

%By formulating tasks as probabilistic programs, however, we open up the possibility for these algorithms to be added. 

Second, the tasks we considered do not cover the variety of possible challenges. Notably, while we have tasks with high dimensional data with and without structure, we have not included tasks with high-dimensional spatial structure, e.g.,~images. Such tasks would require algorithms that automatically learn summary statistics while exploring the structure of the data \citep[e.g., ][]{dinev2018,greenberg2019,hermans2019,chen2020neural}, an active research area.

Third, while we extensively investigated tuning choices and compared implementations, the results might nevertheless reflect our own areas of expertise. 

Fourth, in line with common practice in SBI, results presented in the paper focused on performance as a function of the number of simulation calls. It is important to remember that differences in computation time can be substantial (see \autoref{appendix:runtimes}): For example, \XABC{} was much faster than approaches requiring network training. Overall, sequential neural algorithms exhibited longest runtimes.

Fifth, for reasons described above, we focused on problems for which reference posteriors can be computed. This raises the question of how insights on these problems will generalize to `real-world' simulators. Notably, even these simple problems already identify clear differences between, and limitations of, different SBI approaches. Since it is not possible to rigorously compare the performance of different algorithms directly on `real-world' simulators due to the lack of appropriate metrics, we see the benchmark as a necessary stepping stone towards the development of (potentially automated) selection strategies for practical problems.

Sixth, in practice, the choice of algorithm can depend on aspects that are difficult to quantify: It will depend on the available information about a problem, the inference goal, and the speed of the simulator, among other considerations. We included some practical considerations and recommendations in \hyperref[box:advice]{Box~\ref*{box:advice}}.

Finally, benchmarking is an important tool, but not an end in itself---for example, conceptually new ideas might initially not yield competitive results but only reveal their true value later. Conversely, `overfitting' on benchmarks can lead to the illusion of progress, and result in an undue focus on small implementation details which might not generalize beyond it. It would certainly be possible to cheat on this benchmark: In particular, as the simulators are available, one could use samples (or even likelihoods) to excessively tune hyperparameters \textit{for each task}. This would hardly transfer to practice where such tuning is usually impossible (lack of metrics and expensive simulators). %
Therefore, we carefully compared choices and selected hyperparameters performing best \textit{across tasks} (\autoref{appendix:hyperparams}).

\section{Discussion}
\label{discussion}

Quantitatively evaluating, comparing and improving algorithms through benchmarking is at the core of progress in machine learning. We here provided an initial benchmark for simulation-based inference. If used sensibly, it will be an important tool for clarifying and expediting progress in SBI. We hope that the current results on multiple widely-used algorithms already provide insights into the state of the field, assist researchers with algorithm development, and that our recommendations for practitioners will help them in selecting appropriate algorithms.

We believe that the full potential of the benchmark will be revealed as more researchers participate and contribute. To facilitate this process, and allow users to quickly explore and compare algorithms, we are providing precomputed reference posteriors, a website (\linkwebsite{}), and open-source code (\linkrepo{}). 

\newpage

\section*{Acknowledgements}
\addcontentsline{toc}{section}{\protect\numberline{\thesection}{Acknowledgements}}
\label{acknowledgements}

We thank Álvaro Tejero-Cantero, Auguste Schulz, Conor Durkan, François Lanusse, Leandra White, Marcel Nonnenmacher, Michael Deistler, Pedro Rodrigues, Poornima Ramesh, Sören Becker and Theofanis Karaletsos for discussions and comments on the manuscript. In addition, J.-M.L. would like to thank the organisers and participants of the Likelihood-Free Inference Workshop hosted by the Simons Foundation for discussions, in particular, Danley Hsu, François Lanusse, George Papamakarios, Henri Pesonen, Joeri Hermans, Johann Brehmer, Kyle Cranmer, Owen Thomas and Umberto Simola. We also acknowledge and thank the Python \citep{van1995python} and Julia \citep{bezanson2017julia} communities for developing the tools enabling this work, including \texttt{Altair} \citep{altair}, \texttt{DifferentialEquations.jl} \citep{rackauckas2017}, \texttt{Hydra} \citep{hydra}, \texttt{kernel-gof} \citep{kgof}, \texttt{igms} \citep{igms}, \texttt{NumPy} \citep{harris2020}, \texttt{pandas} \citep{pandas}, \texttt{pyABC} \citep{klinger2018}, \texttt{pyabcranger} \citep{collin2020}, \texttt{Pyro} \citep{bingham2018}, \texttt{PyTorch} \citep{paszke2019}, \texttt{sbi} \citep{sbi}, \texttt{Scikit-learn} \citep{scikit-learn},  \texttt{torch-two-sample} \citep{tts}, and \texttt{vega-lite} \citep{vega-lite}. 

%The authors gratefully acknowledge the computational and data resources provided by the Leibniz Supercomputing Centre. 

This work was supported by the German Research Foundation (DFG; SFB 1233 PN 276693517, SFB 1089, SPP 2041, Germany’s Excellence Strategy – EXC number 2064/1 PN 390727645) and the German Federal Ministry of Education and Research (BMBF; project 'ADIMEM', FKZ 01IS18052 A-D).

%% file: appendix/A_algorithms.tex
\renewcommand{\thesection}{A}
\section{Algorithms}
\label{appendix:algorithms}

\input{appendix/algorithms/rej_abc}
\input{appendix/algorithms/smc_abc}
\input{appendix/algorithms/nle}
\input{appendix/algorithms/snle}
\input{appendix/algorithms/npe}
\input{appendix/algorithms/snpe}
\input{appendix/algorithms/nre}
\input{appendix/algorithms/snre}
\input{appendix/algorithms/rf_abc}
\input{appendix/algorithms/sl}

%% file: appendix/algorithms/rej_abc.tex
\subsection[Rejection Approximate Bayesian Computation (REJ-ABC)]{Rejection Approximate Bayesian Computation (\ABC{})}
\label{appendix:algorithms:mcabc}

\begin{algorithm}
    \SetAlgoLined
    
     \While{in simulation budget}{
      Sample $\btheta^\prime$ from $p(\btheta)$

      Simulate data $\bx^\prime$ from $p(\bx | \btheta^\prime)$

      \eIf{$d(\bx^\prime, \bx_o) \leq \epsilon$}{
       Accept $\btheta^\prime$
       }{
       Reject $\btheta^\prime$
      }
    }
    
    \textbf{return} {Accepted samples $\{\btheta^\prime\}$ from $\hat{p}(\btheta | d(\bx, \bx_o) \leq \epsilon$)}

    \caption{Rejection ABC}
    \label{algorithm:abc}
\end{algorithm}

Classical Approximate Bayesian Computation (ABC) is based on Monte Carlo rejection sampling \citep{tavere1997, pritchard1999}: In rejection ABC, the evaluation of the likelihood is replaced by a comparison between observed data $\bx_o$ and simulated data $\bx$, based on a distance measure $d(\bx, \bx_o)$. Samples $\btheta$ from the approximate posterior are obtained by collecting simulation parameters that result in simulated data that is close to the observed data.

More formally, given observed data $\bx_o$, a prior $p(\btheta)$ over parameters of simulation-based model $p(\bx | \btheta)$, a distance measure $d(\bx, \bx_o)$ and an acceptance threshold $\epsilon$, rejection ABC obtains parameter samples $\btheta$ from the approximate posterior as outlined in \algo{abc}.

In theory, rejection ABC obtains samples from the true posterior $p(\btheta | \bx_o)$ in the limit $\epsilon \to 0$ and $N \to \infty$, where $N$ is the simulation budget. In practice, its accuracy depends on the trade-off between simulation budget and the rejection criterion $\epsilon$. Rejection ABC suffers from the curse of dimensionality, i.e., with linear increase in the dimensionality of $\bx$, an exponential increase in simulation budget is required to maintain accurate results. 

For the benchmark, we did not use a fixed $\epsilon$-threshold, but quantile-based rejection. Depending on the simulation budget (1k, 10k, 100k), we used a quantile of (0.1, 0.01, or, 0.001), so that \ABC{} returned 100 samples with smallest distance to $\bx_o$ in each of these cases (see \autoref{appendix:hyperparams} for different hyperparameter choices). In order to compute metrics on 10k samples, we sampled from a KDE fitted on the accepted parameters (details about KDE resampling in  \autoref{appendix:hyperparams}). \ABC{} requires the choice of the distance measure $d(\bx, \bx_o)$: here we used the $l_2$-norm.

\newpage

%% file: appendix/algorithms/smc_abc.tex
\subsection[Sequential Monte Carlo Approximate Bayesian Computation (SMC-ABC)]{Sequential Monte Carlo Approximate Bayesian Computation (\SABC{})}
\begin{algorithm}
    \SetAlgoLined
    \vspace{1mm}
     Set schedule $\bepsilon$ (including initial $\epsilon_0$), population indicator $t=0$, and population size $N$ \\ \vspace{1mm}
     Initialize weights $W_0 = 1/N$ uniformly \\
     Sample initial population $\{\btheta_0^{(i)}\}$ using rejection sampling with $\epsilon_0$ \\ \vspace{2mm}
    \While{in simulation budget}{
        Increase population indicator $t=t+1$ \\
        Set particle indicator $i=0$ \\ \vspace{1mm}
        \While{$i < N$}{
            Sample $\btheta^{\prime}$ from previous population $\{\btheta_{t-1}^{(i)}\}$ with weights $\{W_{t-1}^{(i)}\}$\;
            Perturb $\btheta^{\prime}$:  $\btheta^{\prime\prime} \sim K_t(\btheta | \btheta^{\prime})$ \\
            Simulate data $x^{\prime\prime}$ from $p(\bx| \btheta^{\prime\prime})$ \\ \vspace{1mm}
            \eIf{$d(\bx^{\prime\prime}, \bx_o) \leq \epsilon_t$}{
                Set $\btheta_t^{(i)} = \btheta^{\prime\prime}$ and $W_t^{i} = \frac{p(\btheta_t^{(i)})}{\sum_{j=1}^{N} W_{t-1}^j K_t(\btheta_t^{(i)} | \btheta_{t-1}^j)}$ \\
                Increase particle indicator $i=i+1$ \\
                }{
                reject $\btheta^{\prime\prime}$ \\
            }
        }
        Normalize weights so that $\sum_i W_t^{(i)} = 1$ \\
    }
    
    \textbf{return} {Weighted samples $\{\btheta_t^{(i)}\}$ from $\hat{p}(\btheta | d(\bx, \bx_o) \leq \epsilon)$} \vspace{1mm}

 \caption{Population Monte Carlo ABC (ABC-PMC) as in \cite{beaumont2009}}
 \label{algorithm:sabc}
\end{algorithm}

Sequential Monte Carlo Approximate Bayesian Computation (\SABC{}) algorithms \citep{beaumont2002, marjoram2006, sisson2007, toni2009} are an extension of the classical rejection ABC approach, inspired by importance sampling and sequential Monte Carlo sampling. Central to \SABC{} is the idea to approach the final set of samples from the approximate posterior by constructing a series of intermediate sets of samples slowly approaching the final set through perturbations. 

Several variants have been developed \citep[e.g., ][]{sisson2007, beaumont2009, toni2009, simola2020}. Here, we used the scheme ABC-PMC scheme of \citet{beaumont2009} and refer to it as \SABC{} in the manuscript. More formally, the description of the ABC-PMC algorithm is as follows: Given observed data $\bx_o$, a prior $p(\btheta)$ over parameters of a simulation-based model $p(\bx| \btheta)$, a distance measure $d(\bx, \bx_o)$, a schedule of acceptance thresholds $\epsilon_i$, and a kernel $K(\btheta | \btheta^{\prime})$ to perturb intermediate samples, weighted samples of the approximate posterior are obtained as described in \algo{sabc}. 

\SABC{} can improve the sampling efficiency compared to \ABC{} and avoids severe inefficiencies due to a mismatch between initial sampling and the target distribution. However, it comes with more hyperparameters that can require careful tuning to the problem at hand, e.g., the choice of distance measure, kernel, and $\epsilon$-schedule. Like, \ABC{}, \SABC{} suffers from the curse of dimensionality. 

For the benchmark, we considered the popular toolbox \texttt{pyABC} \citep{klinger2018}. Additionally, to fully understand the details of the \SABC{} approach, we also implemented our own version.  In the main paper we report results obtained with our implementation because it yielded slightly better results. A careful comparison of the two approaches, and the optimization of hyperparameters like $\epsilon$-schedule, population size and perturbation kernel variance across different tasks are shown in \autoref{appendix:hyperparams}. After optimization, the crucial parameters of \SABC{} were set to: $l_2$-norm as distance metric, quantile-based epsilon decay with 0.2 quantile, population size 100 for simulation budgets 1k and 10k, population size 1000 for simulation budget 100k, Gaussian perturbation kernel with empirical covariance from previous population scaled by 0.5. We obtained 10k samples required for calculation of metrics as follows: If a population is not complete within the simulation budget we completed it with accepted particles from the last population and recalculated all weights. We then fitted a KDE on all those particles and sampled 10k samples from the KDE. 

\newpage

%% file: appendix/algorithms/nle.tex
\subsection[Neural Likelihood Estimation (NLE)]{Neural Likelihood Estimation (\NLE{})}
\label{appendix:algorithms:nle}

\begin{algorithm}
    \SetAlgoLined

    Set $\mathcal{D}=\{\}$
     
    \For{$n = 1 : N$}{
        Sample $\btheta_n \sim p(\btheta)$ \\
        Simulate $\bx_n \sim p(\bx | \btheta_n)$ \\
        Add $(\btheta_n, \bx_n)$ to $\mathcal{D}$ \\
    }
    Train $q_{\bpsi}(\bx | \btheta)$ on $\mathcal{D}$\\
    \textbf{return} {Samples from $\hat{p}(\btheta | \bx_o) \propto q_{\bpsi}(\bx_o | \btheta) p(\btheta)$ via MCMC; $q_{\bpsi}(\bx | \btheta)$} \vspace{1mm}
 \caption{Single round Neural Likelihood as in \cite{papamakarios2019a}}
  \label{algorithm:nle}
\end{algorithm}

Likelihood estimation approaches to SBI use density estimation to approximate the likelihood $p(\bx_o | \btheta)$. After learning a surrogate $q_{\bpsi}$ ($\bpsi$ denoting the parameters of the estimator) for the likelihood function, one can for example use Markov Chain Monte Carlo (MCMC) based sampling algorithms to obtain samples from the approximate posterior $\hat{p}(\btheta | \bx_o)$. This idea dates back to using Gaussian approximations of the likelihood \citep{wood2010, sisson2018_chapter12}, and more recently, was extended to density estimation with neural networks \citep{papamakarios2019a, lueckmann2019}.

We refer to the single-round version of the (sequential) neural likelihood approach by \citet{papamakarios2019a} as \NLE{}, and outline it in \algo{nle}: Given a set of samples $\{\btheta_n, \bx_n\}_{1:N}$ obtained by sampling $\btheta_n \sim p(\btheta)$ from the prior and simulating $\bx_n \sim p(\bx | \btheta_n)$, we train a conditional neural density estimator $q_{\bpsi}(\bx | \btheta)$ modelling the conditional of data given parameters on the set $\{\btheta_n, \bx_n\}_{1:N}$. Training proceeds by maximizing the log likelihood $\sum_n \log q_{\bpsi}(\bx | \btheta)$. Given enough simulations, a sufficiently flexible conditional neural density estimator approximates the likelihood in the support of the prior $p(\btheta)$ \citep{papamakarios2019a}. Once $q_{\bpsi}$ is trained, samples from the approximate posterior $\hat{p}(\btheta | \bx_o)$ are obtained using MCMC sampling based on the approximate likelihood $\hat{p}(\bx_o | \btheta)$ and the prior $p(\btheta)$. 

\phantomsection
\label{appendix:mcmc-sampling}
For MCMC sampling, \citet{papamakarios2019a} suggest to use Slice Sampling \citep{neal2003} with a single chain. However, we observed that the accuracy of the obtained posterior samples can be substantially improved by changing the Slice Sampling scheme as follows: 1) Instead of a single chain, we used 100 parallel MCMC chains; 2) for initialization of the chains, we sampled 10k candidate parameters from the prior, evaluated them under the unnormalized approximate posterior, and used these values as weights to resample initial locations; 3) we transformed parameters to be unbounded as suggested e.g.~in \citet{bingham2018, carpenter2017,hogg2017}. In addition, we reimplemented the slice sampler to allow vectorized evaluations of the likelihood, which yielded significant computational speed-ups. 

For the benchmark, we used as density estimator a Masked Autoregressive Flow \citep[MAF, ][]{papamakarios2017} with five flow transforms, each with two blocks and 50 hidden units, $\tanh$ non-linearity and batch normalization after each layer. For the MCMC step, we used the scheme as outlined above with 250 warm-up steps and ten-fold thinning, to obtain 10k samples from the approximate posterior (1k samples from each chain). In \autoref{appendix:hyperparams} we show results for all tasks obtained with a Neural Spline Flow \citep[NSF, ][]{durkan2019neural} for density estimation, using five flow transforms, two residual blocks of 50 hidden units each, ReLU non-linearity, and 10 bins.

\newpage

%% file: appendix/algorithms/snle.tex
\subsection[Sequential Neural Likelihood Estimation (SNLE)]{Sequential Neural Likelihood Estimation (\SNLE{})}
\label{appendix:algorithms:snle}
\begin{algorithm}
    \SetAlgoLined

    Set $\hat{p}_0(\btheta | \bx_o) = p(\btheta)$ and $\mathcal{D}=\{\}$
    
    \For{r = 1 : R}{
        \For{$n = 1 : N$}{
            Sample $\btheta_n \sim \hat{p}_{r-1}(\btheta | \bx_o)$ with MCMC \\
            Simulate $\bx_n \sim p(\bx | \btheta_n)$\\
            Add $(\btheta_n, \bx_n)$ to $\mathcal{D}$\\
        }

        (Re-)train $q_{\bpsi}(\bx | \btheta)$ on $\mathcal{D}$\\
        Set $\hat{p}_r(\btheta | \bx_o) \propto q_{\bpsi}(\bx_o | \btheta) p(\btheta)$\\
    }
    \textbf{return} {Samples from $\hat{p}(\btheta | \bx_o) \propto q_{\bpsi}(\bx_o | \btheta) p(\btheta)$ via MCMC; $q_{\bpsi}(\bx | \btheta)$} \vspace{1mm}
 \caption{Sequential Neural Likelihood as in  \cite{papamakarios2019a}}
 \label{algorithm:snle}
\end{algorithm}

Sequential Neural Likelihood estimation \citep[SNLE or SNL, ][]{papamakarios2019a} extends the neural likelihood estimation approach described in the previous section to be sequential.

The idea behind sequential SBI algorithms is based on the following intuition: If for a particular inference problem, there is only a single $\bx_o$ one is interested in, then simulating data using parameters from the entire prior space might be inefficient, leading to a training set $\mcD$ that contains training data $(\btheta, \bx)$ which carries little information about the posterior $p(\btheta | \bx_o)$. Instead, to increase sample efficiency, one may draw training data points from a proposal distribution $\tilde{p}(\btheta)$, ideally obtaining $\btheta$ for which $\bx$ is close to $\bx_o$. One candidate that has been commonly used in the literature for such a proposal is the approximate posterior distribution itself.  

\SNLE{} is a multi-round version of \NLE{}, where in each round new training samples are drawn from a proposal $\tilde{p}(\btheta)$. The proposal is chosen to be the posterior estimate at $\bx_o$ from the previous round $\hat{p}(\btheta | \bx_o)$ and its samples are obtained using MCMC. The proposal controls where $q_{\bpsi}(\bx | \btheta)$ is learned most accurately. Thus, by iterating over multiple rounds, a good approximation to the posterior can be learned more efficiently than by sampling all training data from the prior. \SNLE{} is summarized in \algo{snle}. 

For the benchmark, we used as density estimator a Masked Autoregressive Flow \citep{papamakarios2017}, and MCMC to obtain posterior samples after every round, both with the same settings as described for \NLE{}. The simulation budget was equally split across 10 rounds. In \autoref{appendix:hyperparams}, we show results for all tasks obtained with a Neural Spline Flow \citep[NSF, ][]{durkan2019neural} for density estimation, using five flow transforms, two residual blocks of 50 hidden units each, ReLU non-linearity, and 10 bins.

\newpage

%% file: appendix/algorithms/npe.tex
\subsection[Neural Posterior Estimation (NPE)]{Neural Posterior Estimation (\NPE{})}
\label{apppendix:algorithms:npe}

\begin{algorithm}
    \SetAlgoLined
    \For{$j = 1 : N$}{
        Sample $\btheta_j \sim p(\btheta)$\\
        Simulate $\bx_j \sim p(\bx | \btheta_j)$\\
    }
    $\bphi \leftarrow \argmin \sum_j^N - \log q_{F(\bx_{j}, \bphi)}(\btheta_{j})$\\
    Set $\hat{p}(\btheta | \bx_o) = q_{F(\bx_o, \bphi)}(\btheta)$\\
    \textbf{return} {Samples from $\hat{p}(\btheta | \bx_o)$; $q_{F(\bx, \bphi)}(\btheta)$} \vspace{1mm}
 \caption{Single round Neural Posterior Estimation as in \cite{papamakarios2016}}
 \label{algorithm:npe}
\end{algorithm}

\NPE{} uses conditional density estimation to directly estimate the posterior. This idea dates back to regression adjustment approaches \citep{blum2010} and was extended to density estimators using neural networks \citep{papamakarios2016} more recently. 

As outlined in \algo{npe}, the approach is as follows: Given a prior over parameters $p(\btheta)$ and a simulator, a set of training data points 
$(\btheta, \bx)$ is generated. This training data is used to learn the parameters $\bpsi$ of a conditional density estimator $q_{\bpsi}(\btheta | x)$ using a neural network $F(\bx, \bphi)$, i.e., $\bpsi=F(\bx, \bphi)$. The loss function is given by the negative log probability $-\log q_{\bpsi}(\btheta | x)$. If the density
estimator $q$ is flexible enough and training data is infinite, this loss function
leads to perfect recovery of the ground-truth posterior \citep{papamakarios2016}.

For the benchmark, we used the approach by \citet{papamakarios2016} with a Neural Spline Flow \citep[NSF, ][]{durkan2019neural} as density estimator, using five flow transforms, two residual blocks of 50 hidden units each, ReLU non-linearity, and 10 bins. We sampled 10k samples from the approximate posterior $q_{F(\bx_o, \bphi)}(\btheta)$. In \autoref{appendix:hyperparams}, we compare NSFs to Masked Autoregressive Flows \citep[MAFs, ][]{papamakarios2017}, as used in \cite{greenberg2019,durkan2020}, with five flow transforms, each with two blocks and 50 hidden units, $\tanh$ non-linearity and batch normalization after each layer. 

\newpage

%% file: appendix/algorithms/snpe.tex
\subsection[Sequential Neural Posterior Estimation (SNPE)]{Sequential Neural Posterior Estimation (\SNPE{})}
\begin{algorithm}
\label{appendix:algorithms:snpe-alg}
    \SetAlgoLined

    Set $\tilde{p}_1(\btheta) = p(\btheta)$\\
    $c \leftarrow 0$ \\
    \For{$r = 1 : R$}{
        \For{$j = 1 : N$}{
            $c \leftarrow c + 1$\\
            Sample $\btheta_{c} \sim \tilde{p}_{r}(\btheta)$\\
            Simulate $\bx_{c} \sim p(\bx | \btheta_{c})$\\
        }
        $V_r(\Theta) := \begin{cases} \binom{c}{M}^{-1} & \text{ if } \Theta = \{\btheta_{b_1}, \btheta_{b_1}, \ldots,\btheta_{b_M}\} \text{ and } 1 \leq b_1 < b_2 < \ldots < b_M \leq c \\ 0  & \text{ otherwise} \end{cases}$\\

        $\bphi \leftarrow \argmin_{\bphi} \mathbb{E}_{\Theta \sim V_r(\Theta)} \left[ \sum_{\btheta_j \in \Theta} - \log \tilde{q}_{\bx_{j}, \bphi}(\btheta_{j}) \right]$\\
        Set $\tilde{p}_{r+1}(\btheta) := q_{F(\bx_o, \bphi)}(\btheta)$\\
    }
    \textbf{return} {Samples from $\hat{p}_R(\btheta | \bx_o)$; $q_{F(x, \bphi)}(\btheta)$} \vspace{1mm}
 \caption{Sequential Neural Posterior Estimation  with atomic proposals \citep{greenberg2019}}
 \label{algorithm:snpe}
\end{algorithm}

Sequential Neural Posterior Estimation \SNPE{} is the sequential analog of \NPE{}, and meant to increase sample efficiency (see also \autoref{appendix:algorithms:snle}). When the posterior is targeted directly, using a proposal distribution $\tilde{p}(\btheta)$ different from 
the prior requires a correction step---without it, the posterior under the proposal distribution would be inferred \citep{papamakarios2016}. This so-called proposal posterior is denoted by $\tilde{p}(\btheta | \bx)$: 
$$
\begin{aligned}
\label{appendix:algorithms:snpe:proposaleq}
    \tilde{p}(\btheta | \bx) &= p(\btheta | \bx) \frac{\tilde{p}(\btheta) p(\bx)}{p(\btheta) \tilde{p}(\bx)},
\end{aligned}
$$
\vspace{-0.1cm}
where $\tilde{p}(\bx) = \int_{\btheta}\tilde{p}(\btheta) p(\bx | \btheta)$. Note that for $\tilde{p}(\btheta)=p(\btheta)$, it directly follows that $\tilde{p}(\btheta | \bx) = p(\btheta | \bx)$. 

There have been three different approaches to this correction step so far, leading to three versions of SNPE \citep{papamakarios2016,lueckmann2017,greenberg2019}. All three algorithms have in common
that they train a neural network $F(\bx, \bphi)$ to learn the parameters of a family of densities $q_{\bpsi}$ to estimate the posterior. They differ in what is targeted by $q_{\bpsi}$ and which loss is used for $F$.

SNPE-A \citep{papamakarios2016} trains $F$ to target the proposal posterior $\tilde{p}(\btheta | \bx)$ by minimizing the log likelihood loss $-\sum_n \log q_{\bpsi}(\btheta_n | \bx_n)$, and then post-hoc solves for $p(\btheta | \bx)$. The analytical post-hoc step places restrictions on $q_{\bpsi}$, the proposal, and prior. \cite{papamakarios2016} used Gaussian mixture density networks, single Gaussians proposals, and Gaussian or uniform priors. SNPE-B \citep{lueckmann2017} trains $F$ with the importance weighted loss $-\sum_n \frac{p(\btheta_n}{\tilde{p}(\btheta_n)} \log q_{\bpsi}(\btheta_n | \bx_n)$ to directly recover $p(\btheta | \bx)$ without the need for post-hoc correction, removing restrictions with respect to $q_{\bpsi}$, the proposal, and prior. However, the importance weights can have high variance during training, leading to inaccurate inference for some tasks \citep{greenberg2019}. SNPE-C (APT) \citep{greenberg2019} alleviates this issue by reparameterizing the problem such that it can infer the posterior by maximizing an estimated proposal posterior. It trains $F$ to approximate $p(\btheta|\bx)$ with $q_{F(\bx, \bphi)}(\btheta)$, using a loss defined on the approximate proposal posterior $\tilde{q}_{\bx, \bphi}(\btheta)$. \cite{greenberg2019} introduce `atomic' proposals to allow for arbitrary choices of the density estimator, e.g., flows \citep{papamakarios2019c}: The loss on $\tilde{q}_{\bx, \bphi}(\btheta)$ is calculated as the expectation over proposal sets $\Theta$ sampled from a so-called `hyperproposal' $V(\Theta)$ as outlined in \algo{snpe} \citep[see][for full details]{greenberg2019}.

For the benchmark, we used the approach by \citet{greenberg2019} with `atomic' proposals and referred to it as \SNPE{}. As density estimator, we used a Neural Spline Flow \citep{durkan2019neural} with the same settings as for \NPE{}. For the `atomic' proposals, we used $M=10$ atoms (larger $M$ was too demanding in terms of memory). The simulation budget was equally split across 10 rounds and for the final round, we obtained 10k samples from the approximate posterior $\hat{p}_R(\btheta | \bx_o)$. In \autoref{appendix:hyperparams}, we compare NSFs to Masked Autoregressive Flows \citep[MAFs, ][]{papamakarios2017}, as used in \citet{greenberg2019,durkan2020}, with five flow transforms, each with two blocks and 50 hidden units, $\tanh$ non-linearity and batch normalization after each layer.

\newpage

%% file: appendix/algorithms/nre.tex
\subsection[Neural Ratio Estimation (NRE)]{Neural Ratio Estimation (\NRE{})}
\label{appendix:algorithms:nre}
\begin{algorithm}
    \SetAlgoLined
    
    Set optimization criterion $l$ (e.g., BCE)\\
     
    \For{$j = 1 : N$}{
        Sample $\btheta_j \sim p(\btheta)$\\
        Sample $\btheta_j' \sim p(\btheta)$\\
        Simulate $\bx_j \sim p(\bx | \btheta_j)$\\
    }
    $\bphi \leftarrow \argmin l(d_{\bphi}(\bx_n, \btheta_n), 1) + l(d_{\bphi}(\bx_n, \btheta_n'), 0)$\\
    Parameterize $d_{\bphi}(\bx, \btheta)$\\
    \textbf{return} {Samples from $\hat{p}(\btheta | \bx_o)$ via MCMC; $d_{\bphi}(\bx, \btheta)$} \vspace{1mm}
 \caption{Single round Neural Ratio Estimation as in \cite{hermans2019}}
 \label{algorithm:nre}
\end{algorithm}

Neural ratio estimation (\NRE{}) uses neural-network based classifiers to approximate the posterior $p(\btheta | \bx_o)$. While neural-network based approaches described in the previous sections use \textit{density estimation} to either estimate the likelihood (\XNLE) or the posterior (\XNPE), NRE algorithms (\XNRE) use \textit{classification} to estimate a ratio of likelihoods. The ratio can then be used for posterior evaluation or MCMC-based sampling.  % evaluate the posterior, or to use MCMC for sampling from it.

Likelihood ratio estimation can be used for SBI because it allows to perform MCMC without evaluating the intractable likelihood. In MCMC, the transition probability from a current parameter $\btheta_t$ to a proposed parameter $\btheta'$ depends on the posterior ratio and in turn on the likelihood ratio between the two parameters:
$$
\begin{aligned}
    \frac{p(\btheta' | \bx)}{p(\btheta_t | \bx)} &= \frac{p(\btheta') p(\bx | \btheta') / p(\bx)}{p(\btheta_t) p(\bx | \btheta_t) / p(\bx)} = \frac{p(\btheta') p(\bx | \btheta')}{p(\btheta_t) p(\bx | \btheta_t)}.
\end{aligned}
$$

Therefore, given a ratio estimator $r(\bx |\btheta', \btheta_t) = \frac{p(\bx | \btheta')}{p(\bx | \btheta_t)}$ learned from simulations, one can perform MCMC to obtain samples from the posterior, even if evaluating $p(\bx | \btheta)$ is intractable.

\cite{hermans2019} proposed the following approach for MCMC with classifiers to approximate density ratios: A classifier is trained to distinguish samples from an arbitrary $(\btheta, \bx) \sim p(\bx | \btheta)p(\btheta)$ and samples from the marginal model $(\btheta, \bx) \sim p(\btheta)p(\bx)$. This results in a likelihood-to-evidence estimator that needs to be trained only once to be evaluated for any $\btheta$. The training of the classifier $d_{\bphi}(\bx, \btheta)$ proceeds by minimizing the binary cross-entropy loss (BCE), as outlined in \algo{nre}. Once the classifier $d_{\bphi}(\bx, \btheta)$ is parameterized, it can be used to perform MCMC to obtain samples from the posterior. The authors name their approach \textit{Amortized Approximate Likelihood Ratio MCMC} (AALR-MCMC): It is amortized because once the likelihood ratio estimator is trained, it is possible to run MCMC for any $\bx \sim p(\bx)$. 

Earlier ratio estimation algorithms for SBI \citep[e.g.,][]{izbicki2014high,pham2014note,cranmer2015,dutta2016likelihood} and their connections to recent methods are discussed in \cite{thomas2020}, as well as in \cite{durkan2020}. AALR-MCMC is closely related to LFIRE \citep{dutta2016likelihood} but trains an amortized classifier rather than a separate one per posterior evaluation. \cite{durkan2020} showed that the loss of AALR-MCMC is closely related to the atomic SNPE-C/APT approach of \citet{greenberg2019} (\SNPE{}) and that both can be combined in a unified framework. \cite{durkan2020} changed the formulation of the loss function for training the classifier from binary to multi-class. 

For the benchmark, we used neural ratio estimation (\NRE{}) as formulated by \citet{durkan2020} and implemented in the \texttt{sbi} toolbox \citep{sbi}. As a classifier, we used a residual network architecture (ResNet) with two hidden layers of 50 units and ReLU non-linearity, trained with Adam \citep{kingma2014adam}. Following the notation of \cite{durkan2020}, we used $K=10$ as the size of the contrasting set. For the MCMC step, we followed the same procedure as described for \NLE{}, i.e., using Slice Sampling with 100 chains, to obtain 10k samples from each approximate posterior. In \autoref{appendix:hyperparams}, we show results for all tasks obtained with a multi-layer perceptron (MLP) architecture with two hidden layers of 50 ReLu units, and batch normalization. 

\newpage

%% file: appendix/algorithms/snre.tex
\subsection[Sequential Neural Ratio Estimation (SNRE)]{Sequential Neural Ratio Estimation (\SNRE{})}
\label{appendix:algorithms:snre}
\begin{algorithm}
    \SetAlgoLined
     
     Set optimization criterion $l$ (e.g., BCE)\\
     Set $\tilde{p}(\btheta) = p(\btheta)$\\
     
    \For{$r = 1 : R$}{
        \For{$j = 1 : N$}{
            Sample $\btheta_j \sim \tilde{p}(\btheta)$ (via $d_{\bphi}$ and MCMC)\\
            Sample $\btheta_j' \sim \tilde{p}(\btheta)$ (via $d_{\bphi}$ and MCMC)\\
            Simulate $\bx_j \sim p(\bx | \btheta_j)$\\
        }
        $\bphi \leftarrow \argmin l(I'n, \btheta_n), 1) + l(d_{\bphi}(\bx_n, \btheta_n'), 0)$\;
        Parameterize $d_{\bphi}(\bx, \btheta)$\\
    }
    \textbf{return} {Samples from $\hat{p}(\btheta | \bx_o)$ via MCMC; $d_{\bphi}(\bx, \btheta)$} \vspace{1mm}
 \caption{Sequential Neural Ratio Estimation as in \citet{hermans2019}}
    \label{algorithm:snre}
\end{algorithm}

Sequential Neural Ratio Estimation (\SNRE{}) is the sequential version of \NRE{}, and meant to increase sample efficiency, at the cost of needing to train new classifiers for different $\bx_o$. 

A sequential version of neural ratio estimation was proposed by \citet{hermans2019}. As with other sequential algorithms, the idea is to replace the prior by a proposal distribution $\tilde{p}(\btheta)$ that is focused on $\bx_o$ in the sense that the sampled parameters $\btheta$ result in simulated data $\bx$ that are informative about $\bx_o$. The proposal for the next round is the posterior estimate from the previous round. The ratio estimator then becomes $\tilde{r}(\bx, \btheta)$ and is refined over rounds by training the underlying classifier with positive examples $(\bx, \btheta) \sim p(\bx | \btheta) \tilde{p}(\btheta)$ and negative examples $(\bx, \btheta) \sim p(\bx) \tilde{p}(\btheta)$. Exact posterior evaluation is not possible anymore, but samples can be obtained as before via MCMC. These steps are outlined in \algo{snre}.

For the benchmark, we used \SNRE{} as formulated by \citet{durkan2020} and implemented in the \texttt{sbi} toolbox \citep{sbi}. The classifier had the same architecture as described for \NRE{}. For the MCMC step, we followed the same procedure as described for \NLE{}. The simulation budget was equally split across 10 rounds. In \autoref{appendix:hyperparams}, we show results for all tasks obtained with a multi-layer perceptron (MLP) architecture with two hidden layers of 50 ReLu units, and batch normalization. 

\newpage

%% file: appendix/algorithms/rf_abc.tex
\subsection[Random Forest Approximate Bayesian Computation (RF-ABC)]{Random Forest Approximate Bayesian Computation (\RFABC{})}
\label{appendix:algorithms:rfabc}
\begin{algorithm}
    \SetAlgoLined

    Set $\mathcal{D}=\{\}$
    Set simulation budget $N$\\
    Set number of trees $B$\\
    Set minimum node size $N_{min}$
     
    \For{$n = 1 : N$}{
        Sample $\btheta_n \sim p(\btheta)$ \\
        Simulate $\bx_n \sim p(\bx | \btheta_n)$ \\
        Add $(\btheta_n, \bx_n)$ to $\mathcal{D}$ \\
    }
    Run random forest regression of $\bx$ on $\btheta$ using $\mathcal{D}$, $B$ and $N_{min}$\\

    \textbf{return} {$N$ samples $\{\btheta^{(i)}\}$ and associated weights $\{w^{(i)}\}$ for drawing approximate posterior samples} \vspace{1mm}

 \caption{Random Forest ABC (RF-ABC) as in \citet{raynal2018}}
 \label{algorithm:rfabc}
\end{algorithm}

Random forest Approximate Bayesian Computation \citep[\RFABC{},][]{pudlo2016, raynal2018} is a more recently developed ABC algorithm based on a regression approach. Similar to previous regression approaches to ABC \citep{beaumont2002, blum2010}, \RFABC{} aims at improving classical ABC inference (\ABC{}, \SABC{}) in the setting of high-dimensional data. 

The idea of the \RFABC{} algorithm is to use random forests \citep[RF, ][]{breiman2001} to run a non-parametric regression of a set of potential summary statistics of the data on the corresponding parameters. That is, the RF regression is trained on data simulated from the model, such that the covariates are the summary statistics and the response variable is a parameter. For a detailed description of the algorithm, we refer to \citet{raynal2018}. 

The only hyperparameters for the \RFABC{} algorithm are the number of trees and the minimum node size for the RF regression. Following \citet{raynal2018}, we chose the default of 500 trees and a minimum of 5 nodes. The output of the algorithm is a RF weight for each of the simulated parameters. This set of weights can be used to calculate posterior quantiles or to obtain an approximate posterior density as described in \citet{raynal2018}. We obtained 10k posterior samples for the benchmark by using the random forest weights to sample from the simulated parameters. We used the implementation in the \texttt{abcranger} toolbox \citet{collin2020}.

One important property of \RFABC{} is that it can only be applied in the unidimensional setting, i.e., for 1-D dimensional parameter spaces, or for multidimensional parameters spaces with the assumption that the posterior factorizes over parameters (thus ignoring potential posterior correlations). This assumptions holds only for a few tasks in our benchmark (Gaussian Linear, Gaussian Linear Uniform, Gaussian Mixture). Due to this inherent limitation, we report \RFABC{} in the supplement (see \suppfig{rf_abc}).

\newpage

%% file: appendix/algorithms/sl.tex
\subsection[Synthetic Likelihood (SL)]{Synthetic Likelihood (\SL{})}
\label{appendix:algorithms:sl}

\begin{algorithm}
    \SetAlgoLined

    Set number of simulations per step $M$\\
    Set number of MCMC steps $T$\\
    
    \For{$t=1:T$}{
        
        Get new candidate $\btheta_t$ from MCMC scheme \\
        Set $\mathcal{D}_t=\{\}$ \\
        \For{$m=1 : M$}{
            Simulate $\bx_m \sim p(\bx | \btheta_t)$ \\
            Add $(\btheta_t, \bx_m)$ to $\mathcal{D}_t$ \\
        }
        Use $\mathcal{D}_t$ to estimate mean and covariance of a Gaussian approximation of the likelihood $\hat{L}(\bx_o | \theta_t)$ \\
        Perform the next MCMC step using $\hat{L}(\bx_o | \theta_t)$
    }

    \textbf{return} {$N$ samples $\{\btheta^{(i)}\}$ from MCMC chain} \vspace{1mm}

 \caption{Synthetic Likelihood algorithm as in \citet{wood2010}}
 \label{algorithm:sl}
\end{algorithm}

The Synthetic Likelihood (\SL{}) approach circumvents the evaluation of the intractable likelihood by estimating a \textit{synthetic} one from simulated data or summary statistics. This approach was introduced by \citet{wood2010}. Its main motivation is that the classical ABC approach of comparing simulated and observed data with a distance metric can be problematic if parts of the differences are entirely noise-driven. \citet{wood2010} instead approximated the distribution of the summary statistics (the likelihood) of a nonlinear ecological dynamic system as a Gaussian distribution, thereby capturing the underlying noise as well. The approximation of the likelihood can then be used to obtain posterior sampling via Markov Chain Monte Carlo (MCMC) \citep{wood2010}.

The \SL{} approach can be seen as the predecessor of the \XNLE{} approaches: They replaced the Gaussian approximation of the likelihood with a much more flexible one that uses neural networks and normalizing flows (see \ref{appendix:algorithms:nle}). Moreover, there are modern approaches from the classical ABC field that further developed \SL{} using a Gaussian approximation \citep[e.g.,][]{sisson2018_chapter12,priddle2019}.

For the benchmark, we implemented our own version of the algorithm proposed by \citet{wood2010}. We used Slice Sampling MCMC \citep{neal2003} and estimated the Gaussian likelihood from 100 samples at each sampling step. To ensure a positive definite covariance matrix, we added a small value $\epsilon$ to the diagonal of the estimated covariance matrix for some of the tasks. In particular, we used $\epsilon=0.01$ for SIR and Bernoulli GLM Raw tasks, and we tried without success $\epsilon=[0, 0.01, 0.1, 1.0]$ for Lotka-Volterra and SLCP with distractors. For all remaining tasks, we set $\epsilon=0$. For Slice Sampling, we used a single chain initialized with sequential importance sampling (SIR) as described for \NLE{}, 1k warm-up steps and no thinning, in order to keep the number of required simulations tractable. This resulted in an overall simulation budget on the order of $10^8$ to $10^9$ simulations per run in order to generate 10k posterior samples, as new simulations are required for every MCMC step.

The high simulation budget makes it problematic to directly compare \SL{} and other other algorithms in the benchmark. Therefore, we report \SL{} in the supplement (see \suppfig{sl}).

\newpage
\clearpage

%% file: appendix/B_benchmark.tex
\renewcommand{\thesection}{B}
\section{Benchmark}
\label{appendix:benchmark}

\subsection{Reference posteriors}

We generated 10k reference posterior samples for each observation. For the \nameref{appendix:task:gaussian_linear} task, reference samples were obtained by using the analytic solution for the true posterior. Similarly, for \nameref{appendix:task:gaussian_linear_uniform} and  \nameref{appendix:task:gaussian_mixture}, the analytic solution was used, combined with an additional rejection step, in order to account for the bounded support of the posterior due to the use of a uniform prior. For the \nameref{appendix:task:two_moons} task, we devised a custom scheme based on the model equations, which samples both modes and rejects samples outside the prior bounds. 

For \nameref{appendix:task:slcp}, \nameref{appendix:task:sir}, and \nameref{appendix:task:lotka_volterra}, we devised a likelihood-based procedure to ensure obtaining a valid set of reference posterior samples: First, we either used Sampling/Importance Resampling \citep{rubin1988using} (for \nameref{appendix:task:slcp}, \nameref{appendix:task:sir}) or Slice Sampling MCMC \citep{neal2003} (for \nameref{appendix:task:lotka_volterra}) to obtain a set of 10k proposal samples from the unnormalized posterior $f(\btheta) = \tilde{p}(\btheta|\bx_o) = p(\bx_o|\btheta) p(\btheta)$. We used these proposal samples to train a density estimator, for which we used a neural spline flow (NSF) \citep{durkan2019neural}. Next, we created a mixture composed of the NSF and the prior with weights 0.9 and 0.1, respectively, as a proposal distribution $g(\btheta)$ for rejection sampling \citep{martino2018accept}. Rejection sampling relies on finding a constant $M$ such that $f(\btheta) \leq M g(\btheta)$ for all values of $\btheta$: To find this constant, we initialized $M=1$, sampled $\btheta \sim g(\btheta)$, and updated $M=1.2 f(\btheta) / g(\btheta)$ if $f(\btheta) / g(\btheta) > M$. This loop stopped only after at least 100k samples without updating $M$ were reached. We then used $M$, $f$, and $g$ to generate 10k reference posterior samples. We found that the NSF-based proposal distribution resulted in high acceptance rates. We used this custom scheme rather than relying on MCMC directly, since we found that standard MCMC approaches (Slice Sampling, HMC, and NUTS) all struggled with multi-modal posteriors and wanted to avoid bias in the reference samples, e.g.~due to correlations in MCMC chains. 

As a sanity check, we ran this scheme twice on all tasks and observation and found that the resulting reference posterior samples were indistinguishable in terms of C2ST.

\subsection{Code}

We provide \texttt{sbibm}, a benchmarking framework that implements all tasks, reference posteriors, different metrics and tooling to run and analyse benchmark results at scale. The framework is available at:

\linkrepo{}

We make benchmarking new algorithms maximally easy by providing an open, modular framework for \textit{integration} with SBI toolboxes. We here evaluated algorithms implemented in \texttt{pyABC} \citep{klinger2018}, \texttt{pyabcranger} \citep{collin2020}, and \texttt{sbi} \citep{sbi}. We emphasize that the goal of \texttt{sbibm} is orthogonal to any toolbox: It could easily be used with other toolboxes, or even be used to compare results for the same algorithm implemented by different ones. There are currently several SBI toolboxes available or under active development. \texttt{elfi} \citep{elfi2018} is a general purpose toolbox, including ABC algorithms as well as BOLFI \citep{gutmann2015}. There are many toolboxes for ABC algorithms, e.g., \texttt{abcpy} \citep{abcpy-repo}, \texttt{astroABC} \citep{jennings2017astroabc}, \texttt{CosmoABC} \citep{ishida2015cosmoabc}, see also \citet{sisson2018_chapter13} for an overview. \texttt{carl} \citep{louppe2016} implements the algorithm by \citet{cranmer2015}. \texttt{hypothesis} \citep{hypothesis-repo}, and \texttt{pydelfi} \citep{pydelfi-repo} are SBI toolboxes under development.

%We optimized the tasks for speed, so that the simulation cost will be negligible when running benchmark tasks. 

\subsection{Reproducibility}

To ensure reproducibility of our results, we publicly released all code including instructions on how to run the benchmark on cloud-based infrastructure.

%% file: appendix/F_figures.tex
\renewcommand{\thesection}{F}
\section{Figures}
\label{appendix:figures}

\begin{figure*}[h!]
    \centering
    \begin{subfigure}[htbp]{\textwidth}
        \includegraphics[trim=20 49 0 3,clip,width=\textwidth]{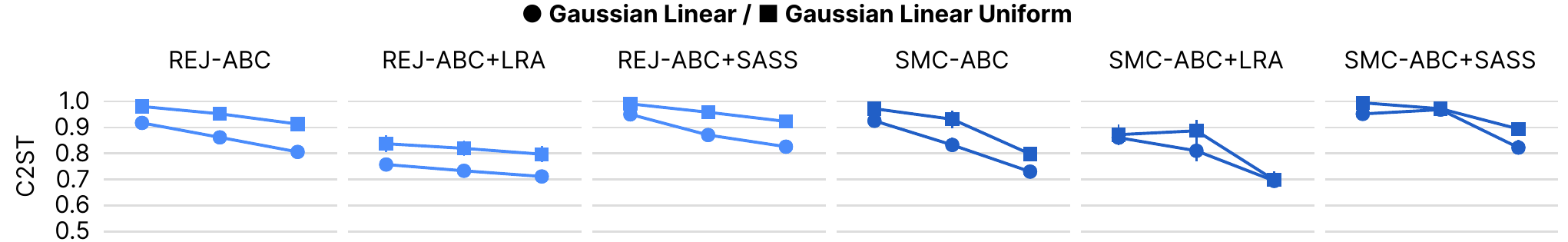}
    \end{subfigure}
    \begin{subfigure}[htbp]{\textwidth}
        \includegraphics[trim=20 49 0 0,clip,width=\textwidth]{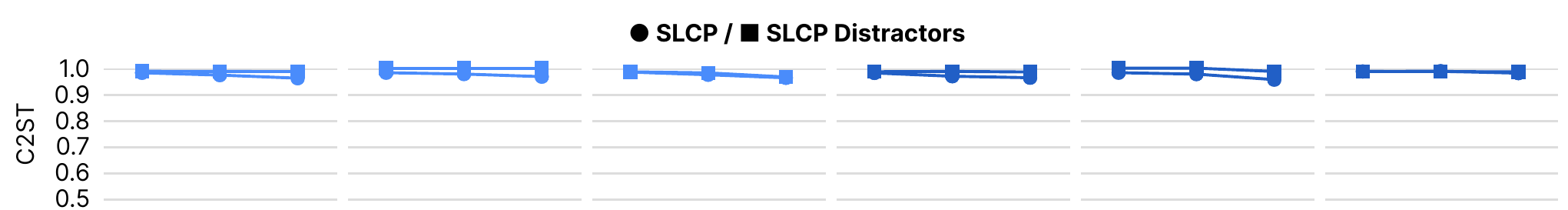}
    \end{subfigure}    
    \begin{subfigure}[htbp]{\textwidth}    
        \includegraphics[trim=20 49 0 0,clip,width=\textwidth]{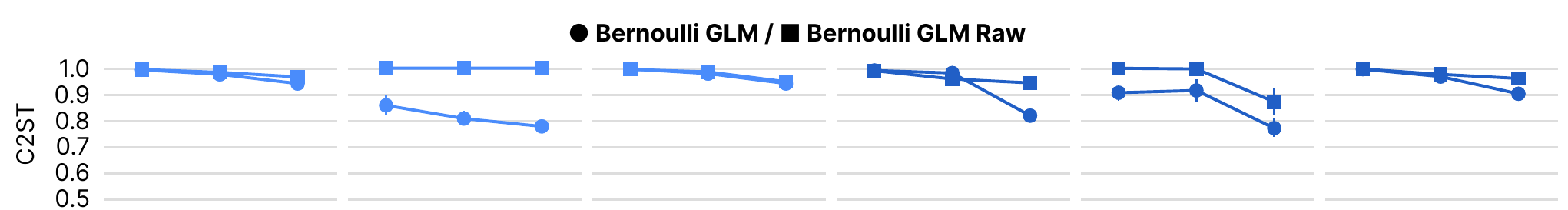}  
    \end{subfigure}
    \begin{subfigure}[htbp]{\textwidth}
        \includegraphics[trim=20 49 0 0,clip,width=\textwidth]{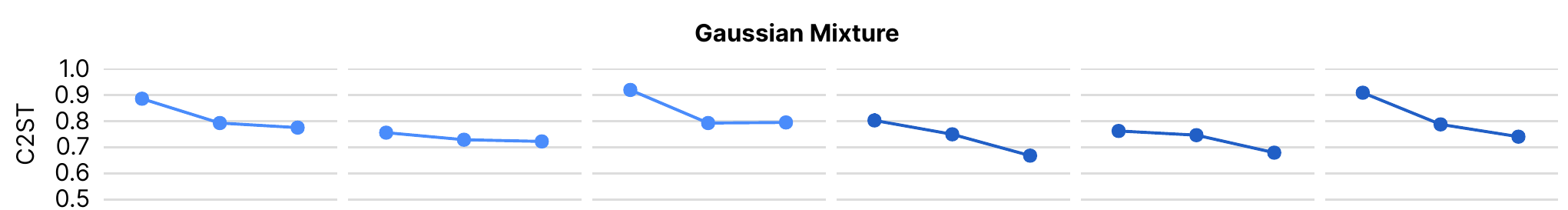}
    \end{subfigure}
    \begin{subfigure}[htbp]{\textwidth}
        \includegraphics[trim=20 49 0 0,clip,width=\textwidth]{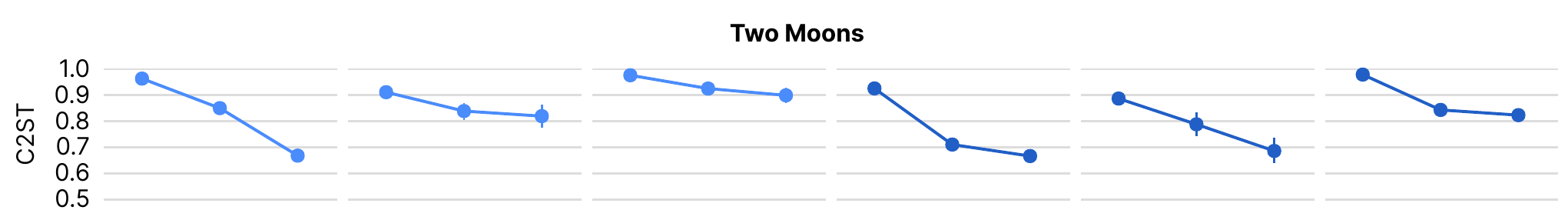}
    \end{subfigure}
    \begin{subfigure}[htbp]{\textwidth}    
        \includegraphics[trim=20 49 0 0,clip,width=\textwidth]{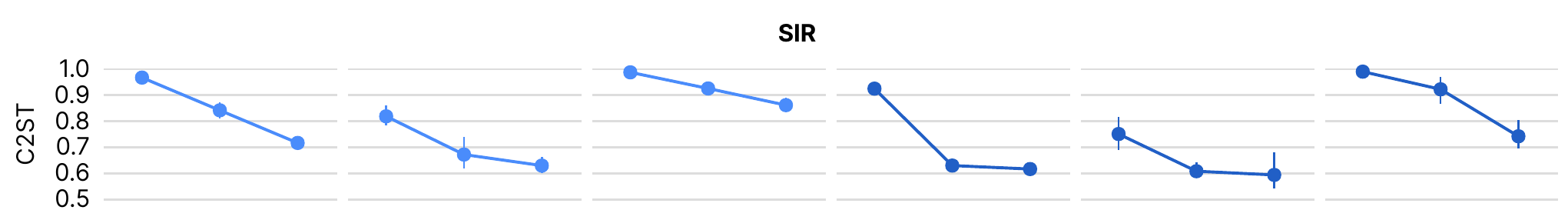}  
    \end{subfigure}
    \begin{subfigure}[htbp]{\textwidth}    
        \includegraphics[trim=20 10 0 0,clip,width=\textwidth]{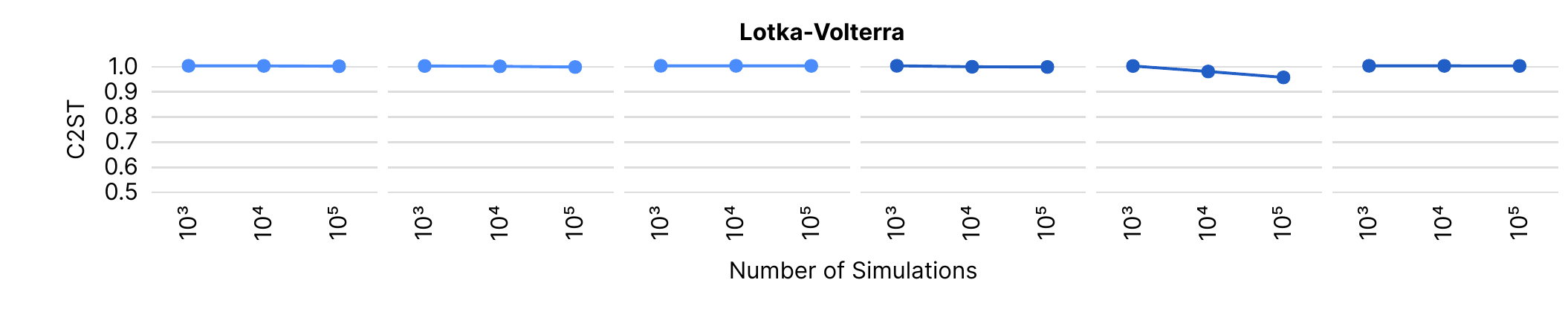}
    \end{subfigure}    %
    \caption{
        {\bf Additional ABC results with linear regression adjustment (LRA) and semi-automatic summary-statistics (SASS).} We ran ABC with post-hoc LRA \citep{beaumont2002, sisson2018_chapter3}. On some tasks, this led to an improvement relative to versions without post-hoc adjustment. On Two Moons (bimodal posterior), linear adjustment decreased performance. We implemented our own SASS \citep{prangle2014semi} with a third order polynomial feature expansion, and observed similar performance as with the implementation in \texttt{abcpy} toolbox \citep{abcpy-repo}. 
    }
    \label{fig:abc_additional}
\end{figure*}

\newpage
\clearpage

\begin{figure*}[h!]
    \centering
    \begin{subfigure}[htbp]{\textwidth}
        \includegraphics[trim=20 49 0 3,clip,width=\textwidth]{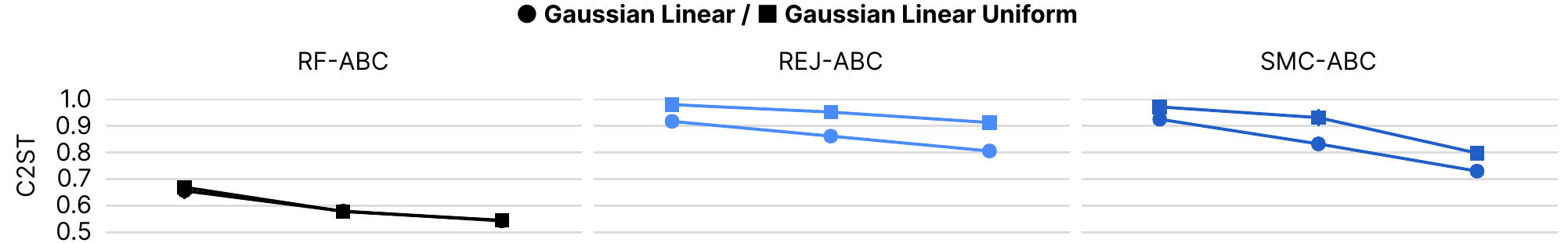}
    \end{subfigure}
    \begin{subfigure}[htbp]{\textwidth}
        \includegraphics[trim=20 49 0 0,clip,width=\textwidth]{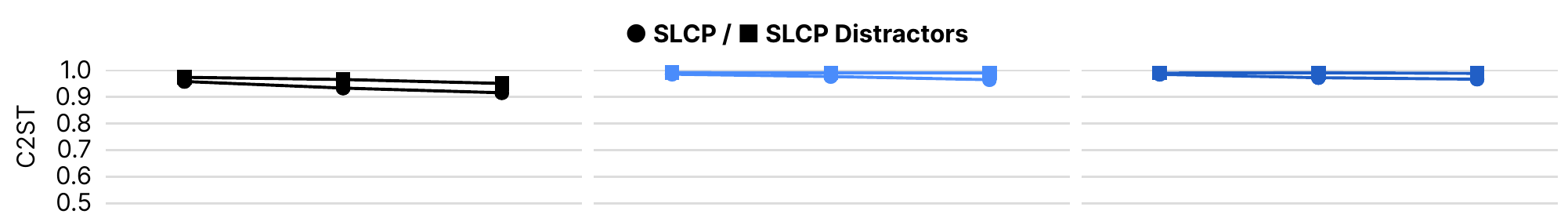}
    \end{subfigure}    
    \begin{subfigure}[htbp]{\textwidth}    
        \includegraphics[trim=20 49 0 0,clip,width=\textwidth]{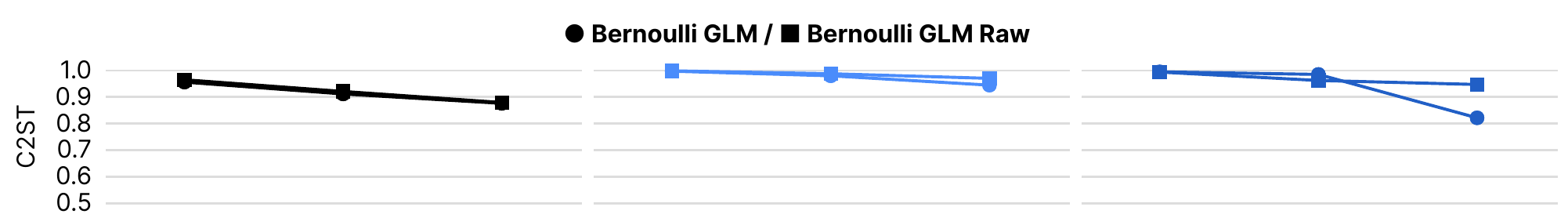}  
    \end{subfigure}
    \begin{subfigure}[htbp]{\textwidth}
        \includegraphics[trim=20 49 0 0,clip,width=\textwidth]{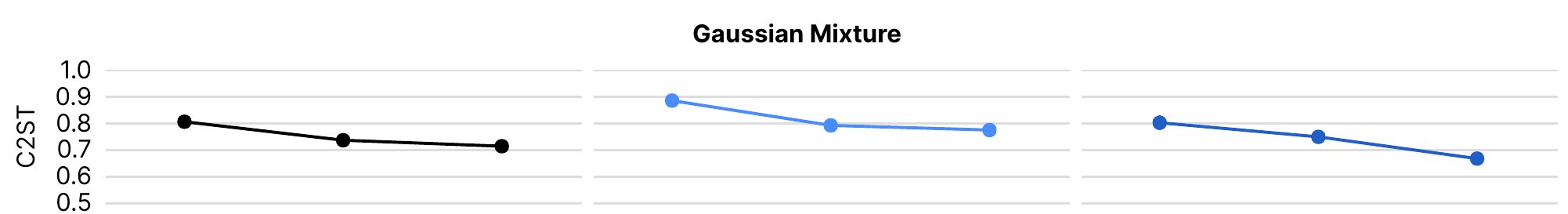}
    \end{subfigure}
    \begin{subfigure}[htbp]{\textwidth}
        \includegraphics[trim=20 49 0 0,clip,width=\textwidth]{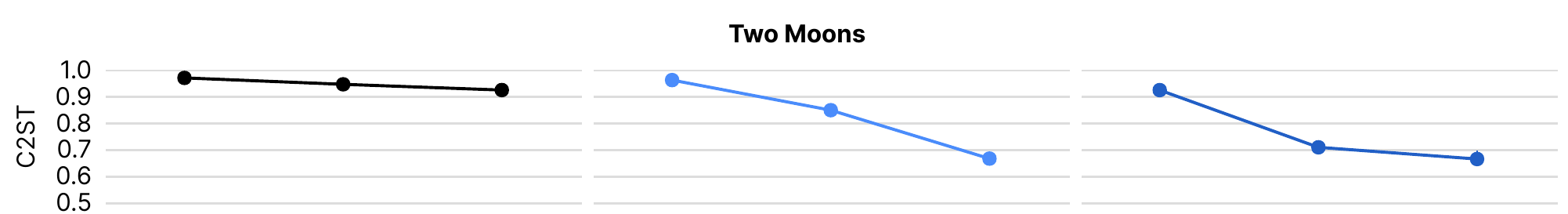}
    \end{subfigure}
    \begin{subfigure}[htbp]{\textwidth}    
        \includegraphics[trim=20 49 0 0,clip,width=\textwidth]{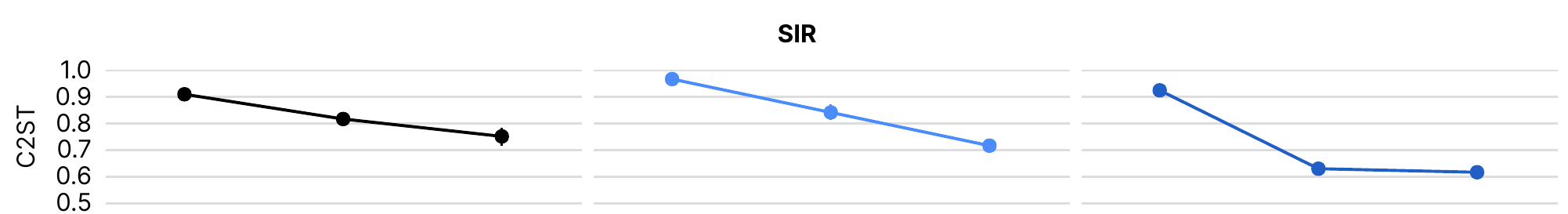}  
    \end{subfigure}
    \begin{subfigure}[htbp]{\textwidth}    
        \includegraphics[trim=20 10 0 0,clip,width=\textwidth]{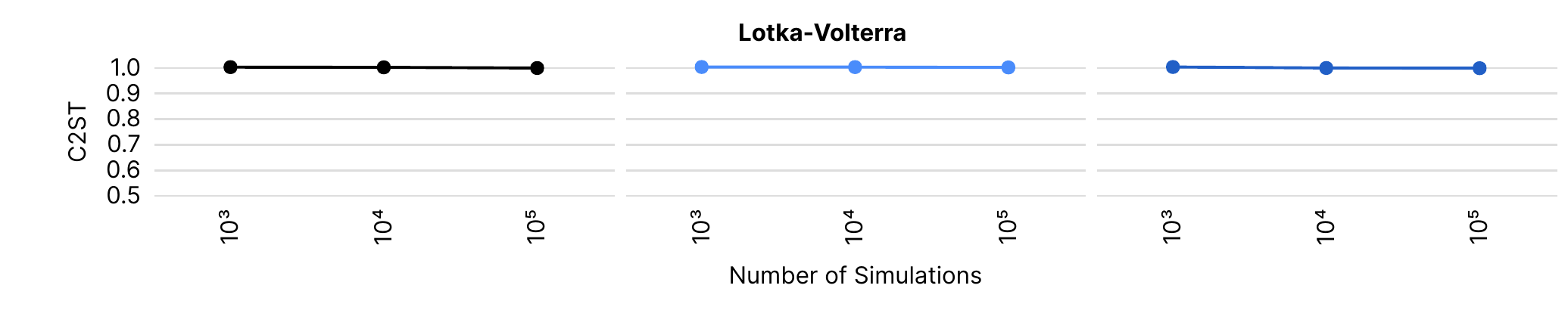}  
    \end{subfigure}
    \caption{
        {\bf \RFABC{} results.} Results for \RFABC{} (as described in \ref{appendix:algorithms:rfabc}) compared to \ABC{} and \SABC{} on all benchmark tasks, using C2ST. Note that \RFABC{} predicts each parameter individually, i.e. effectively assumes the posterior to be factorized-- this is only appropriate for the Gaussian Linear, Gaussian Linear Uniform, and Gaussian Mixture tasks. On other tasks, the posterior deviates markedly from being factorized, and therefore it is to be expected that \RFABC{} performance is limited, even when using many samples. Each data point corresponds to the mean and 95\% confidence interval across 10 observations.
    }
    \label{fig:rf_abc}
\end{figure*}

\newpage
\clearpage

\begin{figure}[h!]
    \centering
    \begin{subfigure}[htbp]{\textwidth}
        \includegraphics[trim=20 55 0 3,clip,width=\textwidth]{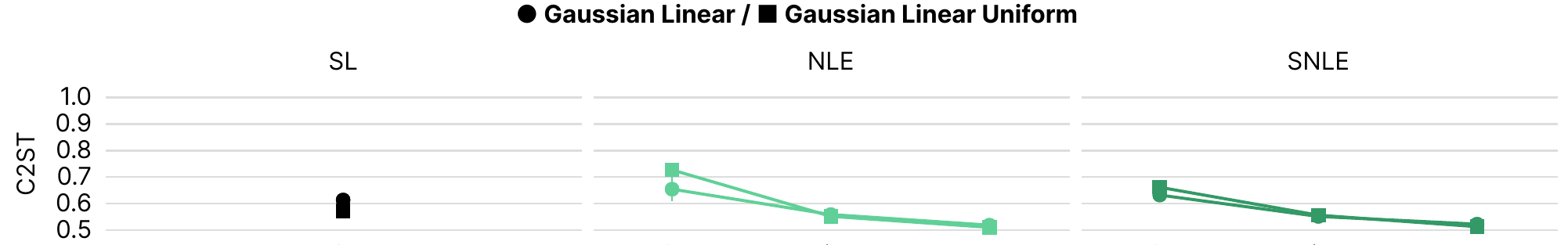}
    \end{subfigure}
    \begin{subfigure}[htbp]{\textwidth}
        \includegraphics[trim=20 55 0 0,clip,width=\textwidth]{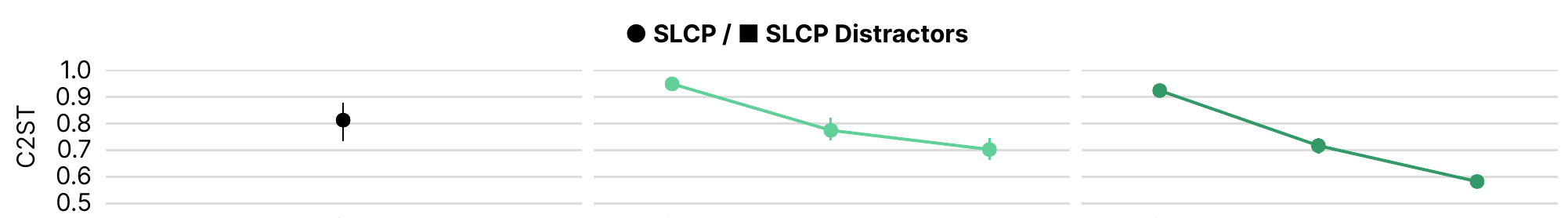}
    \end{subfigure}    
    \begin{subfigure}[htbp]{\textwidth}    
        \includegraphics[trim=20 55 0 0,clip,width=\textwidth]{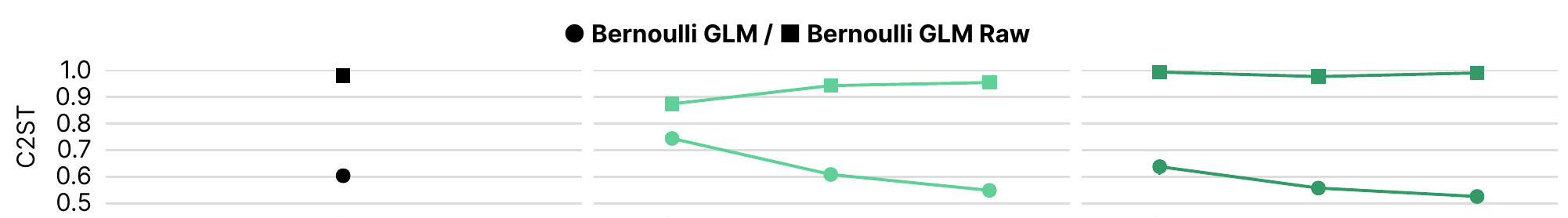}  
    \end{subfigure}
    \begin{subfigure}[htbp]{\textwidth}
        \includegraphics[trim=20 55 0 0,clip,width=\textwidth]{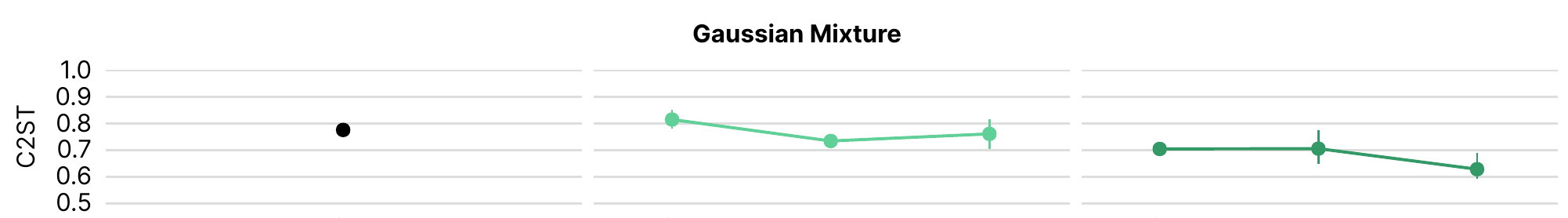}
    \end{subfigure}
    \begin{subfigure}[htbp]{\textwidth}
        \includegraphics[trim=20 55 0 0,clip,width=\textwidth]{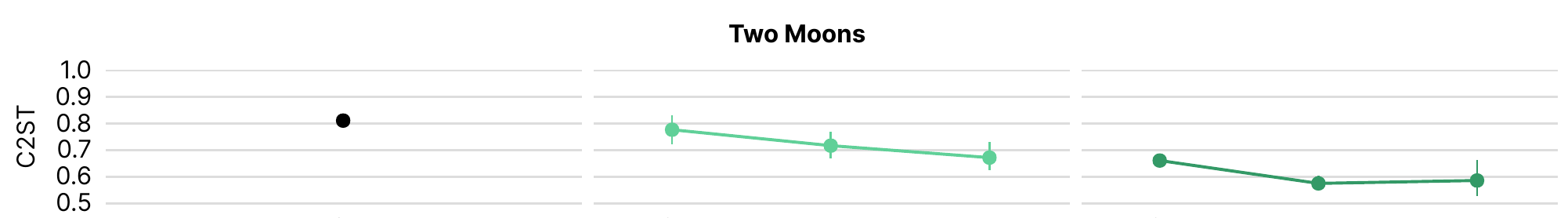}
    \end{subfigure}
    \begin{subfigure}[htbp]{\textwidth}    
        \includegraphics[trim=20 10 0 0,clip,width=\textwidth]{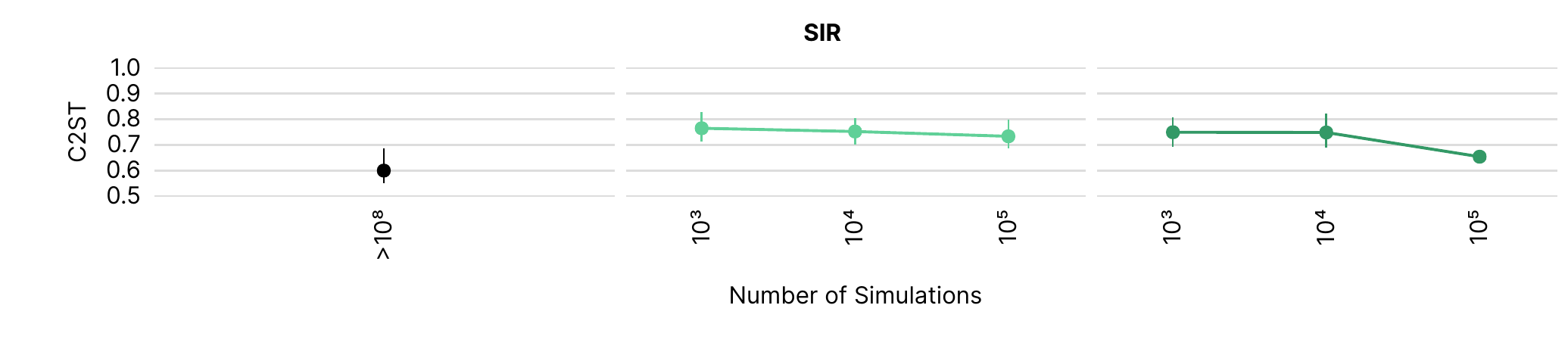}  
    \end{subfigure}
    \caption{
            {\bf \SL{} results.} Results for \SL{} compared to \NLE{} and \SNLE{} on benchmark tasks in terms of C2ST. Note that \SL{} performs simulations at every MCMC step  to approximate a Gaussian likelihood (see \ref{appendix:algorithms:sl} for details), and therefore it does not produce sensible results with the simulation budgets of other algorithms (between 1k and 100k), . In our experiments, \SL{} required on the order of $10^8$ to $10^9$ simulations. For the SLCP Distractors and Lotka-Volterra stable estimation of covariances was not possible, which is why these tasks were omitted (details in \ref{appendix:algorithms:sl}). We do not report \SL{} results in the main paper, given the huge difference in simulation budget. Each data point corresponds to the mean and 95\% confidence interval across 10 observations.
    }
    \label{fig:sl}
\end{figure}

\newpage
\clearpage

\begin{figure}[h!]
    \centering
    \begin{subfigure}[htbp]{0.45\textwidth}
        \includegraphics[trim=0 0 0 0,clip,width=\textwidth]{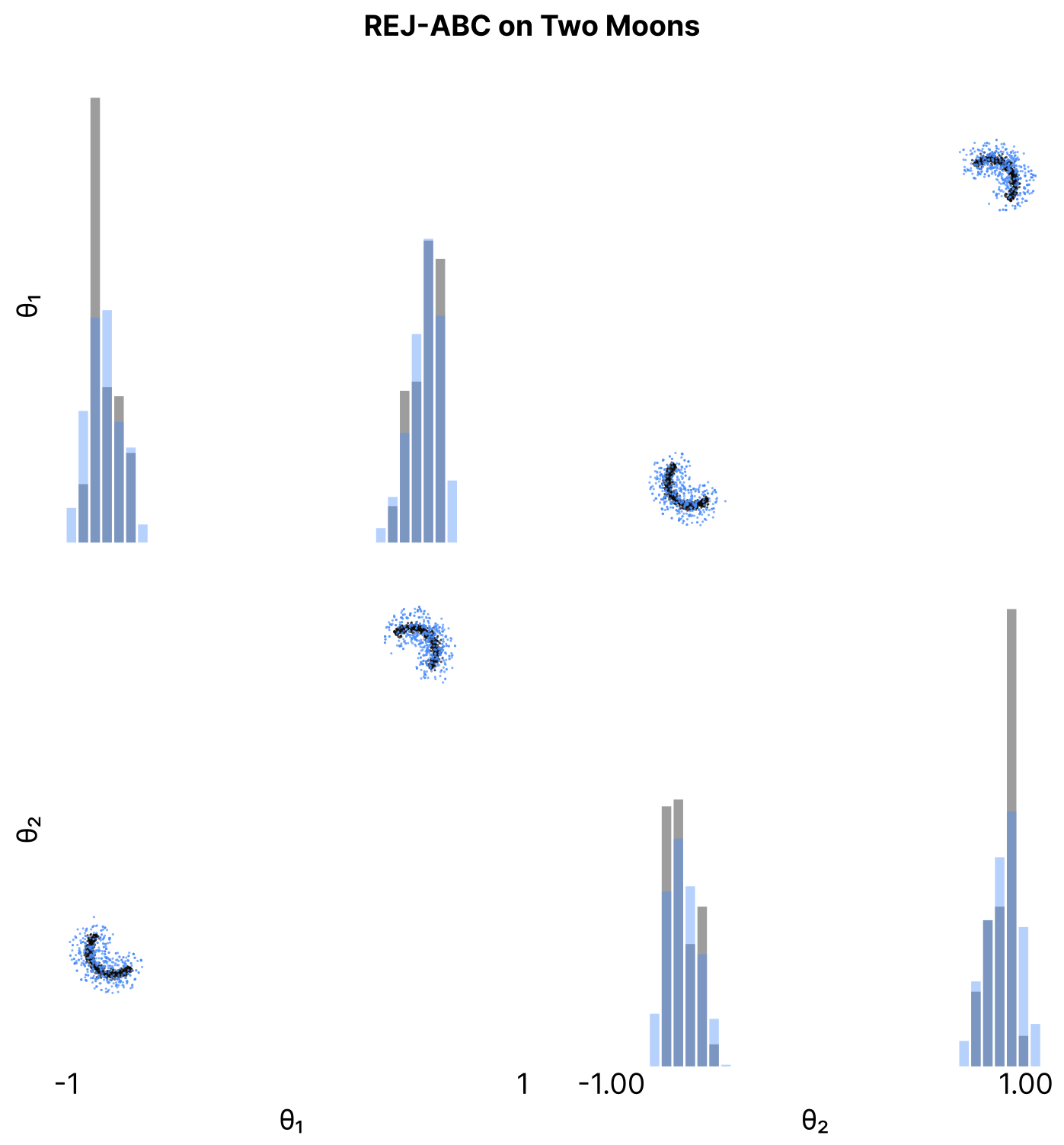}
    \end{subfigure}
    \begin{subfigure}[htbp]{0.45\textwidth}
        \includegraphics[trim=0 0 0 0,clip,width=\textwidth]{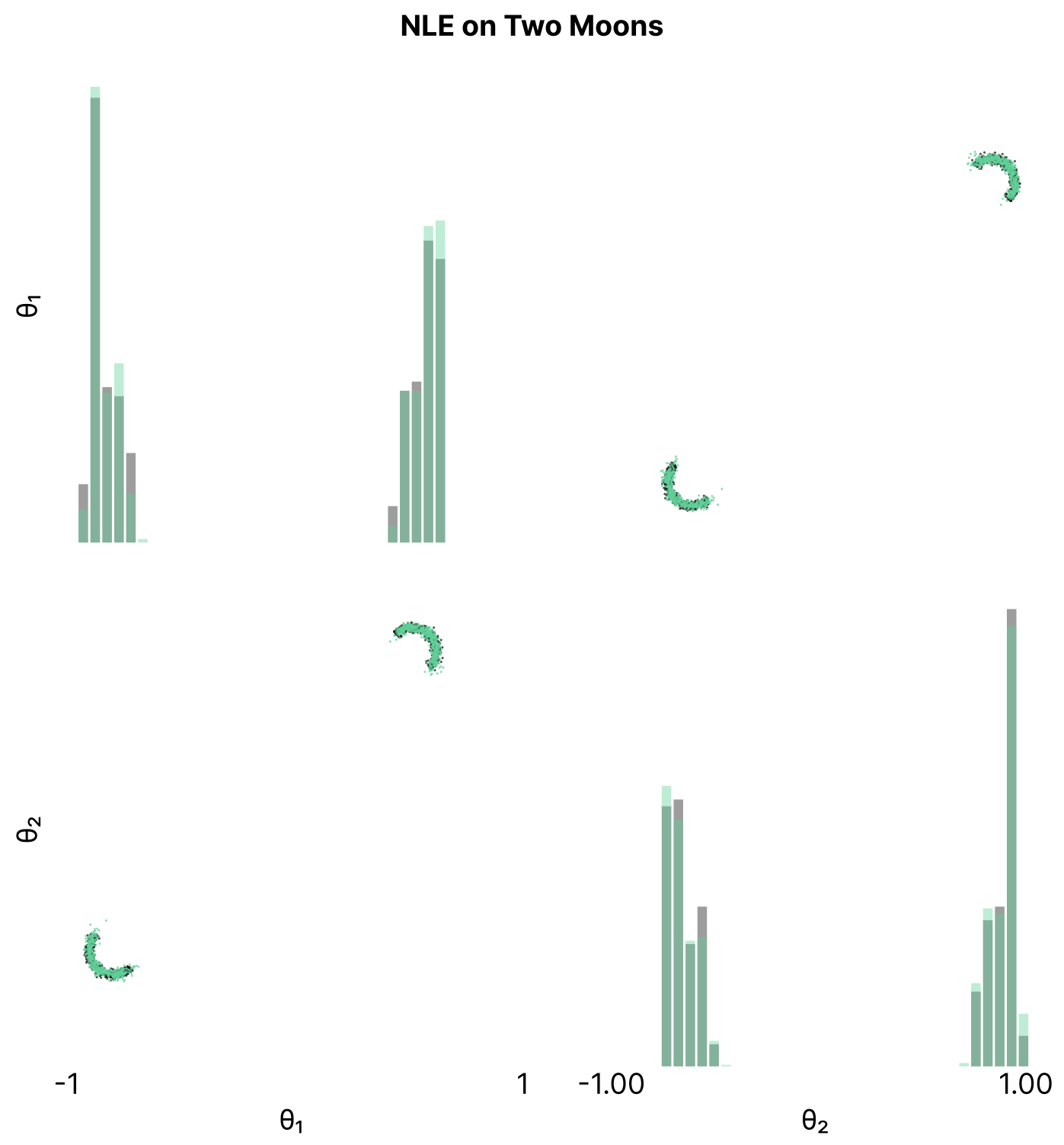}
    \end{subfigure}
    \caption{
            {\bf MMD on Two Moons.} When using MMD with the median heuristic (as commonly done, including in SBI papers), MMD is slightly lower for the posterior obtained by \ABC{} (left, blue samples), than for \SNLE{} samples (right, green samples): 0.00729 (\ABC{}) versus 0.00772 (\NLE{}). This is at odds with the visual impression of the quality of the fit (reference samples in gray) as well as C2ST results: A classifier performed near chance level (.502) for \SNLE{} samples while being able to tell apart \ABC{} samples from the reference with accuracy 0.794. When manually choosing a length scale on the median distance of a \textit{single crescent} (i.e., 0.09 instead of 1.78), MMD results were in agreement with C2ST results: 0.00738 (\ABC{}) versus 0.00035 (\SNLE{}), i.e., they also suggested a better fit for \SNLE{}. In the main paper, we prefer to report C2ST because we found it less sensitive to hyperparameters: reliance on the commonly used median heuristic can be problematic on tasks with complex posterior structure, e.g., multi-modality in Two Moons, as demonstrated here. We refer the interested reader to \citet{liu2020} for further illustrative examples of where MMD with Gaussian kernels can have limited power. We also want to point out that new kernel-based two sample tests are being actively developed which might make them easier to use on such problems in the future.
    }
    \label{fig:mmd}
\end{figure}

\newpage
\clearpage

%
% Correlations
%

\begin{figure*}[h!]
    \centering
    \begin{subfigure}[htbp]{0.28\textwidth}
        \includegraphics[trim=0 0 0 0,clip,width=\textwidth]{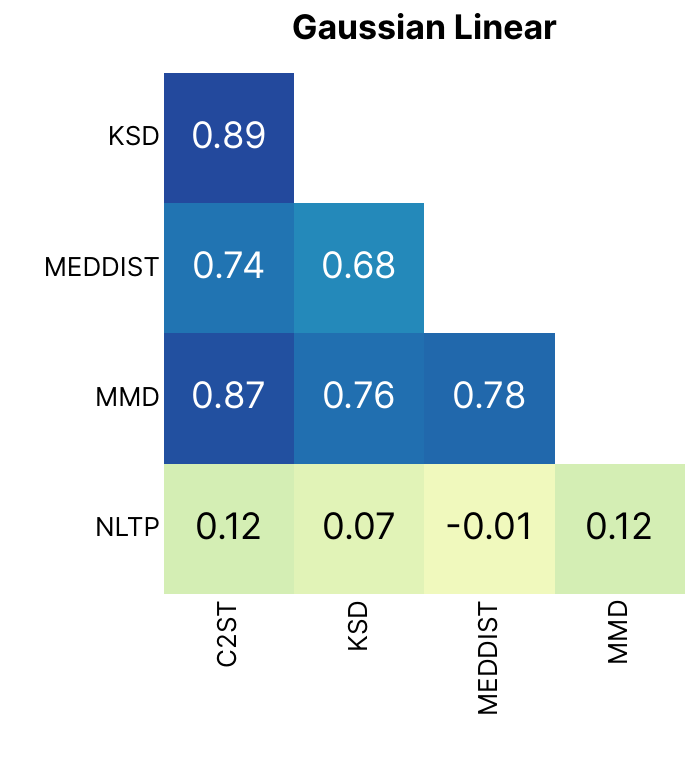}
    \end{subfigure}
    \begin{subfigure}[htbp]{0.28\textwidth}
        \includegraphics[trim=0 0 0 0,clip,width=\textwidth]{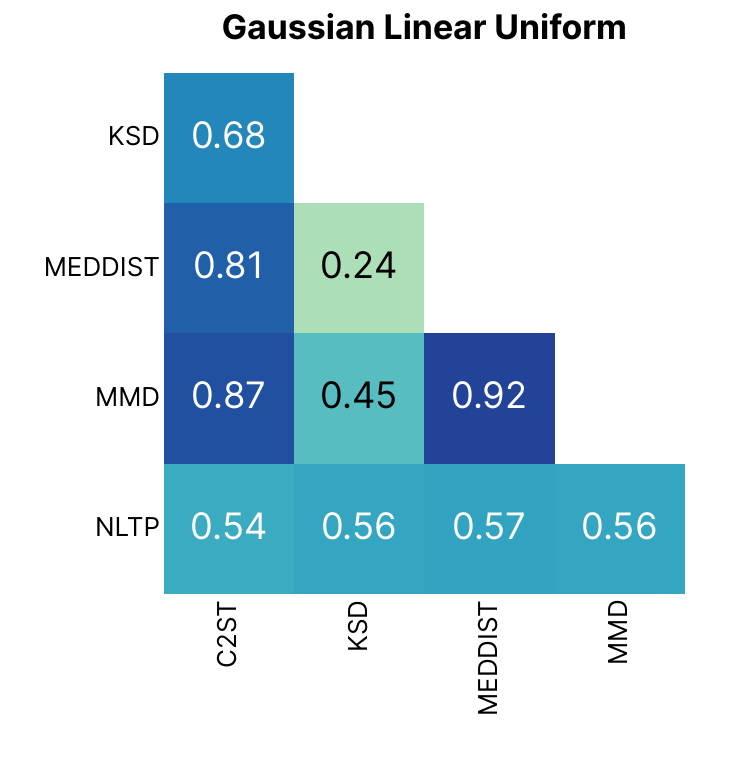}
    \end{subfigure}
    \begin{subfigure}[htbp]{0.28\textwidth}
        \includegraphics[trim=0 0 0 0,clip,width=\textwidth]{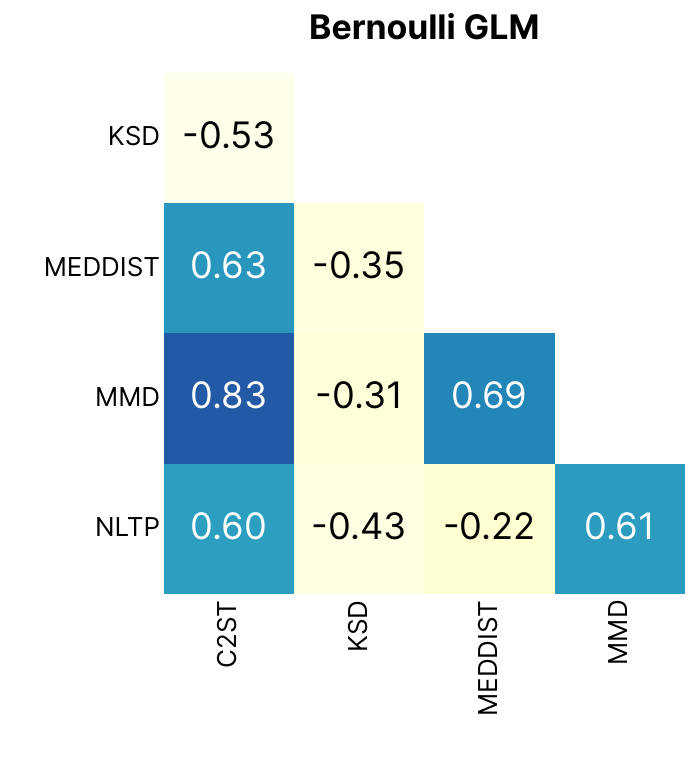}
    \end{subfigure}
    \begin{subfigure}[htbp]{0.28\textwidth}
        \includegraphics[trim=0 0 0 0,clip,width=\textwidth]{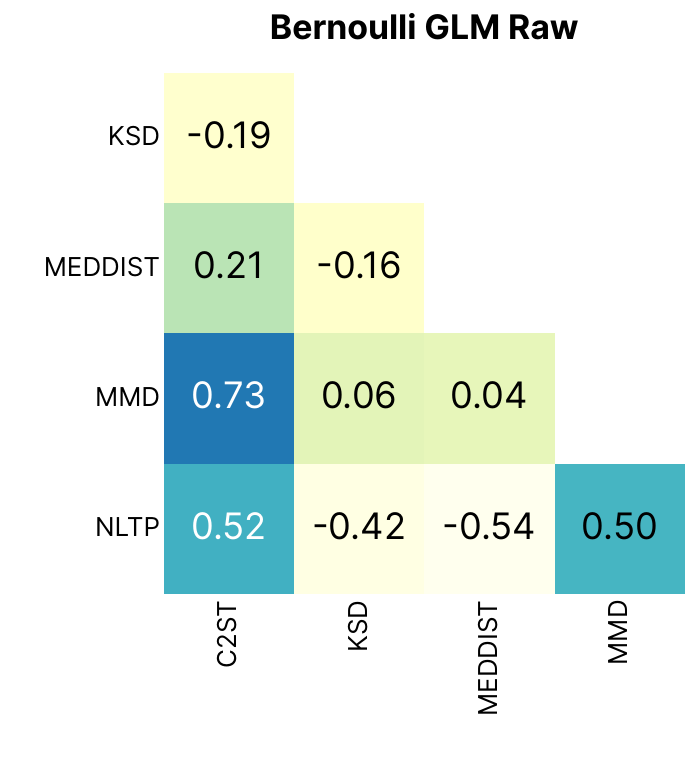}
    \end{subfigure}
    \begin{subfigure}[htbp]{0.28\textwidth}
        \includegraphics[trim=0 0 0 0,clip,width=\textwidth]{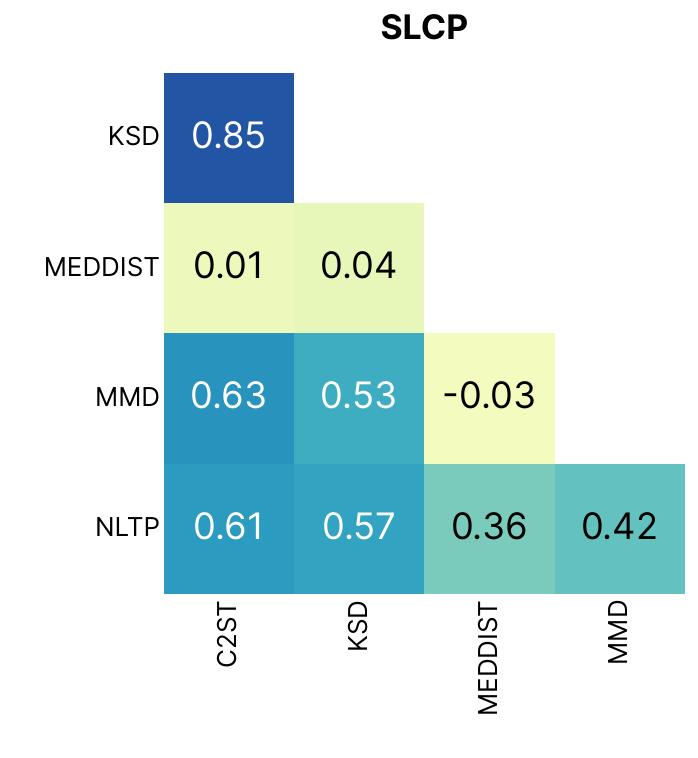}
    \end{subfigure}
    \begin{subfigure}[htbp]{0.28\textwidth}
        \includegraphics[trim=0 0 0 0,clip,width=\textwidth]{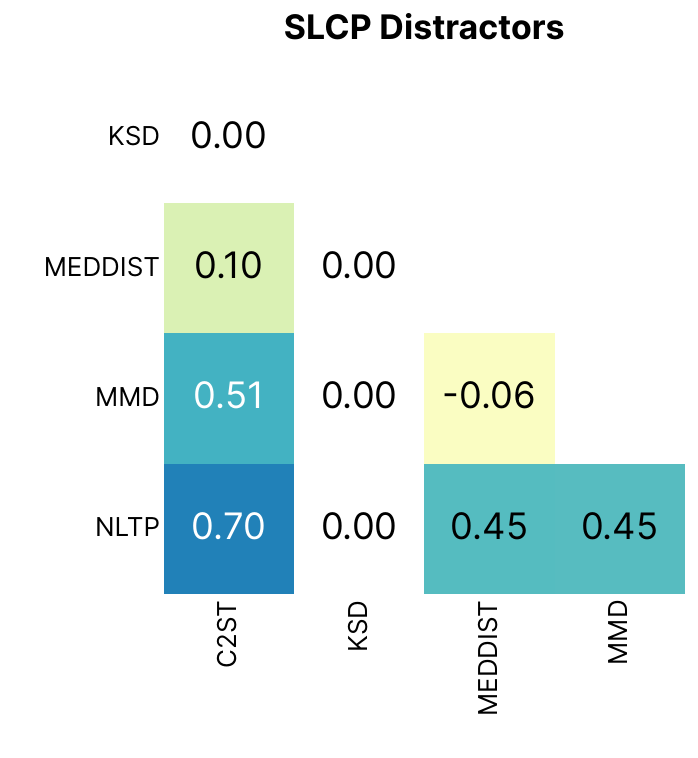}
    \end{subfigure}
    \begin{subfigure}[htbp]{0.28\textwidth}
        \includegraphics[trim=0 0 0 0,clip,width=\textwidth]{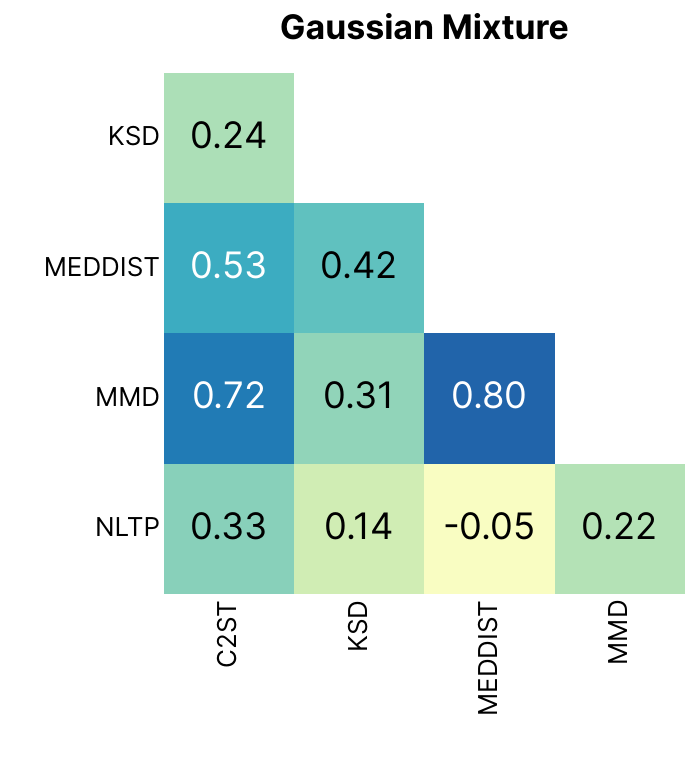}
    \end{subfigure}
    \begin{subfigure}[htbp]{0.28\textwidth}
        \includegraphics[trim=0 0 0 0,clip,width=\textwidth]{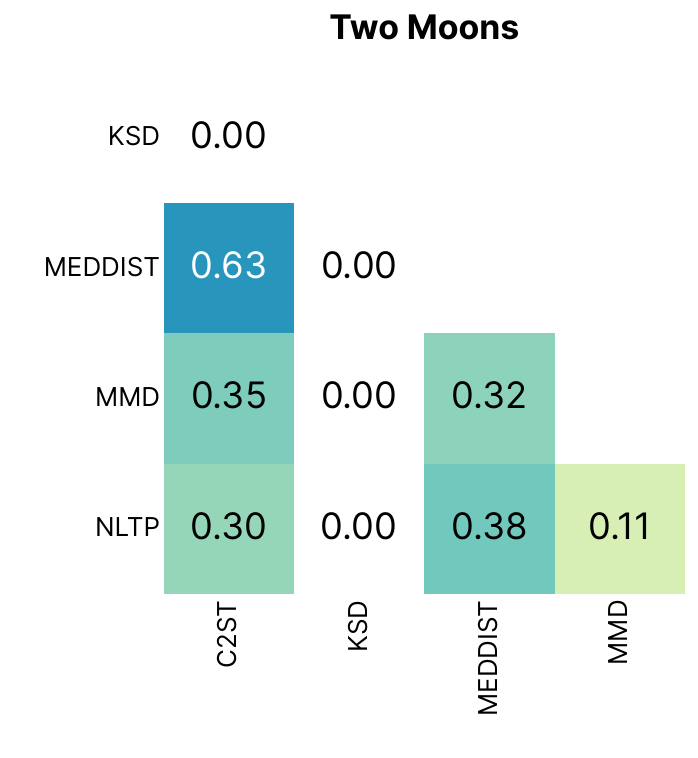}
    \end{subfigure}
    \begin{subfigure}[htbp]{0.28\textwidth}
        \includegraphics[trim=0 0 0 0,clip,width=\textwidth]{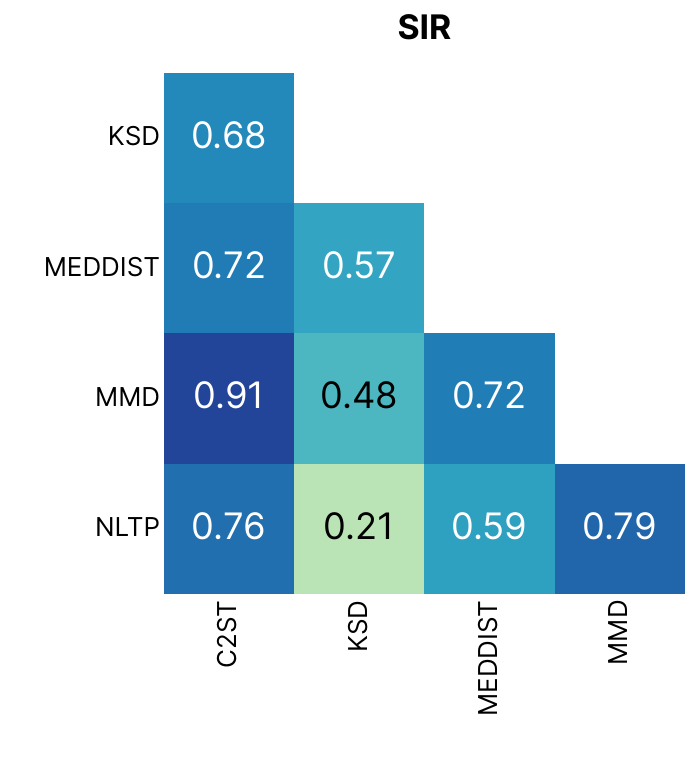}
    \end{subfigure}
    \begin{subfigure}[htbp]{0.28\textwidth}
        \includegraphics[trim=0 0 0 0,clip,width=\textwidth]{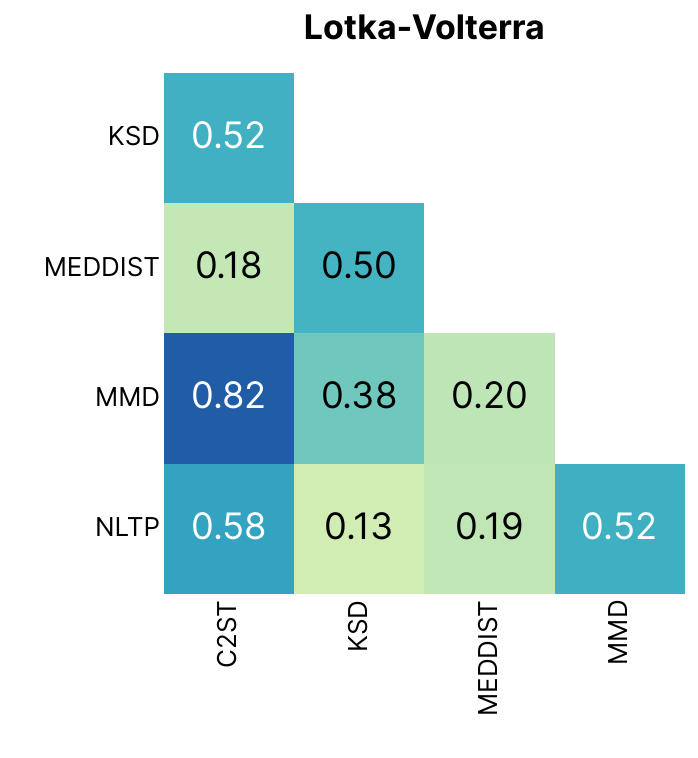}
    \end{subfigure}
    \caption{
        {\bf Correlations between metrics for all tasks.} NLTP is the negative log probability of true parameters. Note that calculation of KSD was numerically unstable when calculating gradients for SLCP Distractors and Two Moons, resulting in correlation of zero for these tasks.
    }
    \label{fig:metrics_correlations}
\end{figure*}

\newpage
\clearpage

%% file: appendix/H_hyperparams.tex
\renewcommand{\thesection}{H}
\section{Hyperparameter Choices}
\label{appendix:hyperparams}

In this section, we address two central questions for any benchmark: (1) how hyperparameters are chosen and (2) how sensitive results are to the respective choices.

Rather than tuning hyperparameters on a per-task basis, we changed hyperparameters on multiple or all tasks at once and selected configurations that worked best across tasks. We wanted to avoid overfitting on individual benchmark tasks and were instead interested in settings that can generalize across multiple tasks. In practice, tuning an algorithm on a given task would typically be impossible,  due to the lack of suitable metrics that can be computed without reference posteriors as well as high computational demands that SBI tasks often have. 

To find good general settings, we performed more than 10 000 individual runs. We explored hyperparameter choices that have not been previously reported, and revealed substantial improvements. The benchmark offers the possibility to systematically compare different choices and design better and more robust SBI algorithms.

\subsection[REJ-ABC]{\ABC{}}

Classical ABC algorithms have crucial hyperparameters, most importantly, the distance metric and acceptance tolerance $\epsilon$. We used our own implementation of \ABC{} as it is straightforward to implement (see \ref{appendix:algorithms:mcabc}). The distance metric was fixed to be the $l_2$-norm for all tasks and we varied different acceptance tolerances $\epsilon$ across tasks on which \ABC{} performed sufficiently well. Our implementation of \ABC{} is quantile based, i.e,. we select a quantile of the samples with the smallest distance to the observed data, which implicitly defines an $\epsilon$. The 10k samples needed for the comparison to the reference posterior samples are then resampled from the selected samples. In order to check whether this resampling significantly impaired performance, we alternatively fit a KDE in order to obtain 10k samples. 

\begin{figure*}[h!]
    \centering
    \begin{subfigure}[htbp]{\textwidth}
        \includegraphics[trim=20 49 0 3,clip,width=\textwidth]{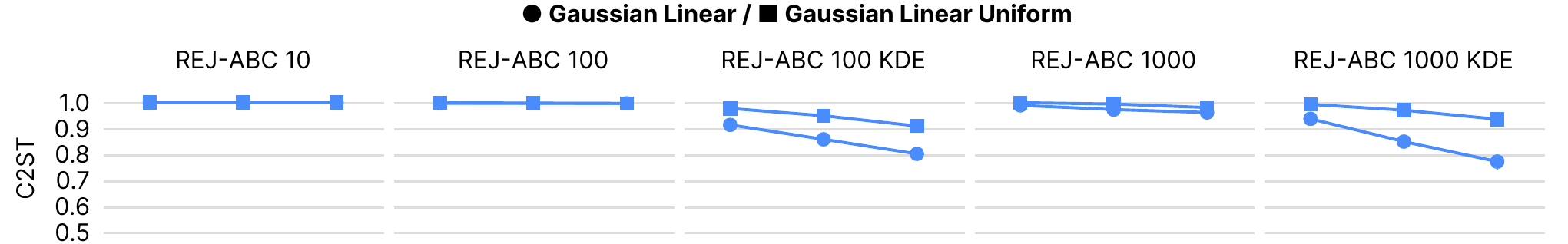}
    \end{subfigure}
    \begin{subfigure}[htbp]{\textwidth}
        \includegraphics[trim=20 49 0 0,clip,width=\textwidth]{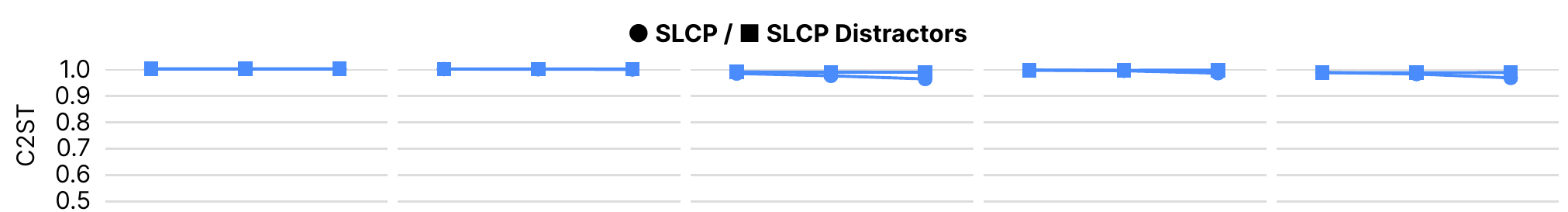}
    \end{subfigure}    
    \begin{subfigure}[htbp]{\textwidth}    
        \includegraphics[trim=20 49 0 0,clip,width=\textwidth]{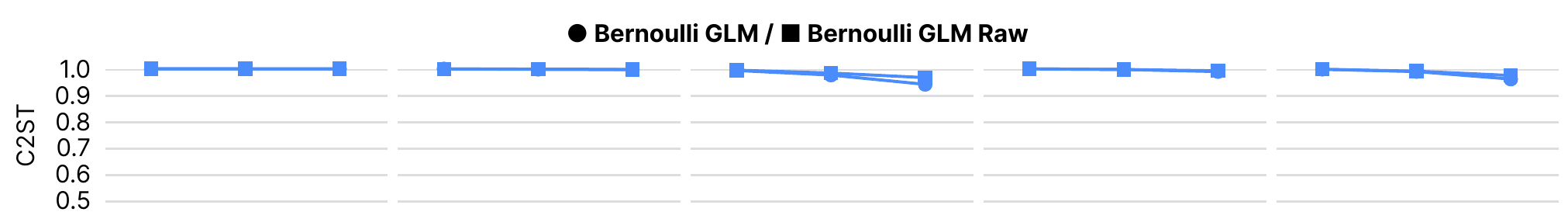}  
    \end{subfigure}
    \begin{subfigure}[htbp]{\textwidth}
        \includegraphics[trim=20 49 0 0,clip,width=\textwidth]{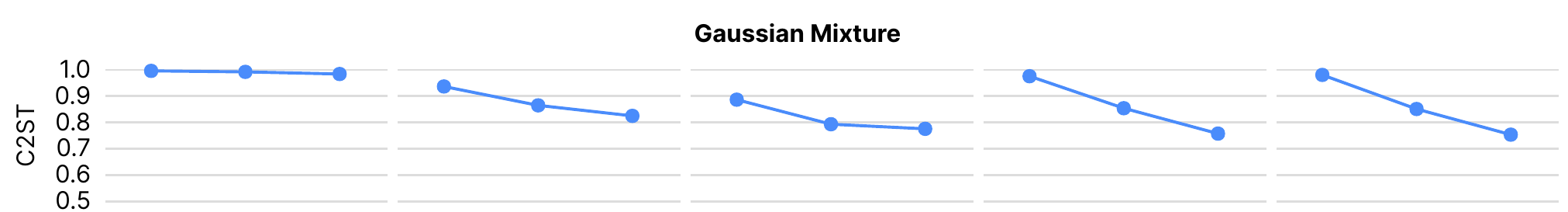}
    \end{subfigure}
    \begin{subfigure}[htbp]{\textwidth}
        \includegraphics[trim=20 49 0 0,clip,width=\textwidth]{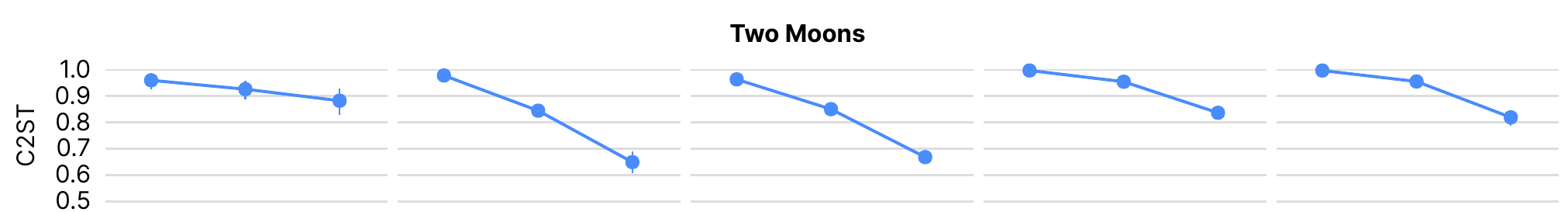}
    \end{subfigure}
    \begin{subfigure}[htbp]{\textwidth}    
        \includegraphics[trim=20 49 0 0,clip,width=\textwidth]{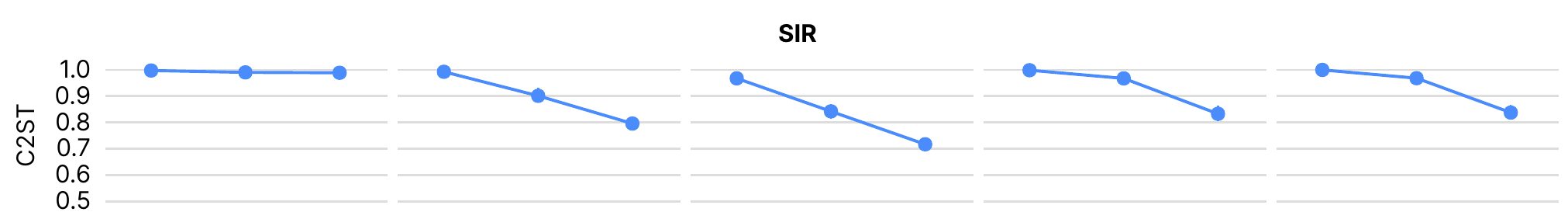}  
    \end{subfigure}
    \begin{subfigure}[htbp]{\textwidth}    
        \includegraphics[trim=20 10 0 0,clip,width=\textwidth]{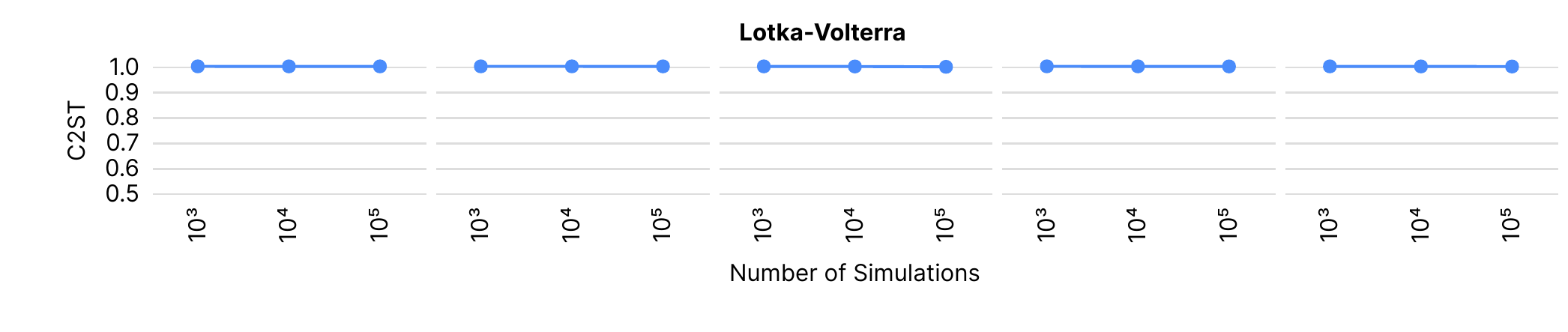}
    \end{subfigure}
    \caption{
        {\bf Hyperparameter selection for \ABC{}.} C2ST performance of different percentile schedules across simulation budgets (columns) for all tasks (rows). Top label for each plot column: number of samples retained, and optional KDE. Across tasks and budgets, the schedule of 0.1, 0.01, 0.001 percentiles, which corresponds to the top 100 samples closest to the observation, performed best. Each data point corresponds to the mean and 95\% confidence interval across 10 observations.
    }
    \label{fig:rej_abc_sweep}
\end{figure*}

Below, we show results for different schedules of quantiles for each simulation budget, e.g., a schedule of 0.1, 0.01, 0.001 corresponds to the 10, 1 and 0.1 percent quantile, or the top 100 samples for each simulation budget. Across tasks and budgets the 0.1, 0.01, 0.001 quantile schedule performed best (\autoref{fig:rej_abc_sweep}). Performance showed improvement by the KDE fit, especially on the Gaussian tasks. We therefore report the version using the top 100 samples and KDE in the main paper.

\newpage
\clearpage
\subsection[SMC-ABC]{\SABC{}}

\SABC{} has several hyperparameters including the population size, the perturbation kernel, the epsilon schedule and the distance metric. In order to ensure that we report the best possible \SABC{} results for a fair comparison, we sweeped over three hyperparameters that are especially critical: the population size, the quantile used to select the epsilon from the distances of the particles of the previous population, and the scaling factor of the covariance of the Gaussian perturbation kernel. The remaining hyperparameters were fixed to values common in the literature: Gaussian perturbation kernel and $l2$-norm distance metric.

Additionally, we compared our implementation against one from the popular pyABC toolbox \citep{klinger2018} to which we refer as versions A and B respectively. We sweeped over these hyperparameters and optionally added a post-hoc KDE fit for drawing the samples needed for two-sample based performance metrics.

\begin{figure*}[h!]
    \centering
    \begin{subfigure}[htbp]{\textwidth}
        \includegraphics[trim=20 49 0 130,clip,width=\textwidth]{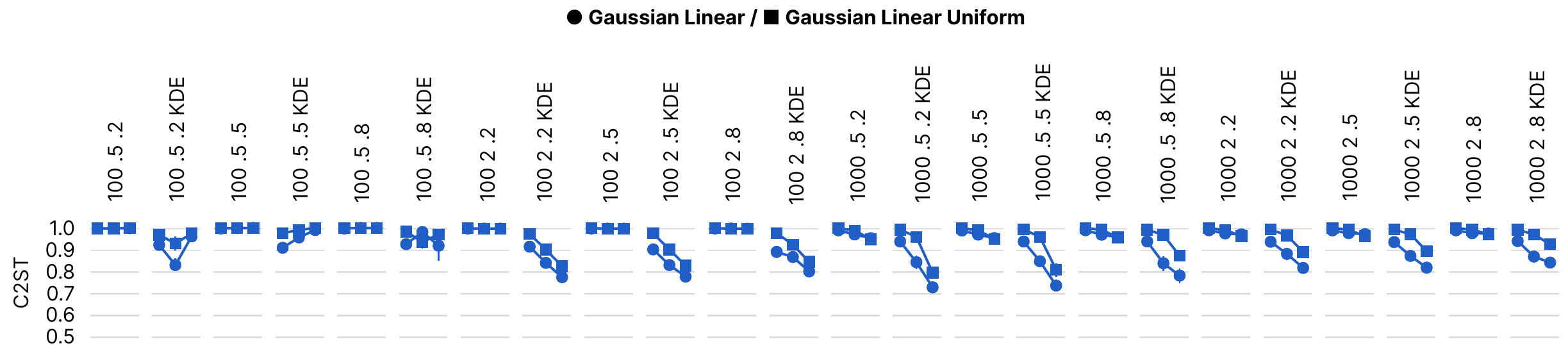}
    \end{subfigure}
    \begin{subfigure}[htbp]{\textwidth}
        \includegraphics[trim=20 49 0 0,clip,width=\textwidth]{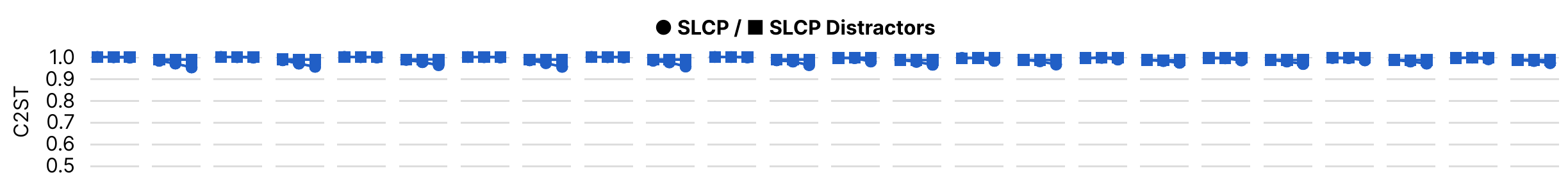}
    \end{subfigure}    
    \begin{subfigure}[htbp]{\textwidth}    
        \includegraphics[trim=20 49 0 0,clip,width=\textwidth]{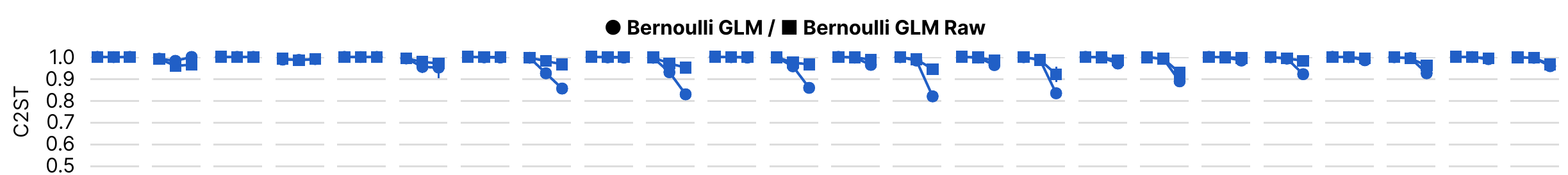}  
    \end{subfigure}
    \begin{subfigure}[htbp]{\textwidth}
        \includegraphics[trim=20 49 0 0,clip,width=\textwidth]{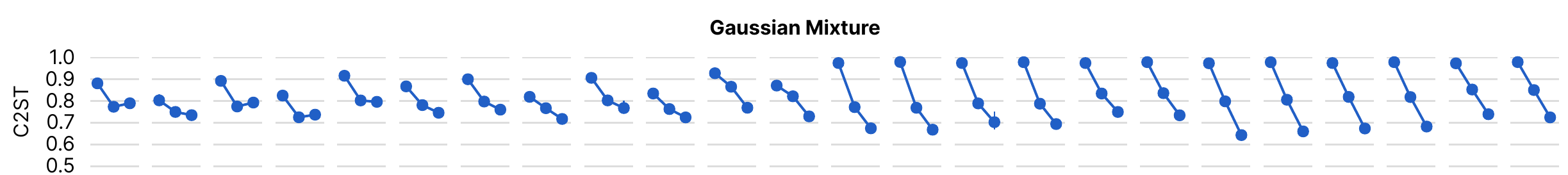}
    \end{subfigure}
    \begin{subfigure}[htbp]{\textwidth}
        \includegraphics[trim=20 49 0 0,clip,width=\textwidth]{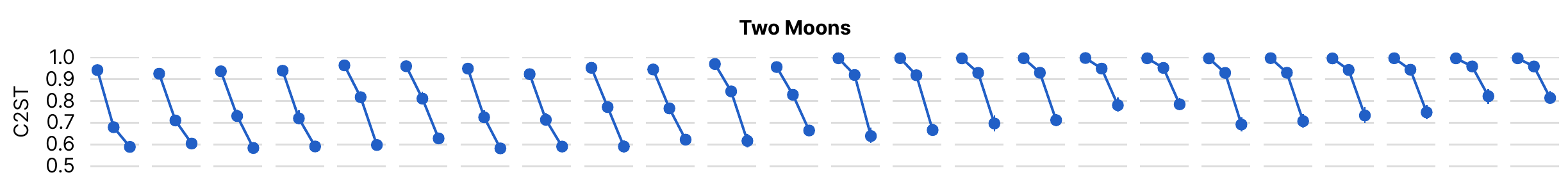}
    \end{subfigure}
    \begin{subfigure}[htbp]{\textwidth}    
        \includegraphics[trim=20 49 0 0,clip,width=\textwidth]{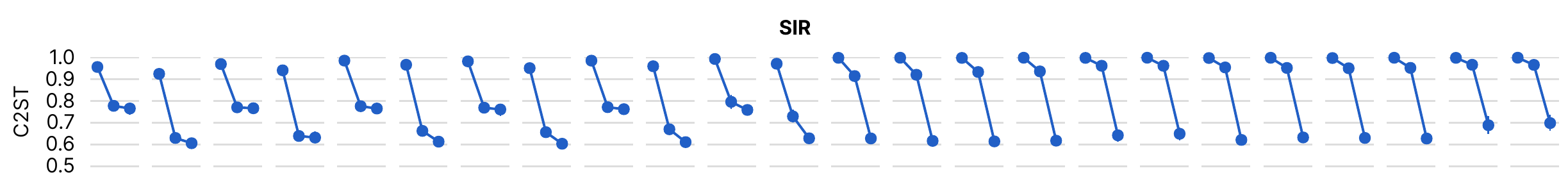}  
    \end{subfigure}
    \begin{subfigure}[htbp]{\textwidth}    
        \includegraphics[trim=20 10 0 0,clip,width=\textwidth]{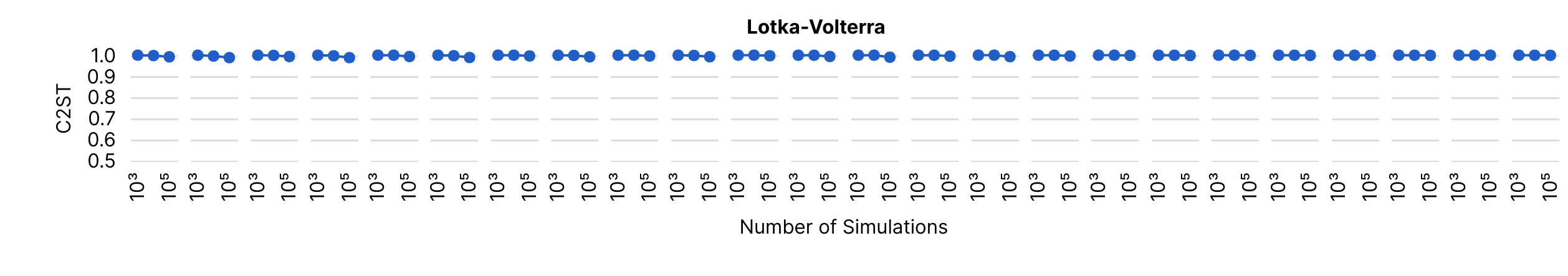}
    \end{subfigure}
    \caption{
        {\bf Hyperparameter selection for \SABC{} with our implementation}. Top label for each plot column: population size, kernel covariance scale, epsilon quantile/epsilon-decay parameter, and optional KDE. Each data point corresponds to the mean and 95\% confidence interval across 10 observations.
    }
    \label{fig:smc_abc_sweep_sbi}
\end{figure*}

\begin{figure*}[h!]
    \centering
    \begin{subfigure}[htbp]{\textwidth}
        \includegraphics[trim=20 49 0 30,clip,width=\textwidth]{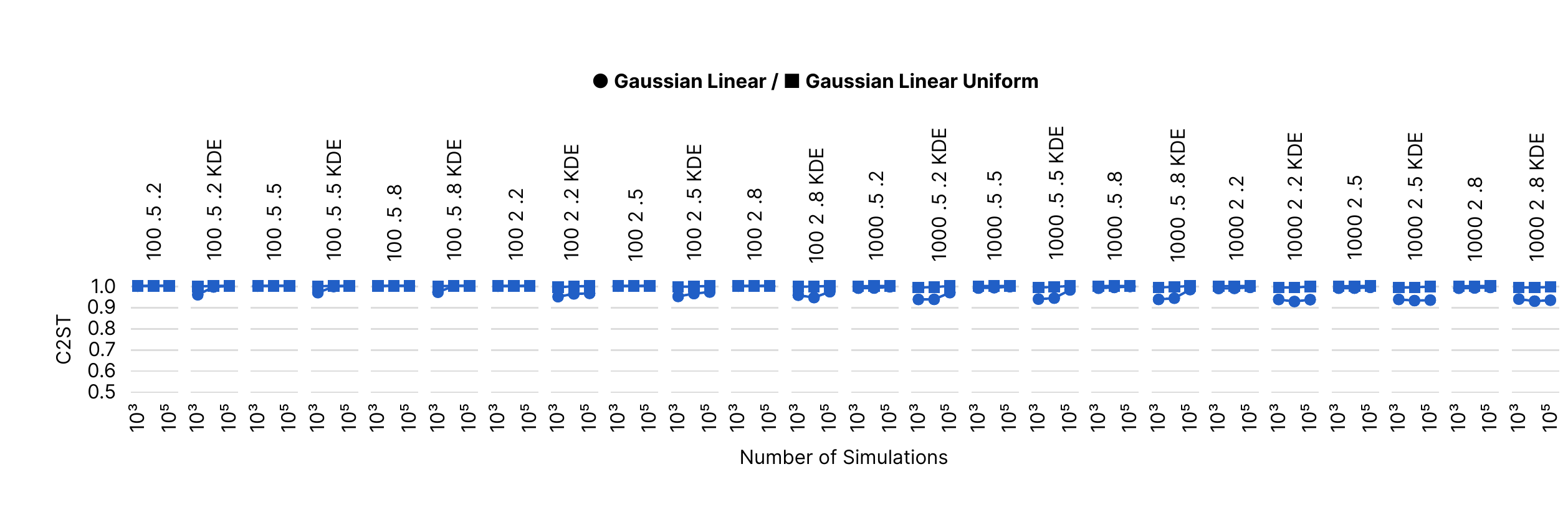}
    \end{subfigure}
    \begin{subfigure}[htbp]{\textwidth}
        \includegraphics[trim=20 49 0 0,clip,width=\textwidth]{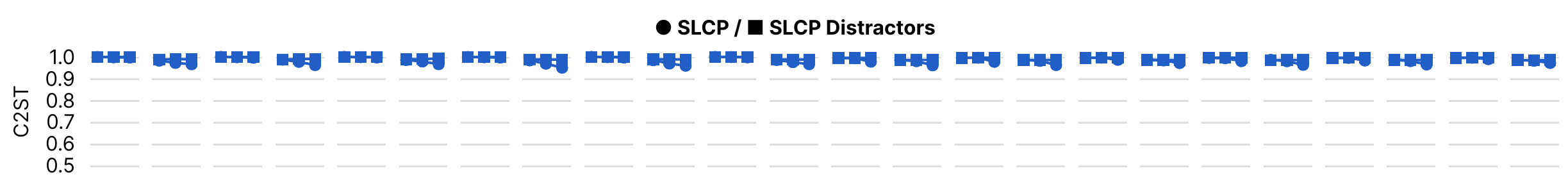}
    \end{subfigure}    
    \begin{subfigure}[htbp]{\textwidth}    
        \includegraphics[trim=20 49 0 0,clip,width=\textwidth]{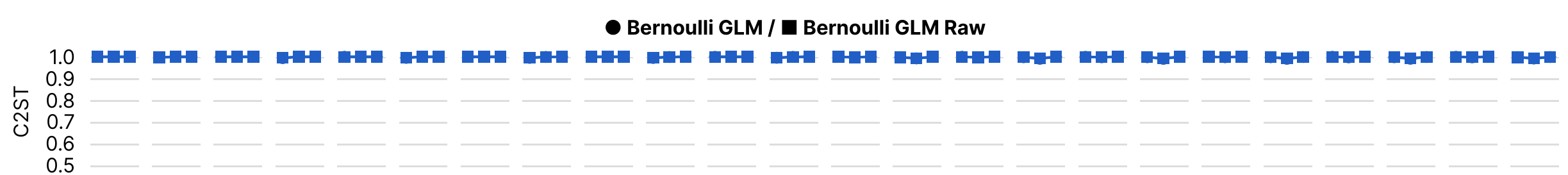}  
    \end{subfigure}
    \begin{subfigure}[htbp]{\textwidth}
        \includegraphics[trim=20 49 0 0,clip,width=\textwidth]{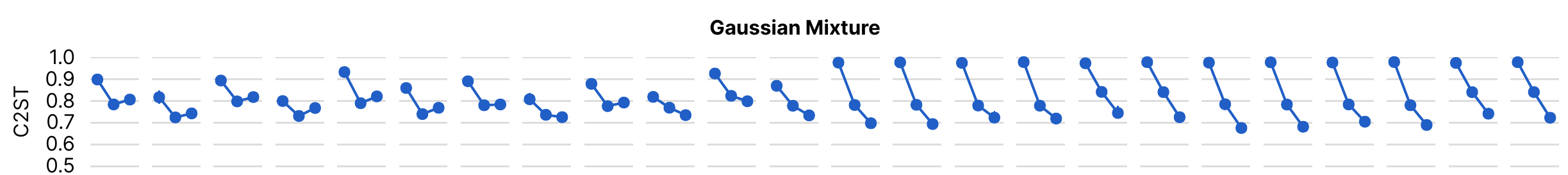}
    \end{subfigure}
    \begin{subfigure}[htbp]{\textwidth}
        \includegraphics[trim=20 49 0 0,clip,width=\textwidth]{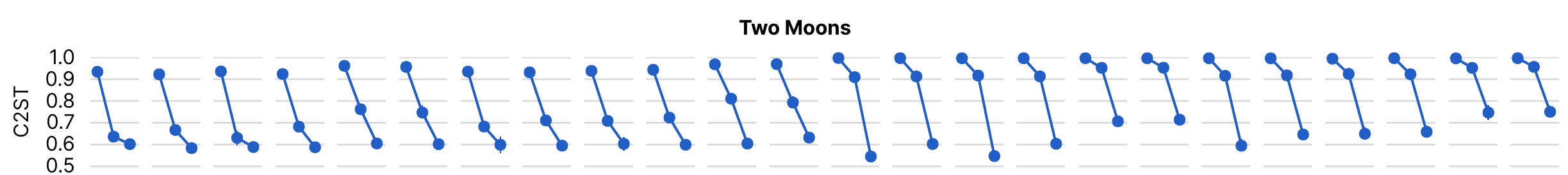}
    \end{subfigure}
    \begin{subfigure}[htbp]{\textwidth}    
        \includegraphics[trim=20 49 0 0,clip,width=\textwidth]{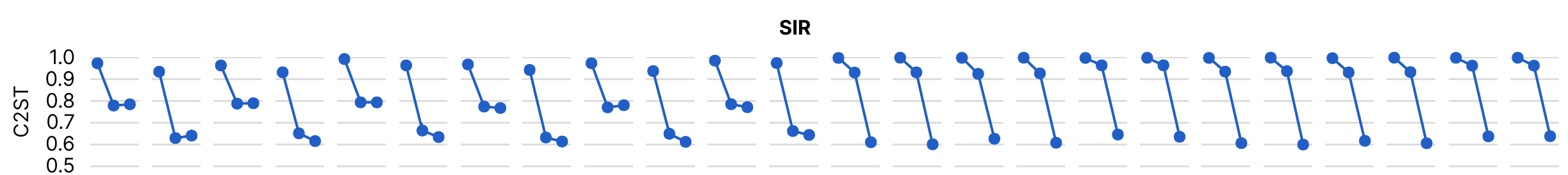}  
    \end{subfigure}
    \begin{subfigure}[htbp]{\textwidth}    
        \includegraphics[trim=20 10 0 0,clip,width=\textwidth]{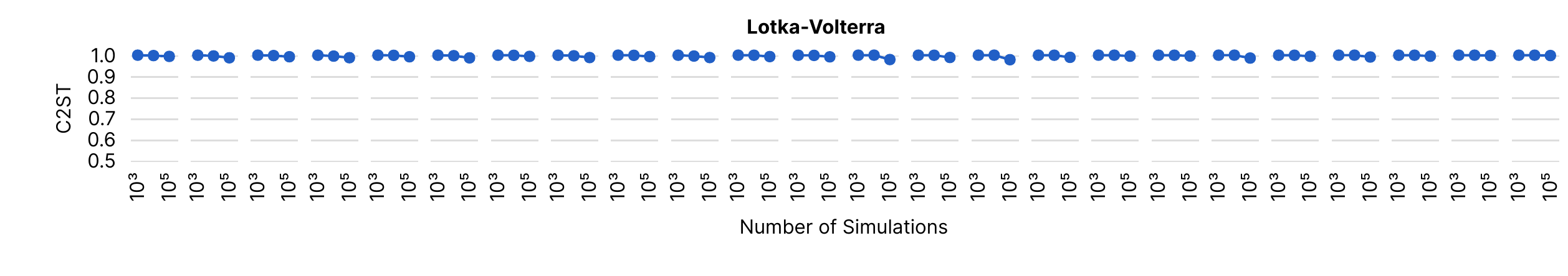}
    \end{subfigure}
    \caption{
        {\bf Hyperparameter selection for \SABC{}. with \texttt{pyABC} implementation}. Top label for each plot column: population size, kernel covariance scale, epsilon quantile/epsilon-decay parameter, and optional KDE. Each data point corresponds to the mean and 95\% confidence interval across 10 observations.
    }
    \label{fig:smc_abc_sweep_pyabc}
\end{figure*}

Overall, the parameter setting with a population size of 100, a kernel covariance scale of 0.5, and an epsilon quantile 0.2 performed best. Although the results of the two different implementations were qualitatively very similar (compare \autoref{fig:smc_abc_sweep_sbi} and \autoref{fig:smc_abc_sweep_pyabc}, respectively), version A was slightly better on the Gaussian tasks. Although we tried to match the implementations and the exact settings, there are small differences between the two, which might explain the difference in the results: Implementation B constructs the Gaussian perturbation kernel using kernel density estimation on the weighted samples of the previous population, whereas A constructs it using the mean and covariance estimated from samples from the previous population. The latter could be advantageous in case of a Gaussian-like (high-dimensional) posterior (Gaussian Mixture and Gaussian linear task) and disadvantageous in a non-Gaussian-like posteriors (e.g., Two Moons). We decided to report results for \SABC{} in the main paper using implementation A (ours) with population size 100 for simulation budgets 1k and 10k, and population size 1000 for simulation budget 100k, a kernel covariance scale of 0.5, and epsilon quantile 0.2. This choice of kernel covariance scale is different from recommendations in the literature \citep{sisson2007, beaumont2009}. We only found very small performance differences for different scales and note that our choice is in line with the recommendation of the \texttt{pyABC} toolbox \citep{pyabcKernel}, i.e., using a scale between 0 and 1. Performance showed improvement by the KDE fit, especially on the Gaussian tasks. We therefore report the version with KDE in the main paper.

\newpage
\clearpage

\newpage
\clearpage

\newpage
\clearpage

\subsection[MCMC for (S)NLE and (S)NRE]{MCMC for \XNLE{} and \XNRE{}}

\XNLE{} and \XNRE{} both rely on MCMC sampling, which has several hyperparameters. In line with \citet{papamakarios2019a} and \citet{durkan2020}, we used Slice Sampling \citep{neal2003}. However, we modified the MCMC schemes used in these papers and obtained significant improvements in performance and speed.

\textbf{Number of chains and initialization.} While \citet{papamakarios2019a,durkan2020} used a single chain with axis-aligned updates, we found that on tasks with multi-modal posteriors, it can be essential to run multiple MCMC chains in order to sample all modes. Performance on Two Moons, for example, was poor with a single chain, since usually only one of the crescent shapes was sampled. Rather than initialising chains by drawing initial locations from the prior, we found the resampling scheme as described in \ref{appendix:algorithms:nle} to work better for initialisation, and used 100 chains instead of a single one.

\textbf{Transformation of variables.} When implementing MCMC, it is common advice to transform problems to have unbounded support \citep{hogg2017}, although this has not been discussed in SBI papers or implemented in accompanying code. We found that without this transformation, MCMC sampling could get stuck in endless loops, e.g., on the Lotka-Volterra task. Apart from the transformation to unbounded space, we found z-scoring of parameters and data to be crucial for some tasks.

\textbf{Vectorization of MCMC sampling}. We reimplemented Slice Sampling so that all chains could perform likelihood evaluations in parallel. Evaluating likelihoods, e.g., in the case of \XNLE{}, requires passes through a flow-based density estimator, which is significantly faster when batched. This allowed us to sample all chains in parallel rather than sequentially and yielded huge speed-ups: For example, \SNLE{} on Gaussian Linear took more than 36 hours on average for 100k simulations without vectorization, and less than 2 hours with vectorization.

\subsection[Density estimator for (S)NLE]{Density estimator for \XNLE{}}

Approaches based on neural networks (NN) tend to have many hyperparameters, including the concrete type of NN architecture and hyperparameters for training. We strove to keep our choices close to \cite{durkan2020}, which are the defaults in the toolbox we used \citep[\texttt{sbi},][]{sbi}. 

\begin{figure*}[h!]
    \centering
    \begin{subfigure}[htbp]{\textwidth}
        \includegraphics[trim=20 49 0 3,clip,width=\textwidth]{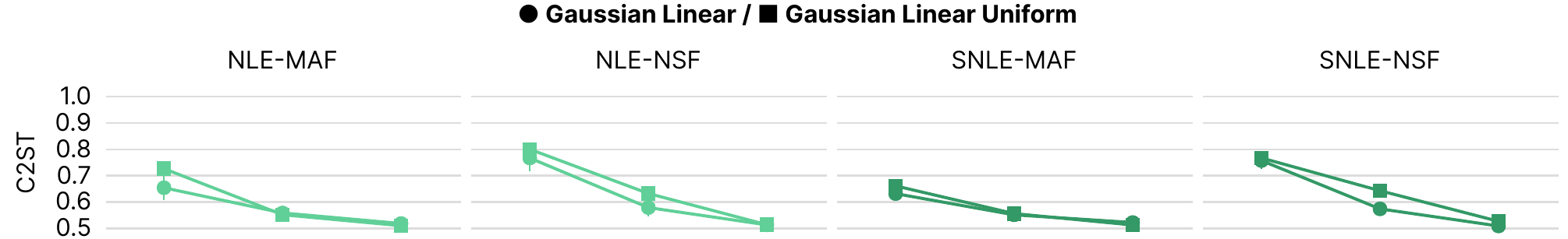}
    \end{subfigure}
    \begin{subfigure}[htbp]{\textwidth}
        \includegraphics[trim=20 49 0 0,clip,width=\textwidth]{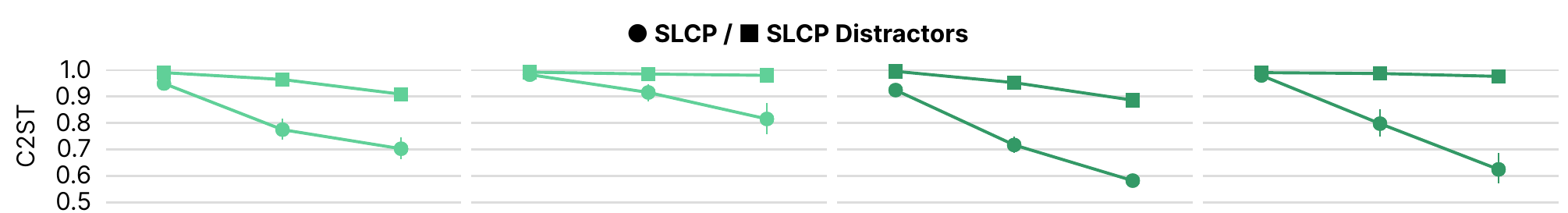}
    \end{subfigure}    
    \begin{subfigure}[htbp]{\textwidth}    
        \includegraphics[trim=20 49 0 0,clip,width=\textwidth]{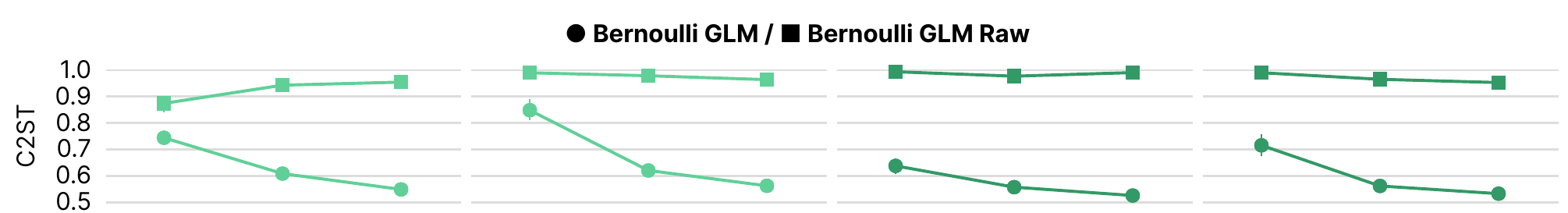}  
    \end{subfigure}
    \begin{subfigure}[htbp]{\textwidth}
        \includegraphics[trim=20 49 0 0,clip,width=\textwidth]{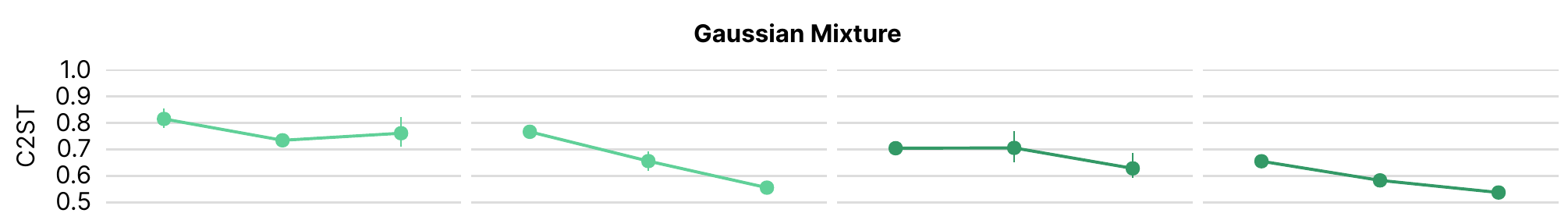}
    \end{subfigure}
    \begin{subfigure}[htbp]{\textwidth}
        \includegraphics[trim=20 49 0 0,clip,width=\textwidth]{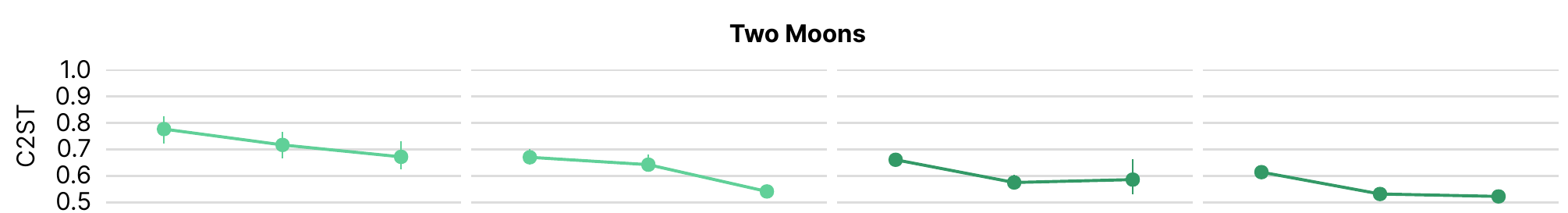}
    \end{subfigure}
    \begin{subfigure}[htbp]{\textwidth}    
        \includegraphics[trim=20 49 0 0,clip,width=\textwidth]{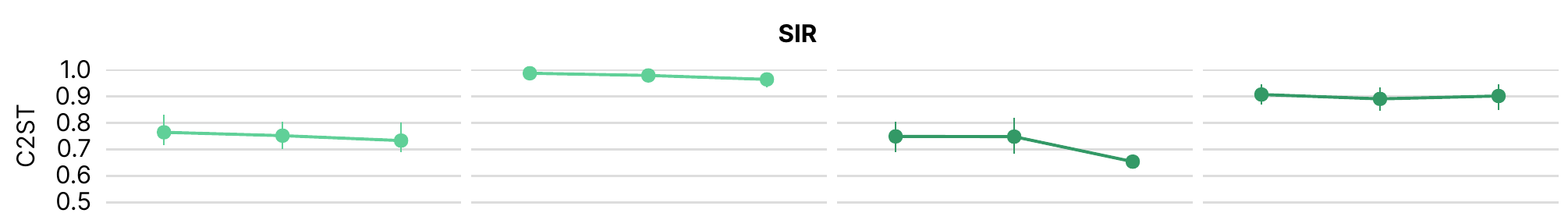}  
    \end{subfigure}
    \begin{subfigure}[htbp]{\textwidth}    
        \includegraphics[trim=20 10 0 0,clip,width=\textwidth]{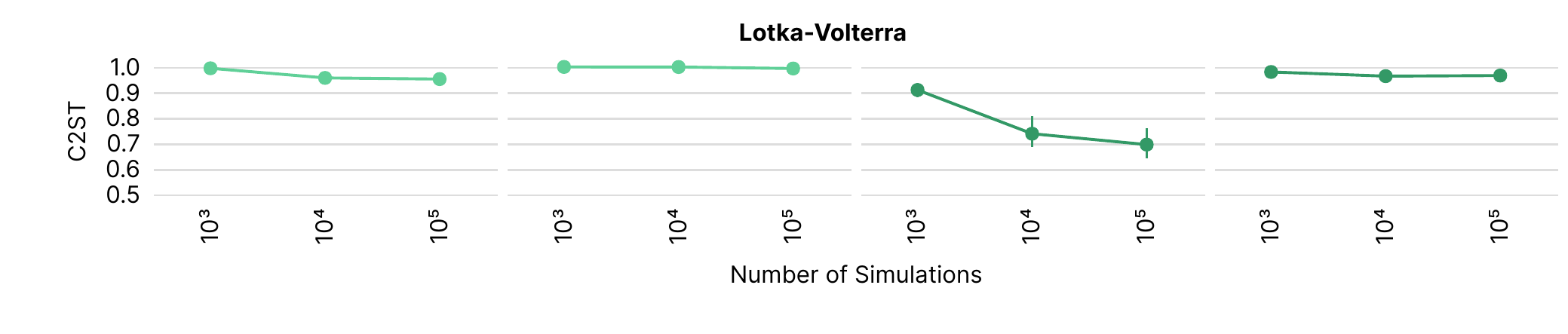}  
    \end{subfigure}
    \caption{
        {\bf Density estimator selection for \XNLE{}.} Performance of \XNLE{} in terms of C2ST across tasks using MAFs or NSFs for density estimation. Considering all tasks, NSFs generally performed worse, e.g., using NSFs significantly reduced performance on SIR and Lotka-Volterra, indicating that the added flexibility of NSFs was not needed for \XNLE{}. We thus reported performance using MAFs in the main paper. Each data point corresponds to the mean and 95\% confidence interval across 10 observations.
    }
    \label{fig:snle_maf_nsf}
\end{figure*}

While \cite{papamakarios2019a,durkan2020} used Masked Autoregressive Flows \citep[MAFs,][]{papamakarios2017} for density estimation, we explored how results change when using Neural Spline Flows \citep[NSFs,][]{durkan2019neural} for density estimation. These results are shown in \autoref{fig:snle_maf_nsf}.

%
% (S)NLE MAF versus NSF
%

\newpage
\clearpage
\subsection[Density estimator for (S)NPE]{Density estimator for \XNPE{}}

\begin{figure*}[h!]
    \centering
    \begin{subfigure}[htbp]{\textwidth}
        \includegraphics[trim=20 49 0 3,clip,width=\textwidth]{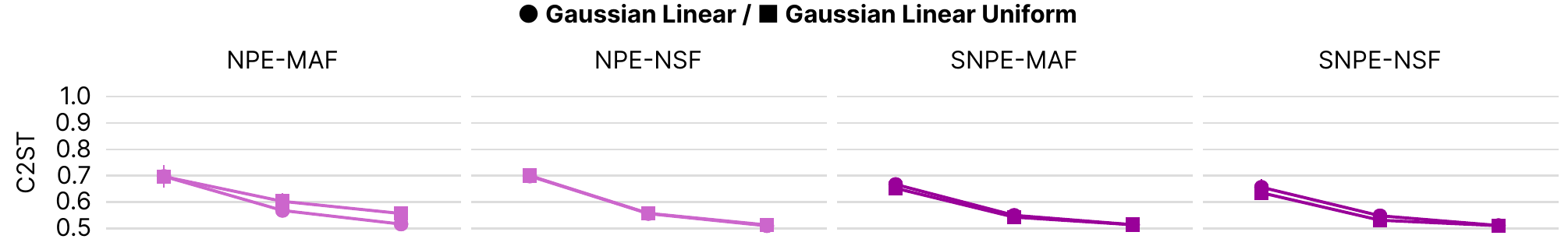}
    \end{subfigure}
    \begin{subfigure}[htbp]{\textwidth}
        \includegraphics[trim=20 49 0 0,clip,width=\textwidth]{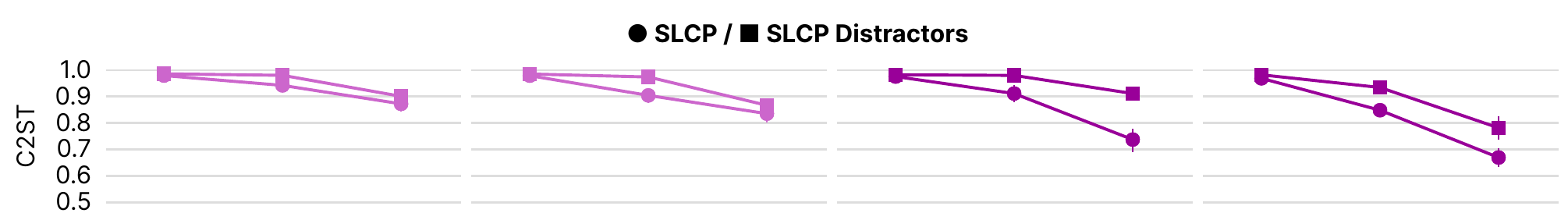}
    \end{subfigure}    
    \begin{subfigure}[htbp]{\textwidth}    
        \includegraphics[trim=20 49 0 0,clip,width=\textwidth]{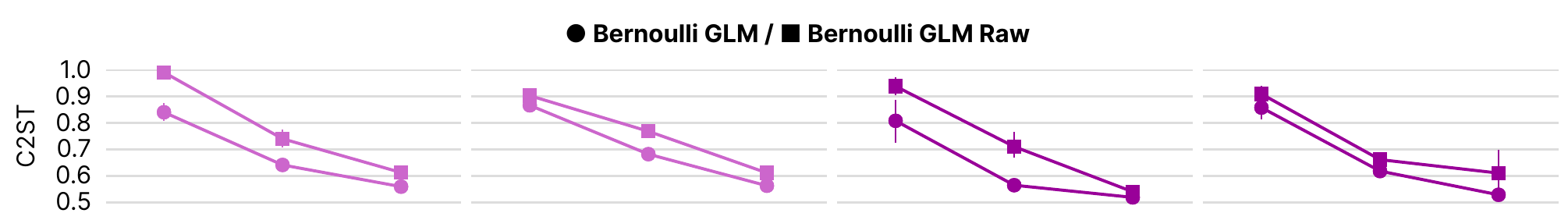}  
    \end{subfigure}
    \begin{subfigure}[htbp]{\textwidth}
        \includegraphics[trim=20 49 0 0,clip,width=\textwidth]{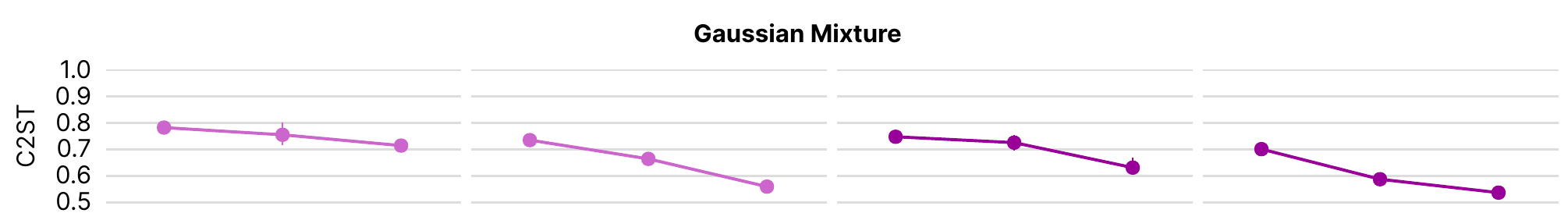}
    \end{subfigure}
    \begin{subfigure}[htbp]{\textwidth}
        \includegraphics[trim=20 49 0 0,clip,width=\textwidth]{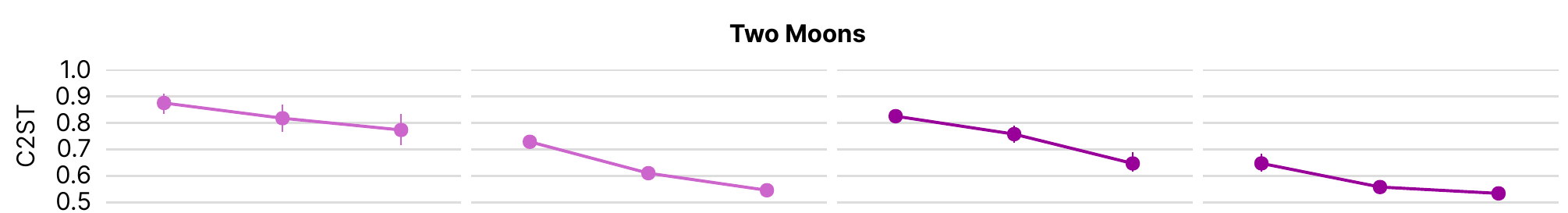}
    \end{subfigure}
    \begin{subfigure}[htbp]{\textwidth}    
        \includegraphics[trim=20 49 0 0,clip,width=\textwidth]{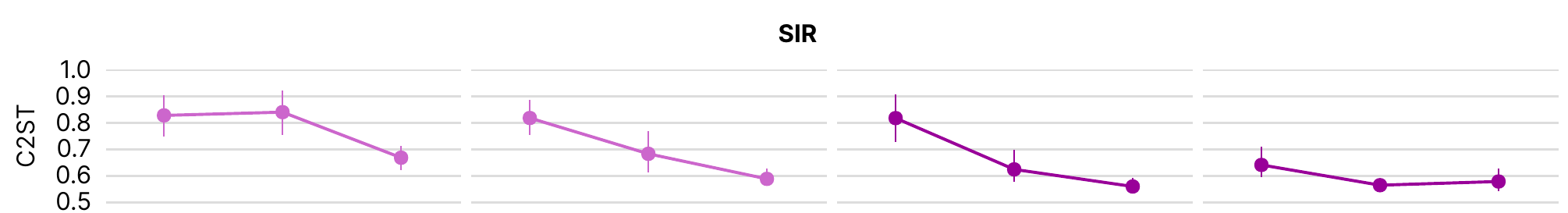}  
    \end{subfigure}
    \begin{subfigure}[htbp]{\textwidth}    
        \includegraphics[trim=20 10 0 0,clip,width=\textwidth]{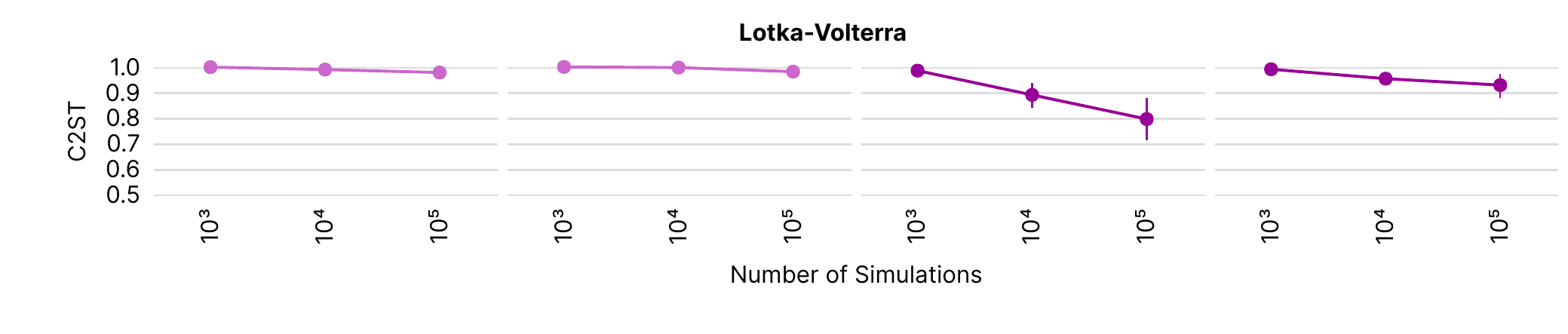}
    \end{subfigure}
    \caption{
        {\bf Density estimator selection for \XNPE{}.} Performance of \XNPE{} in terms of C2ST across tasks using MAFs or NSFs for density estimation. Considering all tasks, NSFs generally performed better, especially on Gaussian Mixture, Two Moons, and SIR. We thus reported performance using NSFs in the main paper. Each data point corresponds to the mean and 95\% confidence interval across 10 observations.
    }
    \label{fig:snpe_maf_nsf}
\end{figure*}

We performed the analogous experiments for \XNPE{} as for \XNLE{}: Here, we found NSFs to increase performance relative to MAFs (\autoref{fig:snpe_maf_nsf}). When directly estimating the posterior distribution, especially on tasks with complex multi-modal structure like Two Moons or SLCP, the additional flexibility offered by NSFs improved performance. With NSFs, artifacts from density transformation that were visible e.g. in Two Moons posteriors, vanished. To our knowledge, results on \XNPE{} with NSFs have not been previously published.

%
% (S)NPE MAF versus NSF
%

\newpage
\clearpage
\subsection[Density estimator for (S)NRE]{Classifier choice for \XNRE{}}

\begin{figure*}[h!]
    \centering
    \begin{subfigure}[htbp]{\textwidth}
        \includegraphics[trim=20 49 0 3,clip,width=\textwidth]{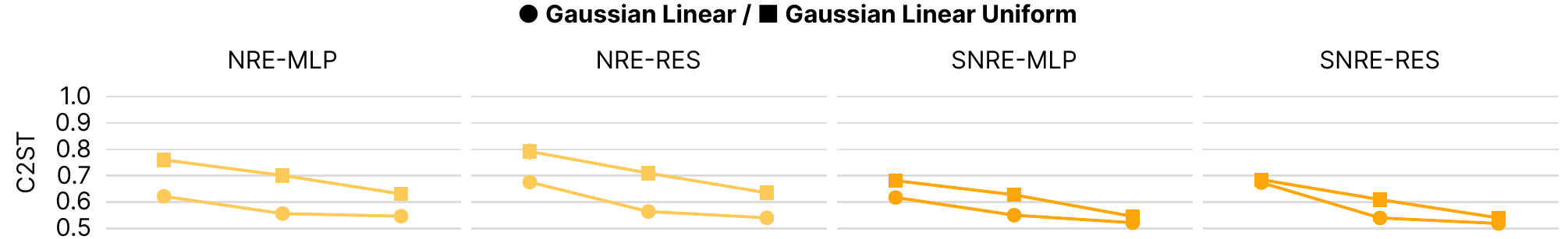}
    \end{subfigure}
    \begin{subfigure}[htbp]{\textwidth}
        \includegraphics[trim=20 49 0 0,clip,width=\textwidth]{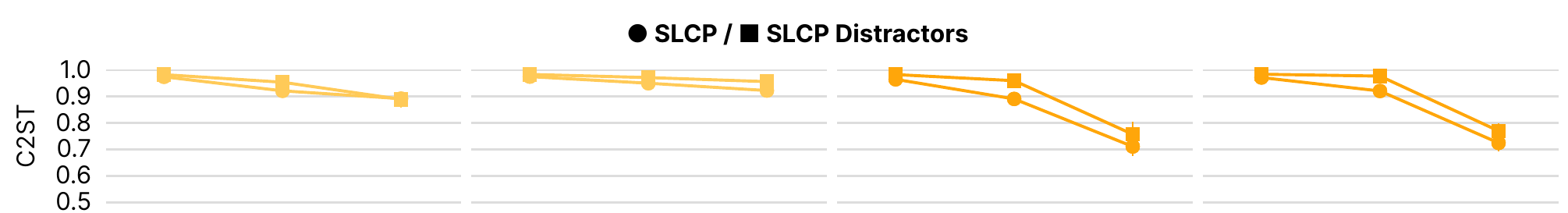}
    \end{subfigure}    
    \begin{subfigure}[htbp]{\textwidth}    
        \includegraphics[trim=20 49 0 0,clip,width=\textwidth]{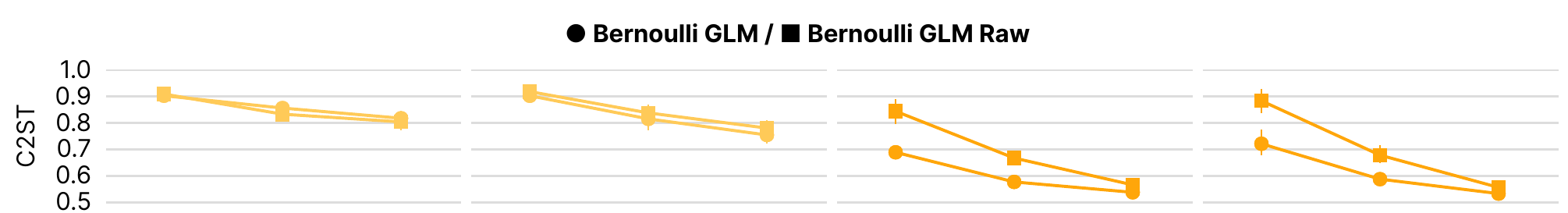}  
    \end{subfigure}
    \begin{subfigure}[htbp]{\textwidth}
        \includegraphics[trim=20 49 0 0,clip,width=\textwidth]{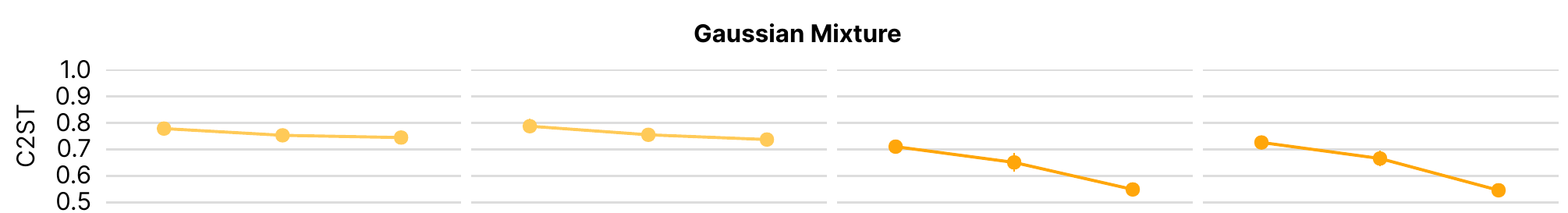}
    \end{subfigure}
    \begin{subfigure}[htbp]{\textwidth}
        \includegraphics[trim=20 49 0 0,clip,width=\textwidth]{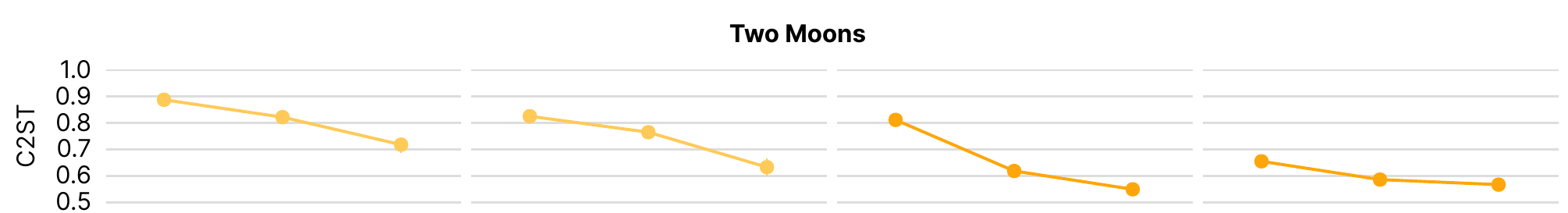}
    \end{subfigure}
    \begin{subfigure}[htbp]{\textwidth}    
        \includegraphics[trim=20 49 0 0,clip,width=\textwidth]{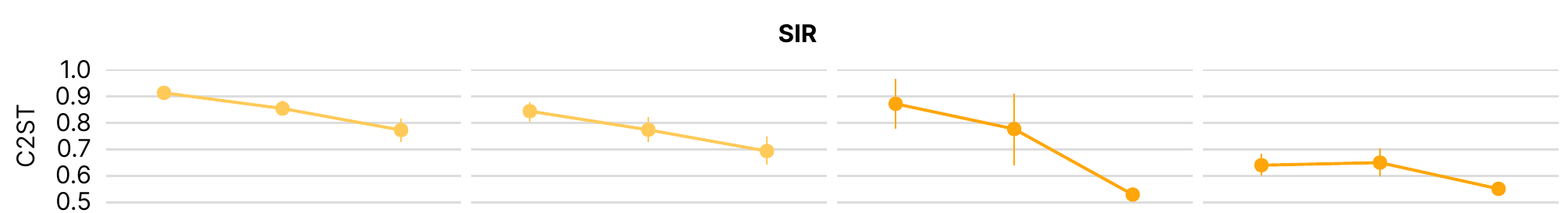}  
    \end{subfigure}
    \begin{subfigure}[htbp]{\textwidth}    
        \includegraphics[trim=20 10 0 0,clip,width=\textwidth]{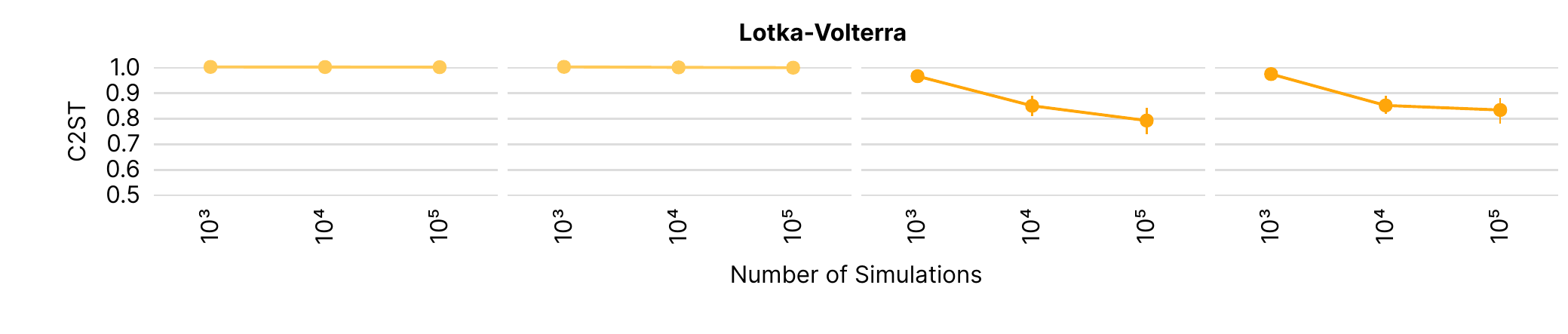}
    \end{subfigure}
    \caption{
        {\bf Classifier architecture for \XNRE{}.} Performance of \XNRE{} in terms of C2ST across tasks using MLPs or ResNets for classification. Considering all tasks, ResNets generally performed better, especially on Two Moons and SIR. We thus reported performance using ResNets in the main paper. Each data point corresponds to the mean and 95\% confidence interval across 10 observations.
    }
    \label{fig:snre_mlp_res}
\end{figure*}

For \XNRE{}, we compared two different choices of classifier architectures: an MLP and a ResNet architecture, as described in \ref{appendix:algorithms:nre}. While results were similar for most tasks (\autoref{fig:snre_mlp_res}), we decided to use the ResNet architecture in the main paper due to the better performance on Two Moons and SIR for low to medium simulation budgets.

%
% (S)NRE MLP versus RES
%

%% file: appendix/M_metrics.tex
\renewcommand{\thesection}{M}
\section{Metrics}
\label{appendix:metrics}

\input{appendix/metrics/true_parameters}
\input{appendix/metrics/simulation_based_calibration}
\input{appendix/metrics/median_distance}

\input{appendix/metrics/mmd}
\input{appendix/metrics/c2st}
\input{appendix/metrics/ksd}

%% file: appendix/metrics/true_parameters.tex
\subsection[Negative log probability of true parameters (NLTP)]{Negative log probability of $\btheta_o$ (NLTP)}
\label{appendix:metrics:true_parameters}

In simulation-based inference, the average negative log likelihood of true parameters $-\E[\log q(\btheta_o|\bx_o)]$ (NLTP) is commonly reported as a performance metric in the literature \citep{papamakarios2016,durkan2018,papamakarios2019a,greenberg2019,hermans2019,durkan2020}. An attractive property of this metric is that the access to the ground-truth posterior is not required.

It is important to point out, however, that calculating this metric on a single or small number of pairs $(\btheta_o, \bx_o)$ is problematic. To illustrate the issue, consider the following example (as discussed in \cite{talts2018}): Consider $\theta \sim \mathcal{N}(0, 1^2), x | \theta \sim \mathcal{N}(\theta, {1}^2)$, and a single pair $(\btheta_o, \bx_o)$ with $\theta_o=0$ and an implausible (but possible) $x_o=2.1$. In this case, the true posterior is $\mathcal{N}(\theta|1.05, 0.5^2)$ under which the $\theta_o$ has low probability since it is more than two standard deviations away from the posterior mean. If an algorithm fitted a wrong posterior, e.g., by overestimating the standard deviation as 1 instead of 0.5, the probability of $\theta_o$ under the estimated posterior would be higher than under the true posterior. 

Therefore, a large number of pairs $(\btheta_o, \bx_o)$ should be used. Indeed, in the limit of infinite number of pairs $(\btheta_o, \bx_o)$, the metric converges to a $\KL$:
$$
\begin{aligned}
&\E_{\btheta_o \sim p(\btheta)} \E_{\bx_o \sim p(\bx|\btheta_o)}\big[ -\log q(\btheta_o|\bx_o) \big] \\
=\ &\E_{\bx_o \sim p(\bx), \btheta_o \sim p(\btheta|\bx_o)}\big[ -\log q(\btheta_o|\bx_o) \big] \\
=\ &\E_{\bx_o \sim p(\bx), \btheta_o \sim p(\btheta|\bx_o)}\big[ -\log q(\btheta_o|\bx_o) + \log p(\btheta_o|\bx_o) \big] - \E_{\bx_o \sim p(\bx), \btheta_o \sim p(\btheta|\bx_o)}\big[\log p(\btheta_o|\bx_o) \big] \\
=\ &\E_{\bx_o \sim p(\bx)} \KL(p(\btheta | \bx_o) || q(\btheta|\bx_o)) + \E_{\bx_o \sim p(\bx)} \mathbb{H}(p(\btheta | \bx_o))
\end{aligned}
$$

The first term in the final equation is the average $\KL$ between true and approximate posteriors over all observations $\bx_o$ that can be generated when sampling parameters $\btheta_o$ from the prior. The second term, the entropy term, would be the same for all algorithms compared.

In the context of this benchmark, we decided against using the probability of $\btheta_o$ as a metric: For all algorithms that are not amortized (all but one), evaluating posteriors at different $\bx_o$ would require rerunning inference. As the computational requirements for running the benchmark at 10 observations per task are already high, running tasks for hundreds of observations would become prohibitively expensive.

%% file: appendix/metrics/simulation_based_calibration.tex
\subsection{Simulation-based calibration (SBC)}

In simulation-based calibration (SBC), samples $\btheta^\prime$ are drawn from the data-averaged posterior, i.e., the posterior obtained by running inference for many observations. When the posterior approximation is exact, $\btheta^\prime$ is distributed according to the prior \citep{talts2018}.

Let us briefly illustrate this: In SBC, we draw $\btheta \sim p(\btheta), \bx \sim p(\bx|\btheta), \btheta^\prime \sim q(\btheta^\prime|\bx)$, which implies a joint distribution $\pi(\btheta, \bx, \btheta^\prime) = p(\btheta) p(\bx|\btheta) q(\btheta^\prime|\bx)$. The marginal $\pi(\btheta^\prime)$ is then:
$$
\begin{aligned}
\pi(\btheta^\prime) = \int\int p(\btheta) p(\bx|\btheta) q(\btheta^\prime|\bx) \d\bx \d\btheta = \int\int p(\btheta, \bx) q(\btheta^\prime|\bx) \d\bx \d\btheta = \int p(\bx)\ q(\btheta^\prime|\bx) \d\bx.
\end{aligned}
$$
If the approximate posterior is the true posterior, the marginal on $\btheta^\prime$ is equal to the prior: If $q(\btheta^\prime|\bx) = p(\btheta^\prime|\bx)$, then $\pi(\btheta^\prime)= \int p(\bx, \btheta') \d\bx = p(\btheta^\prime)$, i.e., one can set up a consistency test that is based on the distribution of $\btheta^\prime$ samples. \cite{talts2018} do this by using frequentist tests per dimension.  

Note that SBC as described above is merely a consistency check. For example, if the approximate posterior were the prior, a calibration test as described above would not be able to detect this. This is a realistic failure mode in simulation-based inference. It could happen with rejection ABC in the limit $\epsilon \rightarrow \infty$, or when learned summary statistics have no information about $\btheta$. One way around this is issue is proposed in \cite{prangle2014diag}, who propose to restrict observations to a subset of all possible $\mathcal{X}$.

SBC is similar to the average negative log likelihood of true parameters described above, in that inference needs to be carried out for many observations generated by sampling from the prior. Running inference for hundreds of observations would become prohibitively expensive in terms of compute for most algorithms, which is why we do not rely on SBC in the benchmark.

%% file: appendix/metrics/median_distance.tex
\subsection{Median distance (MEDDIST)}
\label{appendix:median_distance}

Posterior predictive checks (PPCs) use the posterior predictive distribution to predict new data, $\bx^{\prime} \sim p(\bx^{\prime}|\bx_o) = \int p(\bx^{\prime}|\btheta) q(\btheta|\bx_o) \d\btheta$. The observed data $\bx_o$ should look plausible under the posterior predictive distribution (\cite{gelman2004}, chapter 6). A particular PPC, used for example in \cite{papamakarios2019a, greenberg2019, durkan2020}, is to assess the median L2 distance between $N^{\prime}$ posterior predictive samples $\bx^{\prime}$ and $\bx_o$. The median is used since the mean would be more sensitive to outliers. 

In the benchmark, we refer to this metric as median distance (MEDDIST) and drew $N^{\prime} = 10000$ samples from each posterior predictive distribution to compute it. In contrast with other metrics considered here, the median distance is computed in the space of data $\bx$ and requires additional simulations (which could be expensive, depending on the simulator). The median distance should be considered a mere check rather than a metric and it does not necessarily test the structure of the estimated posterior.

%% file: appendix/metrics/mmd.tex
\subsection{Maximum Mean Discrepancy (MMD)}

Maximum Mean Discrepancy (MMD) is an Integral Probability Metric (IPM). Linear and quadratic time estimates for using MMD as a two-sample test were derived in \cite{gretton2012}. MMD has been commonly used in the SBI literature with Gaussian kernels \citep{papamakarios2019a,greenberg2019,hermans2019}, setting a single length-scale hyperparameter by using a median heuristic \citep{ramdas2015}. We follow the same procedure, i.e., use Gaussian kernels with length-scale determined by the median heuristic on reference samples. MMDs are calculated using 10k samples from reference and approximate posteriors.

If simple kernels are used to compare distributions with complex, multimodal structure, distinct distributions can be mapped to nearby mean embeddings, resulting in low test power. On SLCP and Two Moons, for example, we found a translation-invariant kernel to be limiting, since it cannot adapt to the local structure (see \suppfig{mmd}). This is reflected in the low correlation of MMD and C2ST (\suppfig{metrics_correlations}). We emphasize that these issues are strictly related to simple kernels with hyperparameters commonly used in the literature. Posteriors of the Two Moons task have a structure similar to the blobs example of \citet{liu2020}, who argue for using learned kernels to overcome the aforementioned problem.

%% file: appendix/metrics/c2st.tex
\subsection{Classifier-based tests (C2ST)}

In classifier-based testing, a classifier is trained to distinguish samples of the true posterior $p(\btheta|\bx_o)$ from samples of the estimated posterior $q(\btheta|\bx_o)$. If the samples are indistinguishable, the classification performance should be at chance level, 0.5. Practical use and properties of classifier-based 2-sample testing (C2ST) are discussed in \citet{lopez-paz2018} \citep[see][for examples in the context of SBI]{gutmann2018likelihood,dalmasso2019a}.

To compute C2ST, we trained a two-layer neural network with 10 times as many ReLU units as the dimensionality of parameters, and optimize with Adam \citep{kingma2014adam}. Classifiers were trained on 10k z-scored samples from reference and approximate posterior each. Classification accuracy was reported using 5-fold cross-validation.  

%% file: appendix/metrics/ksd.tex
\subsection{Kernelized Stein Discrepancy (KSD)}

Kernelized Stein Discrepancy (KSD) is a 1-sample goodness-of-fit test proposed independently by \citet{chwialkowski2016} and \citet{liu2016}. KSD tests samples from algorithms against the gradient of unnormalized true posterior density, $\nabla_{\btheta}\ \tilde{p}(\btheta|\bx_o)$. We used KSD with Gaussian kernels, setting the length-scale through the median heuristic, and 10k samples from each algorithm.

%% file: appendix/R_runtimes.tex
\renewcommand{\thesection}{R}
\section{Runtimes}
\label{appendix:runtimes}

In applications of SBI, simulations are commonly assumed to be the dominant cost. In order to make the benchmark feasible at this scale, we focused on simple simulators and optimized runtimes, e.g. we developed a new package bridging \texttt{DifferentialEquations.jl} \citep{rackauckas2017,bezanson2017julia} and \texttt{PyTorch} \citep{paszke2019} so that generating simulations for all implemented tasks is extremely fast. This differs from many cases in practice, where the runtime costs for an algorithm are often negligible compared to the cost of simulations. Having said that, algorithms show significant differences in runtime costs, which we measured and report here. 

We recorded runtimes for all algorithms on all tasks. In principle, runtimes could be reduced by employing multi-CPU architectures, however, we decided for the single CPU setup to accurately compare runtimes across all algorithms and tasks. We did not employ GPUs for training neural-networks (NN). This is because the type of NNs used in the algorithms currently in the benchmark do not benefit much from GPU versus CPU training (e.g., no CNN architecture, rather shallow and narrow networks). In fact, running \SNPE{} on SLCP using a GeForce GTX 1080 showed slightly longer runtimes than on CPU, due to the added overhead resulting from copying data back and forth to the device. Therefore, it was more economical and comparable to run the benchmark on CPUs.

All neural network-based algorithms were run on single 3.6 GHz CPU cores of AWS C5-instances. ABC algorithms were run on single CPU cores of an internal cluster with 2.4 GHz CPUs. We observed a difference in runtimes of less than 100ms when running ABC algorithms on the same hardware as used for neural network-based algorithms.

\begin{figure*}[h!]
    \centering
    \begin{subfigure}[htbp]{\textwidth}
        \includegraphics[trim=20 49 0 3,clip,width=\textwidth]{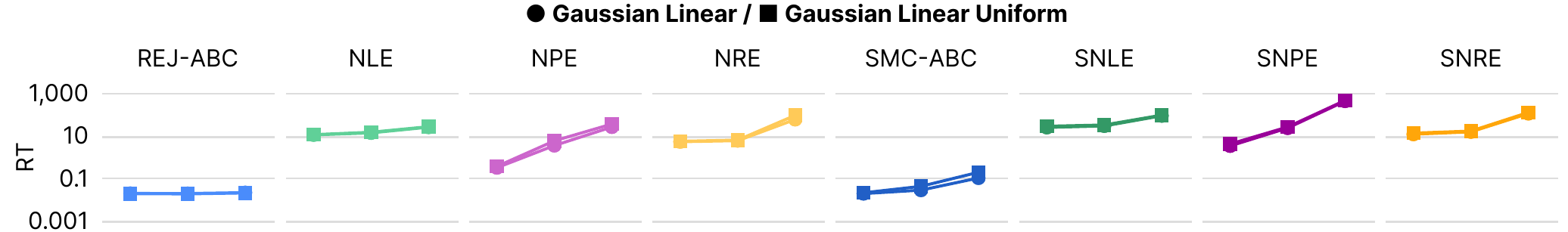}
    \end{subfigure}
    \begin{subfigure}[htbp]{\textwidth}
        \includegraphics[trim=20 49 0 0,clip,width=\textwidth]{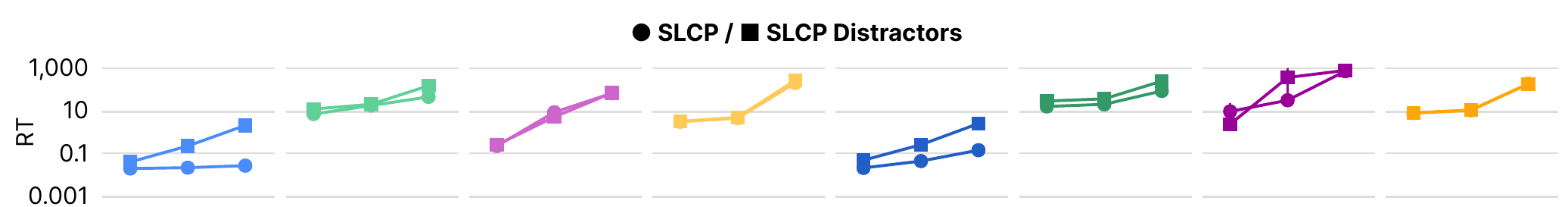}
    \end{subfigure}    
    \begin{subfigure}[htbp]{\textwidth}    
        \includegraphics[trim=20 49 0 0,clip,width=\textwidth]{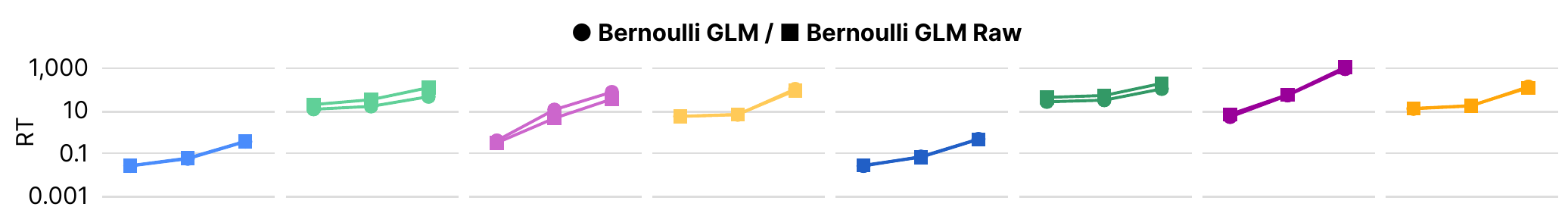}  
    \end{subfigure}
    \begin{subfigure}[htbp]{\textwidth}
        \includegraphics[trim=20 49 0 0,clip,width=\textwidth]{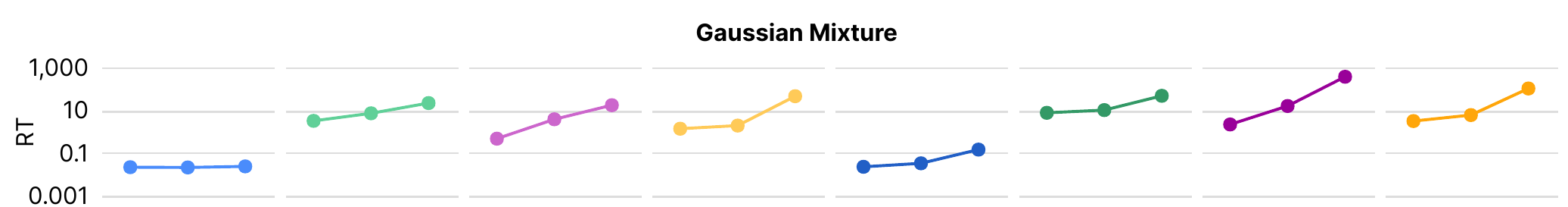}
    \end{subfigure}
    \begin{subfigure}[htbp]{\textwidth}
        \includegraphics[trim=20 49 0 0,clip,width=\textwidth]{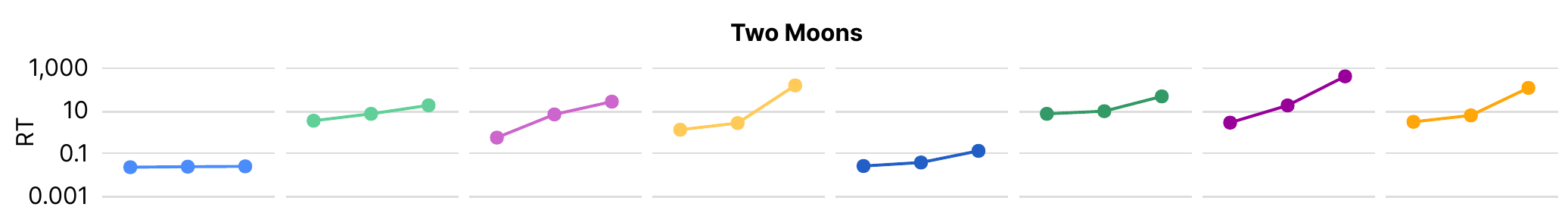}
    \end{subfigure}
    \begin{subfigure}[htbp]{\textwidth}    
        \includegraphics[trim=20 49 0 0,clip,width=\textwidth]{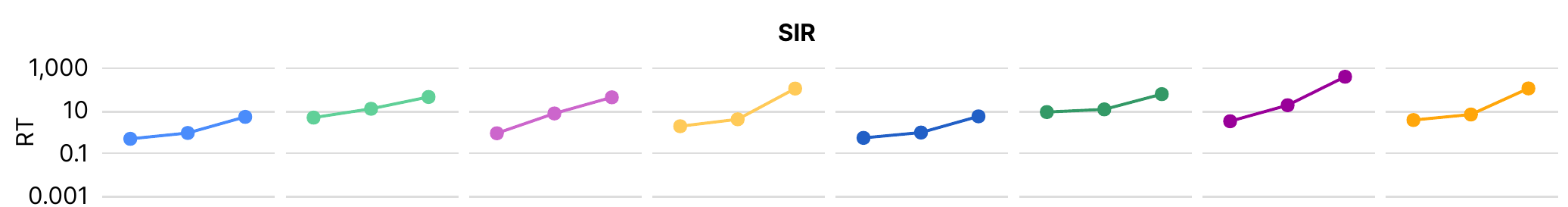}  
    \end{subfigure}
    \begin{subfigure}[htbp]{\textwidth}    
        \includegraphics[trim=20 10 0 0,clip,width=\textwidth]{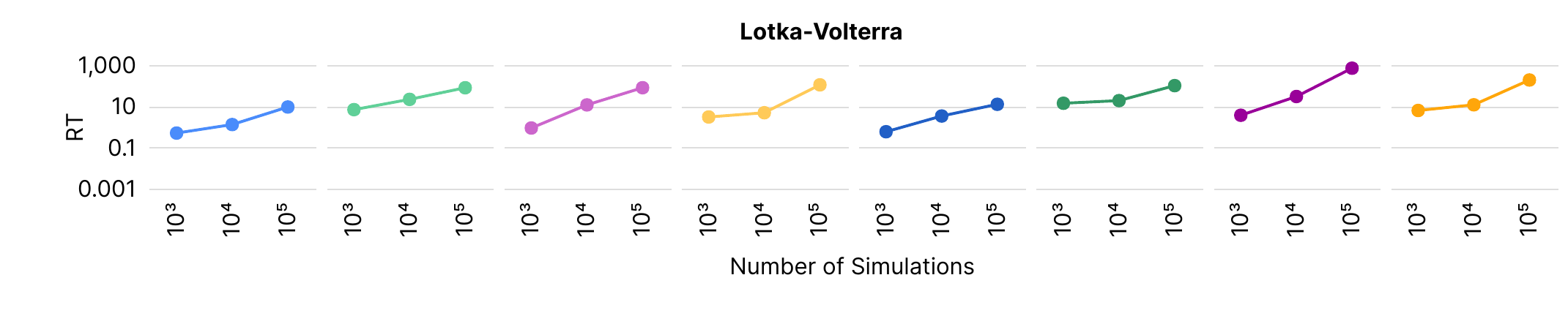}  
    \end{subfigure}
    \caption{
        {\bf Runtime on benchmark tasks.} Runtime of \ABC{}, \SABC{}, \NLE{}, \SNLE{}, \NPE{}, \SNPE{}, \NRE{}, \SNRE{} in minutes, for 10 observations each, means and 95\% confidence intervals. Each run was allocated a single CPU core, see \autoref{appendix:runtimes} for details. 
    }
    \label{fig:metrics_runtime}
\end{figure*}

Figure \ref{fig:metrics_runtime} shows the recorded runtimes in minutes. We observed short runtimes for \ABC{} and \SABC{}, as these do not require NN training or MCMC. The sequential versions of all three NN-based algorithms yielded longer runtimes than the non-sequential versions because these involve 10 rounds of NN training. Among the sequential algorithms, \SNPE{} showed the longest runtimes. Runtimes with MAFs instead of NSFs tend to be faster, e.g.~the difference between MAFs and NSFs using \SNPE{} on SLCP at 100k simulations was about 50 minutes on average. We also emphasize that the speed of \XNLE{} reported here was only obtained after vectorizing MCMC sampling. Without vectorization, runtime on the Gaussian Linear for \SNLE{} was more than 36 hours instead of less than 2 hours (see \autoref{appendix:hyperparams}).

\newpage
\clearpage

%% file: appendix/T_tasks.tex
\renewcommand{\thesection}{T}
\section{Tasks}
\label{appendix:tasks}

\input{appendix/tasks/gaussian_linear}
\input{appendix/tasks/gaussian_linear_uniform}
\input{appendix/tasks/slcp}
\input{appendix/tasks/slcp_distractors}
\input{appendix/tasks/bernoulli_glm}
\input{appendix/tasks/bernoulli_glm_raw}
\input{appendix/tasks/gaussian_mixture}
\input{appendix/tasks/two_moons}
\input{appendix/tasks/sir}
\input{appendix/tasks/lotka_volterra}

%% file: appendix/tasks/gaussian_linear.tex
\subsection{Gaussian Linear}
\label{appendix:task:gaussian_linear}

Inference of the mean of a 10-d Gaussian model, in which the covariance is fixed. The (conjugate) prior is Gaussian:

{\setlength{\extrarowheight}{5pt}
\begin{tabularx}{\textwidth}{@{}ll@{}}
    \textbf{Prior} & %
        $\Normal(\bzero, 0.1 \odot \bI)$ \\
    \textbf{Simulator} & %
        \bigcell{l}{
            $ \bx|\btheta \sim \Normal(\bx|\bmm_{\btheta}=\btheta, \bS = 0.1 \odot \bI) $
        } \\
    \textbf{Dimensionality} & %
        $\btheta \in \bbR^{10}, \bx \in \bbR^{10}$ \\
\end{tabularx}}

%% file: appendix/tasks/gaussian_linear_uniform.tex
\subsection{Gaussian Linear Uniform}
\label{appendix:task:gaussian_linear_uniform}

Inference of the mean of a 10-d Gaussian model, in which the covariance is fixed. The prior is uniform:

{\setlength{\extrarowheight}{5pt}
\begin{tabularx}{\textwidth}{@{}ll@{}}
    \textbf{Prior} & %
        $\Uniform(-\bone, \bone)$ \\
    \textbf{Simulator} & %
        \bigcell{l}{
            $ \bx|\btheta \sim \Normal(\bx|\bmm_{\btheta}=\btheta, \bS = 0.1 \odot \bI) $
        } \\
    \textbf{Dimensionality} & %
        $\btheta \in \bbR^{10}, \bx \in \bbR^{10}$ \\
\end{tabularx}}

%% file: appendix/tasks/slcp.tex
\subsection{SLCP}
\label{appendix:task:slcp}

A challenging inference task designed to have a simple likelihood and a complex posterior. The prior is uniform over five parameters $\btheta$ and the data are a set of four two-dimensional points sampled from a Gaussian likelihood whose mean and variance are nonlinear functions of $\btheta$:

{\setlength{\extrarowheight}{5pt}
\begin{tabularx}{\textwidth}{@{}ll@{}}
    \textbf{Prior} & %
        $\Uniform(-\bthree, \bthree)$ \\
    \textbf{Simulator} & %
        \bigcell{l}{
            $ \bx|\btheta = (\bx_{1}, \ldots, \bx_{4})$, $ \bx_i \sim \Normal(\bmm_{\btheta}, \bS_{\btheta}) $, \\
            where $ \bmm_{\btheta} = \begin{bmatrix} {\theta_{1}} \\ {\theta_{2}} \end{bmatrix} $,
            $ \bS_{\btheta} = \begin{bmatrix} {s_{1}^{2}} & {\rho s_{1} s_{2}} \\ {\rho s_{1} s_{2}} & {s_{2}^{2}}\end{bmatrix} $,
            $ s_{1}=\theta_{3}^{2}, s_{2}=\theta_{4}^{2}, \rho=\tanh \theta_{5}$
        } \\
    \textbf{Dimensionality} & %
        $\btheta \in \bbR^5, \bx \in \bbR^8$ \\
    \textbf{References} & %
        \bigcell{l}{\cite{papamakarios2019a,greenberg2019,hermans2019} \\
        \cite{durkan2020}}
\end{tabularx}}

%% file: appendix/tasks/slcp_distractors.tex
\subsection{SLCP with Distractors}
\label{appendix:task:slcp_distractors}

This task is similar to \ref{appendix:task:slcp}, with the difference that we add uninformative dimensions (distractors) to the observation:

{\setlength{\extrarowheight}{5pt}
\begin{tabularx}{\textwidth}{@{}ll@{}}
    \textbf{Prior} & %
        $\Uniform(-\bthree, \bthree)$ \\
    \textbf{Simulator} & %
        \bigcell{l}{
            $ \bx|\btheta = (\bx_{1}, \ldots, \bx_{100})$,
            $\bx = p(\by)$, where $p$ re-orders the dimensions of $\by$ with a fixed
            random \\ permutation, \\
            $ \by_{[1:8]} \sim \Normal(\bmm_{\btheta}, \bS_{\btheta}) $,
            $ \by_{[9:100]} \sim \frac{1}{20}\sum_{i=1}^{20} t_2(\bmu^i,\bSigma^i)$ \\
            where $ \bmm_{\btheta} = \begin{bmatrix} {\theta_{1}} \\ {\theta_{2}} \end{bmatrix} $,
            $ \bS_{\btheta} = \begin{bmatrix} {s_{1}^{2}} & {\rho s_{1} s_{2}} \\ {\rho s_{1} s_{2}} & {s_{2}^{2}}\end{bmatrix} $,
            $ s_{1}=\theta_{3}^{2}, s_{2}=\theta_{4}^{2}, \rho=\tanh \theta_{5}$, \\
            $ \bmu^i \sim \Normal(0,15^2 \bI)$,
            $ \bSigma_{j,k}^i \sim \Normal(0,9)$, for $j>k$,
            $ \bSigma_{j,j}^i = 3 e^a$, where $a \sim \Normal(0,1)$,
            $ \bSigma_{j,k}^i = 0$ otherwise
        } \\
    \textbf{Dimensionality} & %
        $\btheta \in \bbR^5, \bx \in \bbR^{100}$ \\
    \textbf{References} & %
        \cite{greenberg2019}
\end{tabularx}}

%% file: appendix/tasks/bernoulli_glm.tex
\subsection{Bernoulli GLM}
\label{appendix:task:bernoulli_glm}

Inference of a 10-parameter Generalized linear model (GLM) with Bernoulli observations, and Gaussian prior with covariance matrix which encourages smoothness by penalizing the second-order differences in the vector of parameters \citep{DeNicolao1997}. The observations are the sufficient statistics for this GLM:

{\setlength{\extrarowheight}{5pt}
\begin{tabularx}{\textwidth}{@{}ll@{}}
    \textbf{Prior} & %
        \bigcell{l}{
            $\beta \sim \Normal(0,2)$,
            $\bff \sim \Normal(\bzero, (\bF^{\top} \bF)^{-1})$, \\
            $\bF_{i,i-2} = 1$, 
            $\bF_{i,i-1} = -2$,
            $\bF_{i,i} = 1+\sqrt{\frac{i-1}{9}}$,
            $\bF_{i,j} = 0$ otherwise,
            $1\leq i,j \leq9$
        }\\
    \textbf{Simulator} & %
        \bigcell{l}{
        $\bx|\btheta = (\bx_{1}, \ldots, \bx_{10})$,
        $\bx_1 = \sum_i^T z_i$,
        $\bx_{2:10} = \frac{1}{\bx_1} \bV \bz$, \\
        $z_i \sim \mathrm{Bern}(\eta(\mathbf{v}_i^{\top} \bff + \beta))$,
        $\eta(\cdot)=\exp(\cdot)/(1 + \exp(\cdot))$, \\
        frozen input between time bins $i-8$ and $i$:
        $\bv_i \sim \Normal(\bzero,\bI)$,
        $\bV = [v_1, v_2, \ldots, v_T]$
        } \\
    \textbf{Dimensionality} & %
        $\btheta \in \bbR^{10}, \bx \in \bbR^{10}$ \\
    \textbf{Fixed parameters} & %
        \bigcell{l}{
            Duration of task $T=100$.
        } \\
    \textbf{References} & %
        \cite{lueckmann2017,gonccalves2019training}
\end{tabularx}}

%% file: appendix/tasks/bernoulli_glm_raw.tex
\subsection{Bernoulli GLM Raw}
\label{appendix:task:bernoulli_glm_raw}

This task is similar to \ref{appendix:task:bernoulli_glm}, the sole difference being that the observations are not the sufficient statistics for the Bernoulli GLM process but the raw observations:

{\setlength{\extrarowheight}{5pt}
\begin{tabularx}{\textwidth}{@{}ll@{}}
    \textbf{Prior} & %
        \bigcell{l}{
            $\beta \sim \Normal(0,2)$,
            $\bff \sim \Normal(\bzero, (\bF^{\top} \bF)^{-1})$, \\
            $\bF_{i,i-2} = 1$, $\bF_{i,i-1} = -2$,
            $\bF_{i,i} = 1+\sqrt{\frac{i-1}{9}}$,
            $\bF_{i,j} = 0$ otherwise
            $1\leq i,j \leq9$
        }\\
    \textbf{Simulator} & %
        \bigcell{l}{
        $\bx|\btheta = (\bx_{1}, \ldots, \bx_{100})$,
        $x_i \sim \mathrm{Bern}(\eta(\mathbf{v}_i^{\top} \bff + \beta))$,
        $\eta(\cdot)=\exp(\cdot)/(1 + \exp(\cdot))$ \\
        frozen input between time bins $i-8$ and $i$:
        $\bv_i \sim \Normal(\bzero,\bI)$,
        } \\
    \textbf{Dimensionality} & %
        $\btheta \in \bbR^{10}, \bx \in \bbR^{100}$ \\
    \textbf{Fixed parameters} & %
            Duration of task $T=100$.\\
\end{tabularx}}

%% file: appendix/tasks/gaussian_mixture.tex
\subsection{Gaussian Mixture}
\label{appendix:task:gaussian_mixture}

This task is common in the ABC literature. It consists of inferring the common mean of a mixture of two two-dimensional Gaussian distributions, one with much broader covariance than the other:

{\setlength{\extrarowheight}{5pt}
\begin{tabularx}{\textwidth}{@{}ll@{}}
    \textbf{Prior} & %
        $\Uniform(-\mathbold{10}, \mathbold{10})$ \\
    \textbf{Simulator} & %
        \bigcell{l}{
            $ \bx|\btheta \sim 0.5 \; \Normal(\bx|\bmm_{\btheta}=\btheta, \bS = \bI) + $ $0.5 \; \Normal(\bx|\bmm_{\btheta}=\btheta, \bS = 0.01 \odot \bI) $
        } \\
    \textbf{Dimensionality} & %
        $\btheta \in \bbR^{2}, \bx \in \bbR^{2}$ \\
    \textbf{References} & %
        \cite{sisson2007,beaumont2009,toni2009,simola2020}
\end{tabularx}}

\newpage

%% file: appendix/tasks/two_moons.tex
\subsection{Two Moons}
\label{appendix:task:two_moons}

A two-dimensional task with a posterior that exhibits both global (bimodality) and local (crescent shape) structure to illustrate how algorithms deal with multimodality:

{\setlength{\extrarowheight}{5pt}
\begin{tabularx}{\textwidth}{@{}ll@{}}
    \textbf{Prior} & %
        $\Uniform(-\bone, \bone)$ \\
    \textbf{Simulator} & %
        \bigcell{l}{
            $ \boldsymbol{x} | \btheta = \begin{bmatrix}
              r \cos(\alpha)+0.25\\
              r \sin(\alpha)
            \end{bmatrix} + $
            $\begin{bmatrix}
              {-|\theta_1+\theta_2|}/{\sqrt{2}} \\
              {(-\theta_{1}+\theta_{2}})/{\sqrt{2}} \\
            \end{bmatrix} $,
            where $ \alpha \sim \Uniform(-\pi/{2}, \pi/2) $,
            $ r \sim \Normal(0.1,0.01^{2}) $
        } \\
    \textbf{Dimensionality} & %
        $\btheta \in \bbR^2, \bx \in \bbR^2$ \\
    \textbf{References} & %
        \cite{greenberg2019}

\end{tabularx}}

%% file: appendix/tasks/sir.tex
\subsection{SIR}
\label{appendix:task:sir}

The SIR model is an epidemiological model describing the dynamics of the number of individuals in three possible states: susceptible $S$, infectious $I$, and recovered or deceased $R$.

The SIR task consists in inferring the contact rate $\beta$ and the mean recovery rate $\gamma$, given a sampled number of individuals in the infectious group $I$ in 10 evenly-spaced points in time:

{\setlength{\extrarowheight}{5pt}
\begin{tabularx}{\textwidth}{@{}ll@{}}
    \textbf{Prior} & %
        \bigcell{l}{
            $\beta \sim \text{LogNormal}(\log(0.4), 0.5)$
            $\gamma \sim \text{LogNormal}(\log(1/8), 0.2)$
        } \\
    \textbf{Simulator} & %
        \bigcell{l}{
            $ \bx|\btheta = (x_{1}, \ldots, x_{10})$,
            $x_i \sim \mathcal{B}(1000,\frac{I}{N}) $, where I is simulated from \\
            $\frac{dS}{dt} = -\beta \frac{SI}{N}$ \\
            $\frac{dI}{dt} = \beta \frac{SI}{N}-\gamma I$ \\
            $\frac{dR}{dt} = \gamma I$
        } \\
    \textbf{Dimensionality} & %
        $\btheta \in \bbR^2, \bx \in \bbR^{10}$ \\
    \textbf{Fixed parameters} & %
        \bigcell{l}{
            Population size $N=1000000$ and duration of task $T=160$. \\
            Initial conditions:
            $(S(0),I(0),R(0)) = (N-1,1,0)$
        } \\
    \textbf{References} & %
        \cite{kermack1927}
\end{tabularx}}

%% file: appendix/tasks/lotka_volterra.tex
\subsection{Lotka-Volterra}
\label{appendix:task:lotka_volterra}

This is an influential model in ecology describing the dynamics of two interacting species, most commonly prey and predator interactions. Our task consists in the inference of four parameters $\btheta$ related to species interaction, given 20 summary statistics consisting of the number of individuals in both populations in 10 evenly-spaced points in time:

{\setlength{\extrarowheight}{5pt}
\begin{tabularx}{\textwidth}{@{}ll@{}}
    \textbf{Prior} & %
        \bigcell{l}{
            $\alpha \sim \text{LogNormal}(-0.125, 0.5)$,
            $\beta \sim \text{LogNormal}(-3, 0.5)$, \\
            $\gamma \sim \text{LogNormal}(-0.125, 0.5)$,
            $\delta \sim \text{LogNormal}(-3, 0.5)$
        } \\
    \textbf{Simulator} & %
        \bigcell{l}{
            $ \bx|\btheta = (\bx_{1}, \ldots, \bx_{10})$,
            $\bx_{1,i} \sim \text{LogNormal}(\log(X),0.1)$,
            $\bx_{2,i} \sim \text{LogNormal}(\log(Y),0.1)$, \\
            $X$ and $Y$ are simulated from \\
            $\frac{dX}{dt} = \alpha X - \beta X Y$ \\
            $\frac{dY}{dt} =  -\gamma Y + \delta X Y$
        } \\
    \textbf{Dimensionality} & %
        $\btheta \in \bbR^4, \bx \in \bbR^{20}$ \\
    \textbf{Fixed parameters} & %
        \bigcell{l}{
            Duration of task $T=20$. Initial conditions:
            $(X(0),Y(0))= (30,1)$
        }\\
    \textbf{References} & %
        \cite{lotka1920}
\end{tabularx}}

%% file: appendix/W_website.tex
\renewcommand{\thesection}{W}
\section{Website}
\label{appendix:website}

\begin{figure*}[h!]
    \centering
    \includegraphics[width=0.85\textwidth]{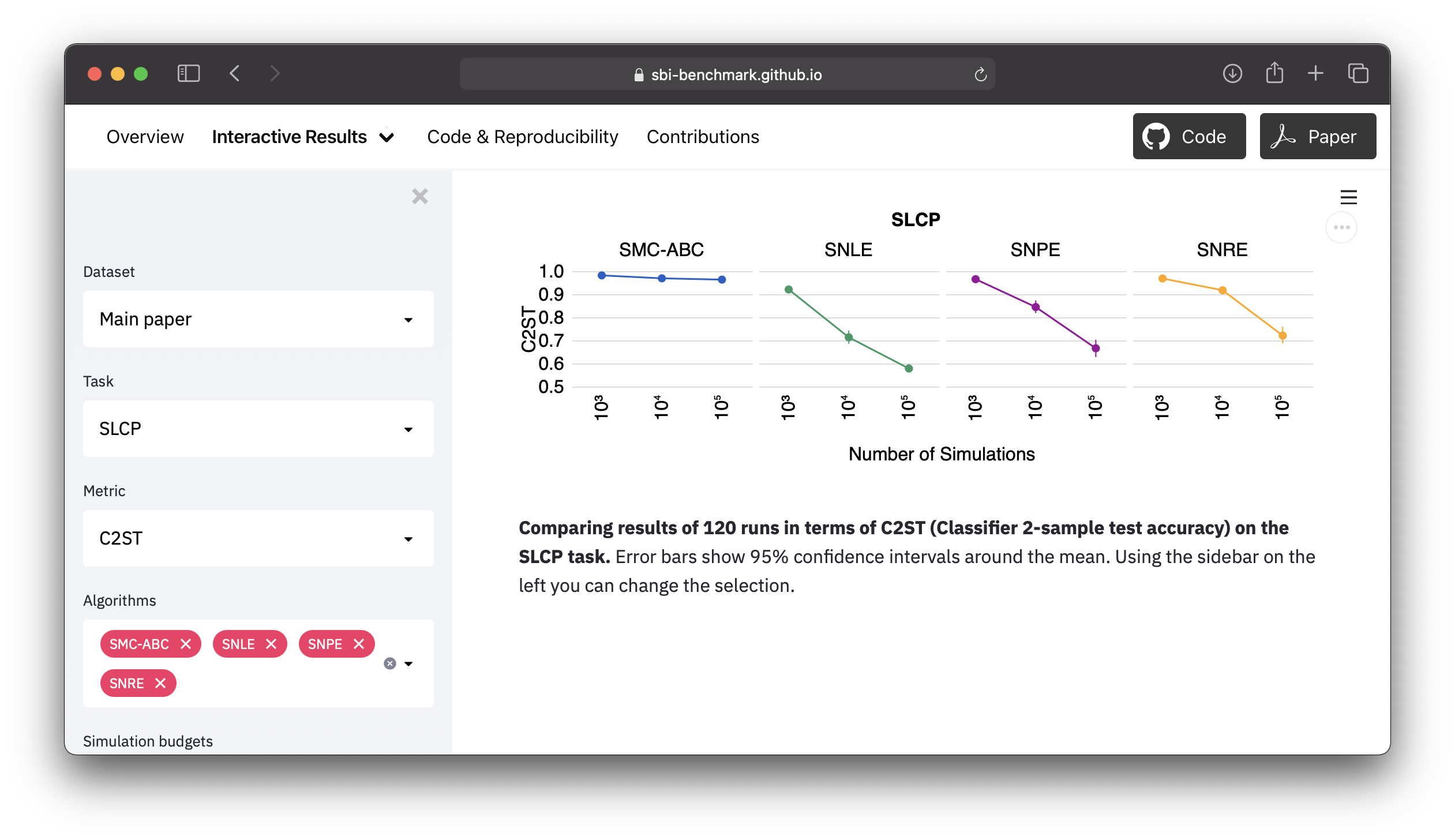}
    \includegraphics[width=0.85\textwidth]{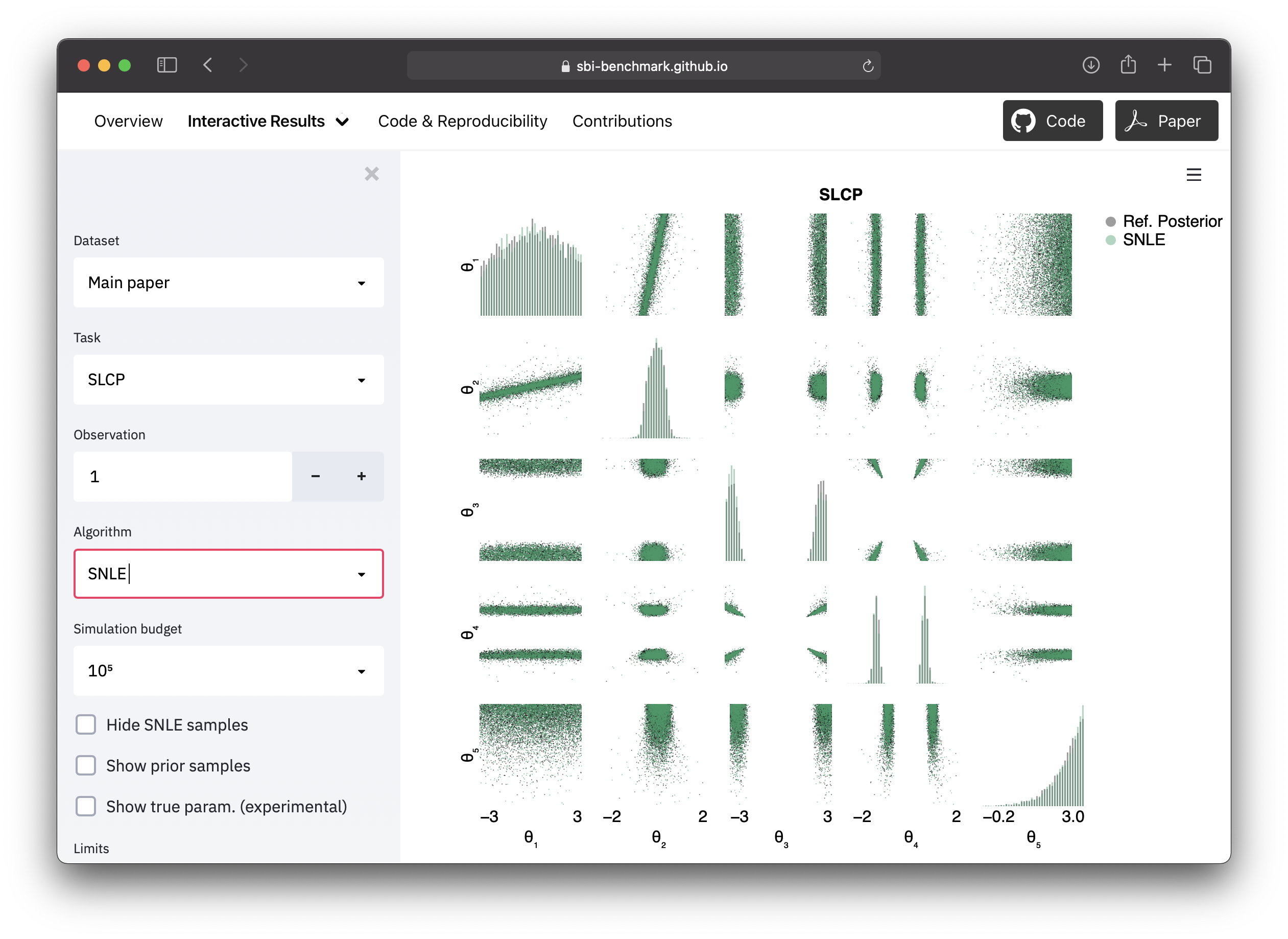}
    \caption{
        {\bf Screenshots from the companion website}. Top: Classification accuracy (C2ST) for a subset of sequential algorithms on the SLCP task. Bottom: \SNLE{} posterior on SLCP for $\bx_o^{(1)}$ at 100k simulations.
    }
    \label{fig:website}
\end{figure*}

The companion website (\linkwebsite{}) allows interactive comparisons in terms of all metrics. It also allows inspection of posterior samples of all runs, which we found insightful when choosing hyperparameters and diagnosing implementation issues. Two screenshots are provided in \autoref{fig:website}. \\[-0.5cm]

%
% Screenshots
%

